\documentclass{article}

\usepackage{microtype}
\usepackage{subcaption}
\usepackage{booktabs} 

\usepackage{hyperref}

\usepackage{caption}
\usepackage{tabularx}

\usepackage[T1]{fontenc}    
\usepackage{amsfonts}       
\usepackage{amssymb}
\usepackage{amsmath}

\usepackage[utf8]{inputenc}

\usepackage[pdftex]{graphicx}
\graphicspath{{images/}{./}}
\DeclareGraphicsExtensions{.pdf,.jpeg,.png}

\usepackage{algorithm}
\usepackage{algorithmic}
\usepackage{url}
\usepackage{wrapfig}

\usepackage{tikz}
\usetikzlibrary{bayesnet}

\usepackage{xr}

\makeatletter
\newcommand*{\addFileDependency}[1]{
  \typeout{(#1)}
  \@addtofilelist{#1}
  \IfFileExists{#1}{}{\typeout{No file #1.}}
}
\makeatother

\usepackage{multirow}
\usepackage{placeins}

\newcommand{\bs}{\boldsymbol}

\newcommand{\x}{{\bs{x}}}
\renewcommand{\xi}{\x^{i}}

\newcommand{\z}{{\bs z}}

\newcommand{\params}{{\bs \theta}}

\newcommand{\M}{\mathcal{M}}
\newcommand{\Mmodel}{\M_{\params}}
\newcommand{\Msamp}{\M_{\mathcal{S}}}

\newcommand{\pjoint}{\mathcal{P}}

\newcommand{\pdec}{p}
\newcommand{\penc}{q}

\newcommand{\Mdec}{\pdec_{\params}}
\newcommand{\Menc}{\penc_{\params}}

\newcommand{\encoder}{\Menc(\z \mid \x)}
\newcommand{\decoder}{\Mdec(\x \mid \z)}

\newcommand{\Pencoderjoint}{\Menc(\x , \z)}
\newcommand{\Pdecoderjoint}{\Mdec(\x , \z)}

\newcommand{\Penc}{\Menc(\z \mid \x)\, \pjoint(\x)}
\newcommand{\Pdec}{\Mdec(\x \mid \z)\, \pjoint(\z)}
\newcommand{\Mdecjoint}{\Mdec \left(\x, \z \right)}
\newcommand{\Mencjoint}{\Menc \left(\x, \z \right)}
\newcommand{\dxz}{\mathrm{d}\x\mathrm{d}\z}

\newcommand{\MIM}{{MIM }}

\newcommand{\AMIMloss}{\mathcal{L}_\text{A-MIM}}
\newcommand{\VAEloss}{\mathcal{L}_\text{VAE}}

\newcommand{\MIMloss}{\mathcal{L}_\text{MIM}}
\newcommand{\EMIMloss}{\hat{\mathcal{L}}_{{\text{MIM}}}}
\newcommand{\EMIMlosssub}[1]{\hat{\mathcal{L}}_{{\text{MIM}}_{#1}}}
\newcommand{\CEloss}{\mathcal{L}_\text{CE}}

\newcommand{\DKL}[2]{\mathcal{D}_\text{KL}\left(#1\,\|\, #2\right)}

\newcommand{\JSdivergence}{\mathcal{D}_\text{JS}}
\newcommand{\E}[2]{\mathbb{E}_{#1}\left[#2\right]}

\newcommand{\HD}[2]{H_{#1} \left( #2 \right)}
\newcommand{\CE}[2]{CE \left( \, #1 \,,\, #2 \, \right)}

\newcommand{\RMIM}{\mathrm{R}_{\mathrm{MIM}}}
\newcommand{\RH}{\mathrm{R}_\mathrm{H}}

\newcommand{\eg}{{\em e.g.}}
\newcommand{\ie}{{\em i.e.}}

\usepackage[preprint]{icml2020}

\icmltitlerunning{Mutual Information Machine}

\begin{document}
\twocolumn[
\icmltitle{MIM: Mutual Information Machine}



\icmlsetsymbol{equal}{*}
\begin{icmlauthorlist}
\icmlauthor{Micha Livne}{uoft,vector}
\icmlauthor{Kevin Swersky}{goo}
\icmlauthor{David J.~ Fleet}{uoft,vector}
\end{icmlauthorlist}

\icmlaffiliation{uoft}{Department of Computer Science, University of Toronto}
\icmlaffiliation{vector}{Vector Institute}
\icmlaffiliation{goo}{Google Research}

\icmlcorrespondingauthor{Micha Livne}{mlivne@cs.toronto.edu}
\icmlcorrespondingauthor{Kevin Swersky}{kswersky@google.com}
\icmlcorrespondingauthor{David J.~ Fleet}{fleet@cs.toronto.edu}

\icmlkeywords{Machine Learning, Autoencoders, ICML}

\vskip 0.3in
]



\printAffiliationsAndNotice{}  




\begin{abstract}
    We introduce the Mutual Information Machine (MIM), a probabilistic auto-encoder 
    for learning joint distributions over observations and latent variables.
    MIM reflects three design principles:
    1) low divergence, to encourage the encoder and decoder to learn 
    consistent factorizations of the same underlying distribution;
    2) high mutual information, to encourage an informative relation between data and latent variables;
    and
    3) low marginal entropy, or compression, which tends to encourage clustered latent representations.
    We show that a combination of the Jensen-Shannon divergence and the joint 
    entropy of the encoding and decoding distributions satisfies these 
    criteria, and admits a tractable cross-entropy bound that can be optimized
    directly with Monte Carlo and stochastic gradient descent.
    We contrast MIM learning with maximum likelihood and VAEs.
    Experiments show that MIM learns representations with high mutual 
    information, consistent encoding and decoding distributions, effective latent clustering,
    and data log likelihood comparable to VAE, while avoiding posterior collapse.
\end{abstract}

\section{Introduction} \label{sec:introduction}

Latent Variable Models (LVMs) are probabilistic models that enhance the distribution over
observations into a joint distribution over observations and latent variables, 
with VAE \citep{Kingma2013} being a canonical example. 
It is hoped that the learned representation will capture salient information in the observations,
which in turn can be used in downstream tasks (\eg, classification, inference, generation).
In addition, a fixed-size representation enables comparisons of observations with variable size (\eg, time series).
The VAE popularity stems, in part, from its versatility, serving as a generative model, 
a probability density estimator, and a representation learning framework.

Mutual information (MI) is often considered a useful measure of the quality 
of a latent representation (\eg, \cite{hjelm2018learning,Hjelm2018,Chen2016}).
Indeed, many common generative LVMs can be seen as optimizing an 
objective involving a sum of mutual information terms and a divergence
between encoding and decoding distributions~\citep{zhao2018LagrangeVAE}.
These include the VAE \citep{Kingma2013}, 
$\beta$-VAE \citep{Higgins2017betaVAELB}, and InfoVAE \citep{Zhao2017InfoVAEIM}, 
as well as GAN models including ALI/BiGAN \citep{dumoulin2016adversarially, Donahue2016BiGAN}, 
InfoGAN \citep{Chen2016}, and CycleGAN \citep{CycleGAN2017}.
From an optimization perspective, however, different objectives can be challenging, 
as MI is notoriously difficult to estimate \citep{Hjelm2018}, and many choices of 
divergence require adversarial training.

In this paper, we introduce a class of generative LVMs over data $\x$ and 
latent variables $\z$ called the Mutual Information Machine (MIM).
The learning objective for MIM is designed around three fundamental principles,
namely
\vspace*{-0.2cm}
\begin{enumerate}
    \itemsep 0cm
    \item Consistency of  encoding and decoding distributions;
    \item High mutual information between $\x$ and $\z$;
    \item Low marginal entropy.
\end{enumerate}
\vspace*{-0.2cm}
Consistency enables one to both generate data and infer latent
variables from the same underlying joint
distribution~\cite{dumoulin2016adversarially, Donahue2016BiGAN, pu2017adversarial}.
High MI ensures that the latent variables accurately capture the
factors of variation in the data.
Beyond consistency and mutual information, our third criterion ensures that each distribution 
efficiently encodes the required information, and does not also model spurious correlations.

MIM is formulated using 1) the Jensen-Shannon divergence (JSD),
a symmetric divergence that also forms the basis of ALI/BiGANs and ALICE,
\citep{dumoulin2016adversarially,Donahue2016BiGAN, li2017alice}, and 
2) the entropy of the encoding 
and decoding distributions, encouraging high mutual information and low marginal entropy.
We show that the sum of these two terms reduces to the entropy of 
the mixture of the encoding and decoding distributions defined by the JSD.
Adding parameterized priors, we obtain a novel upper bound on this entropy,
which forms the MIM objective.
Importantly, the resulting cross-entropy objective enables direct optimization using 
stochastic gradients with reparameterization, thereby avoiding the need for 
adversarial training and the estimation of mutual information.
MIM learning is stable and robust, yielding consistent encoding and decoding 
distributions while avoiding over- and under-estimation of marginals 
(cf.\ \cite{pu2017adversarial}), and avoiding posterior collapse in the conventional VAE.

We provide an analysis of the MIM objective and further broaden the MIM framework,
exploring different sampling distributions and forms of consistency.
We also demonstrate the benefits of MIM when compared to VAEs experimentally,
showing that MIM in particular benefits greatly from more expressive model architectures
by utilizing the additional capacity for better representation (\ie, compressed representation 
with higher mutual information).

\section{Generative LVMs}
\label{sec:lvm}

We consider the class of generative latent variable models (LVMs) over data $\x \in \mathcal{X}$ and latent variables $\z \in \mathcal{Z}$. 
These models generally assume an explicit prior over latent variables, $\pjoint(\z)$, and an unknown distribution over data, $\pjoint(\x)$, specified implicitly through data examples. As these distributions are fixed and not learned, we call  them  \emph{anchors}. The joint distribution over $\x$ and $\z$ is typically expressed 
in terms of an encoding distribution, $\Penc$, or a decoding distribution $\Pdec$, 
where $\encoder$ and $\decoder$ are known as the encoder and decoder.

\citet{zhao2018LagrangeVAE} describe a learning objective that succinctly 
encapsulates many LVMs proposed to date,
\begin{align}
    \mathcal{L}(\theta) &= \alpha_1 I_q(\x;\z) + \alpha_2 I_p(\x;\z)  + \mathbf{\lambda}^\top \mathcal{D}~.
    \label{eq:zhao-objective}
\end{align}
where $\theta$ comprises the model parameters, $\alpha_1, \alpha_2, \mathbf{\lambda}$ are 
weights, $\mathcal{D}$ is a set of divergences, and $I_q(\x; \z)$ and $I_p(\x, \z)$ 
represent mutual information under the encoding and decoding distributions. 
The divergences measure inconsistency between the encoding and decoding distributions.
By encouraging high MI one hopes to learn meaningful relations between $\x$ and $\z$
\cite{Higgins2017betaVAELB, Zhao2017InfoVAEIM, Chen2016, CycleGAN2017, li2017alice}.


One of the best-known examples within this framework is the variational auto-encoder 
(VAE)~\cite{Kingma2013, Rezende2014}, which sets $\alpha_1\!=\!\alpha_2\!=\!0$, 
$\mathbf{\lambda}\!=\!1$, and the divergence $\mathcal{D}$ to
\begin{align}
    \DKL{\Penc}{\Pdec} ~. 
    \label{eq:vaekl}
\end{align}
One can show that this objective is equivalent, up to an additive constant, to the evidence lower bound typically 
used to specify VAEs. 
Although widely used, two issues can arise when training a VAE. First, the asymmetry of the KL divergence can lead it to assign high probability to unlikely regions of the data distribution~\cite{pu2017adversarial}. Second, it sometimes learns an encoder that essentially 
ignores the input data and instead models the latent prior. 
This is known as \emph{posterior collapse}. 
The consequence is that latent states convey little useful information about observations.

Another example is the ALI/BiGAN model \cite{dumoulin2016adversarially, Donahue2016BiGAN}, 
which is instead defined by the Jensen-Shannon divergence:
\begin{eqnarray}
    \JSdivergence &=&
    \frac{1}{2} \Big( \, \DKL{\Pdec}{\Msamp} \nonumber \\
    && \quad + ~ \DKL{\Penc}{\Msamp}\, \Big) ~, 
    \label{eqn:JSD}
\end{eqnarray}
where $\Msamp$ is an equally weighted mixture of the encoding
and decoding distributions; \ie,
\begin{eqnarray}
   \Msamp ~=~ \frac{1}{2} \big( \,\Pdec + \Penc \, \big)  ~.
   \label{eqn:msamp}
\end{eqnarray}
The symmetry in the objective helps to keep the marginal distributions consistent, as with the symmetric KL objective used in \cite{pu2017adversarial}.

As described by \citet{zhao2018LagrangeVAE}, many such methods belong to a difficult class 
of objectives that usually rely on adversarial training, which can be unstable.
In what follows we show how by further encouraging low marginal entropy, and with a particular
combination of MI and divergence, we obtain a framework that accomplishes many of the 
aims of the previous methods while also allowing for stochastic gradient-based optimization.

\section{Mutual Information Machine}
\label{sec:mim}

\subsection{Symmetry and Joint Entropy}
\label{sec:symmetry_and_entropy}
We begin with an objective that satisfies our desiderata of symmetry, high mutual information, 
and low marginal entropy, \ie, the sum of $\JSdivergence$ in \eqref{eqn:JSD} and $\RH$,
defined by
\begin{eqnarray}
    \RH &=& \frac{1}{2}(H_q(\x, \z) + H_p(\x, \z)) ~ .
    \label{eq:RH}  
\end{eqnarray}
Here, 
$H_q(\x, \z)$
is the joint entropy under the encoding distribution, which can also be expressed 
as $H_q(\x) + H_q(\z) - I_q(\x; \z)$, 
the sum of marginal entropies w.r.t.\ $\x$ and $\z$ minus their mutual information.
Analogously, $H_p(\x, \z)$ is the joint entropy under the decoding distribution.
Note that if symmetry is ever achieved, \ie, $\JSdivergence=0$, then the
marginal entropy terms become constant and $\RH$ directly targets mutual
information. This is unlikely to happen in practice, but would roughly
hold for small divergence values.

The objective $\JSdivergence + \RH$ encourages consistency through the JSD,
and high MI and low marginal entropy through $\RH$.
Interestingly, one can also show that this specific combination of objective 
terms reduces to the entropy of the mixture distribution $\Msamp$; \ie,
\begin{eqnarray}
    H_{\Msamp}(\x, \z) &=& \JSdivergence + \RH ~ . \label{eq:jsd-h}
\end{eqnarray}
While $H_{\Msamp}(\x, \z)$ is intractable to compute, as it contains the 
unknown $\pjoint(\x)$ in the log term, it does suggest the existence of a
tractable upper bound that can be optimized straightforwardly with stochastic 
gradients and reparameterization~\cite{Kingma2013, Rezende2014}.
This bound, introduced below, forms the MIM objective.

\subsection{The MIM Objective}

Section \ref{sec:symmetry_and_entropy} formulates a loss function
\eqref{eq:jsd-h} that reflects our desire for a consistent and expressive model,
however direct optimization is intractable. Adversarial optimization is feasible,
however it can also be unstable. As an alternative, we introduce parameterized approximate priors,
$\Menc(\x)$ and $\Mdec(\z)$, to derive a tractable upper-bound.
This is similar in spirit to VAEs, which introduce a parameterized approximate posterior.
These parameterized priors, together with the encoder and decoder,
$\Menc(\z | \x)$ and $\Mdec(\x | \z)$, comprise a new pair of joint distributions,
$\Menc (\x, \z ) \equiv \Menc\left(\z| \x\right) \Menc\left(\x\right)$, 
and $\Mdec (\x, \z )  \equiv \Mdec\left(\x| \z\right) \Mdec\left(\z\right)$.

These new joint distributions allow us to bound \eqref{eq:jsd-h}; \ie,
\begin{eqnarray}
    \CEloss
   &\equiv& \CE{\Msamp}{\Mmodel} 
    \\
    &=& \DKL{\Msamp}{\Mmodel} + H_{\Msamp}(\x, \z)
    \nonumber \\
    &\geq&  H_{\Msamp}(\x, \z) ~ , \nonumber 
    \label{eq:LCE}
\end{eqnarray}
where $\CE{\Msamp}{\Mmodel} $ denotes the cross-entropy between $\Msamp$ and $\Mmodel$,
and $\Mmodel =  \frac{1}{2} (  \Mdecjoint + \Mencjoint )$.
This cross-entropy loss aims to match the model prior distributions to the anchors, while also minimizing $H_{\Msamp}(\x, \z)$. 
Importantly, it can be trained by Monte Carlo sampling from the anchor distributions with reparameterization.

The penalty we pay in exchange for stable optimization
is the need to specify prior distributions $\Menc(\x)$ and $\Mdec(\z)$.
We could set $\Mdec(\z)=\pjoint(\z)$, though allowing it to vary provides more flexibility
to minimize other aspects of the objective. We do not have a closed form expression
for $\pjoint(\x)$, however $\Menc(\x)$ can be any data likelihood model,
\ie, a normalizing flow~\citep{Dinh2014, Dinh2016a, Rezende2015},
an auto-regressive model~\citep{Larochelle2011, van2016conditional},
or even as an implicit distribution (see Section~\ref{sec:learning-implicit}).
In this paper, for fair comparison with a VAE, we use an implicit distribution.
In this way, MIM has the same set of parameters as the corresponding VAE,
and they only differ in their respective loss functions.
The use of an auxiliary likelihood model has been explored in other contexts,
\eg, the noise distribution in noise contrastive estimation~\citep{gutmann2010noise}.


While $\CEloss$ encourages consistency between model and anchored 
distributions, \ie, $\Mdec (\x, \z ) \approx \Mdec(\x | \z)\pjoint(\z)$ and
$\Menc(\x, \z) \approx \Menc(\z | \x)\pjoint(\x)$, it does not directly
encourage model consistency, $\Mdec (\x, \z ) \approx \Menc(\x, \z)$.
To remedy this, we upper-bound $\CEloss$ using Jensen's inequality to obtain
the MIM loss, $\MIMloss \ge  \CEloss$, where
\begin{eqnarray}
   \MIMloss &\equiv& \frac{1}{2}
    \big(\, \CE{\Msamp}{\Menc(\x, \z)} \nonumber \\
    && \qquad + ~ \CE{\Msamp}{\Mdec (\x, \z )} \, \big) ~ .
    \label{eq:mimloss}
\end{eqnarray}
The loss for the Mutual Information Machine is an average of cross entropy terms between 
the mixture $\Msamp$ and the model encoding and decoding distributions.
To see that this encourages model consistency, it can be shown that $\MIMloss$ is
equal to $\CEloss$ plus a non-negative regularizer:
\begin{align}
    \MIMloss ~= ~\CEloss + \RMIM ~,
\end{align}
where, as shown in supplementary material, 
\begin{eqnarray}
   \! \RMIM \!\!\!\!\!  &=& \! \!\! \frac{1}{2}
    \big( \DKL{\Msamp}{\Mdecjoint} \! +\! \DKL{\Msamp}{\Mencjoint} \!\big) \nonumber \\
    & & \quad - \DKL{\Msamp}{\Mmodel} 
    \label{eq:RMIM-DKL}  
\end{eqnarray}
One can conclude from \eqref{eq:RMIM-DKL} that the regularizer $\RMIM$ is zero 
only when the two model distributions, $\Menc \left(\x, \z \right)$
and $\Mdec \left(\x, \z \right)$, are identical under fair samples from 
the joint sample distribution $\Msamp \left(\x, \z \right)$.
In practice we find that encouraging model consistency also helps stabilize learning.

To understand the MIM objective in greater depth, it is helpful to express $\MIMloss$ 
as a sum of fundamental terms that provide intuition for its behavior.  \ie,
one can show that:
\begin{align}
   \! \MIMloss\, &=  ~
    \mathrm{R}_\mathrm{H} \, +\,
    \frac{1}{4}\big(\, \DKL{\pjoint(\z)}{ \Mdec(\z)}  \nonumber \\
     & \quad ~+ \,\DKL{\pjoint(\x)}{\Menc(\x)} \big)
    \nonumber \\
    & \quad ~ +\,\, \frac{1}{4}\big(\,\DKL{\Penc }{ \Mdec(\x, \z)} \nonumber \\
    & \quad ~ +\, \DKL{\Pdec}{ \Menc(\z , \x)} \big) ~ .
    \label{eq:MIM-parts}
\end{align}
The first term in Eqn.\ \eqref{eq:MIM-parts} encourages high mutual
information between observations and latent states. The second shows that MIM
directly encourages the model priors to match the anchor distributions.
Indeed, the KL term between the data anchor and the model prior is the maximum
likelihood objective.
The third term encourages consistency between model and anchored distributions, 
in effect fitting the model decoder to samples drawn from the anchored encoder 
(cf.\ VAE), and, via symmetry, fitting the model encoder to samples drawn from 
the anchored decoder (both with reparameterization).
As such, MIM can be seen as simultaneously training and distilling a model
distribution over the data into a latent variable model.
The idea of distilling density models has been used in other domains, \eg,
for parallelizing auto-regressive models~\citep{oord2017parallel}.


\subsection{Asymmetric Mutual Information Machine}
\label{sec:amim}

With some models, like auto-regressive distributions (\eg, PixelHVAE~\cite{DBLP:journals/corr/TomczakW17}),
sampling from $\Mdec(\x \mid \z)$ becomes impractical during learning.
This is problematic for MIM, as it draws samples from $\Msamp$, a mixture of the anchored
encoding and decoding distributions.
As an alternative, we introduce a variant of MIM, called A-MIM,
that only samples from the encoding distribution.
To that end, collecting terms in Eqn.\ \eqref{eq:MIM-parts} which 
depend on samples from $\Penc$, we obtain
\begin{eqnarray}
    \!\AMIMloss\!\! &\equiv& \!\!
     \frac{1}{2} \big(\, \CE{\Msamp^\penc}{\Menc (\x, \z)} \nonumber  \\
     &~& \quad \, + ~   \CE{\Msamp^\penc}{\Mdec (\x, \z )} \, \big)  
     \label{eq:amimloss} \\
    &=& \!\!\! \frac{1}{2}H_q(\x, \z)  
    ~+~ \frac{1}{2} \DKL{\pjoint(\x)}{\Menc(\x)}  \nonumber \\
    &~& \,+\, \frac{1}{2} \DKL{\Penc }{ \Mdec(\x, \z)} ~,~
            \label{eq:AMIM-parts}
\end{eqnarray}
where $\Msamp^\penc = \Menc(\z\,|\,\x)\, \pjoint (\x)$.
Minimization of $\AMIMloss(\params)$ learns a consistent encoder-decoder model
with an encoding distribution with high mutual information and low entropy.
Formally, one can derive the following bound:
\begin{eqnarray}
    \AMIMloss
    ~\geq~ \CE{\Msamp^\penc}{\Mmodel}
    ~\geq~ H_q(\x, \z) ~,
    \label{eq:A-MIM-asymmetric-bound}
\end{eqnarray}
where the gap between $\AMIMloss$ and the cross-entropy term has the same form 
as $\RMIM$ in Eqn.\ \eqref{eq:RMIM-DKL}, but with $\Msamp$ replaced by $\Msamp^\penc$. 
It is still $0$ when $\Menc(\x, \z) = \Mdec(\x, \z)$.

The main difference between MIM and A-MIM is the lack of symmetry in the sampling distribution.
This is similar to the VAE formulation in Eqn.~\eqref{eq:vaekl}.
Indeed, the final term in Eqn.~\eqref{eq:AMIM-parts} is exactly
the VAE loss, with an optionally parameterized latent prior $\Mdec(\z)$.
Importantly, $\Menc(\x)$ must be defined implicitly using the marginal
of the decoding distribution, otherwise it will be completely independent
from the other terms. Without this, A-MIM can be seen as a VAE with a joint
entropy penalty, and we have found that this doesn't work as well in practice.

In essence, A-MIM trades symmetry of the JSD for speed. 
Regardless, we show that A-MIM is often as effective as MIM at learning representations.


\subsection{A-MIM, VAEs, and Posterior Collapse}
The VAE loss in 
Eqn.\ \eqref{eq:vaekl} can be expressed in a form that bears similarity 
to the A-MIM loss in Eqn.\ \eqref{eq:amimloss} (details in the supplementary material); \ie, 
\begin{eqnarray}
\! \VAEloss \!\!\!\! &=& \!\!\!\!
\frac{1}{2} \! \Big(  \CE{\Msamp^\penc}{\Menc(\z|\x)\,\pjoint(\x)} \nonumber \\
&& \!\!\!\! +\, \CE{\Msamp^\penc}{\Mdec(\x|\z)\,\pjoint(\z)} \Big) - H_q(\x, \z) .
~~~~~~~
\label{eq:vi-as-ml-objective-text}
\end{eqnarray}
Like $\AMIMloss$ in  Eqn.\ \eqref{eq:amimloss}, the first term in 
Eqn.\ \eqref{eq:vi-as-ml-objective-text} is the average of two cross-entropy 
terms between a sample distribution and the encoding and decoding distributions. 
Like $\AMIMloss$, here they are asymmetric, as samples are drawn only from the 
encoding distribution, and the last term in Eqn.\ \eqref{eq:vi-as-ml-objective-text}
encourages high marginal entropies and low MI under the encoding distribution.
This plays a significant role in allowing for posterior collapse (\eg, see 
\cite{zhao2018LagrangeVAE,he2018lagging}).

\section{Learning} 
\label{sec:learning}

\subsection{Standard MIM Learning}
Here we provide a detailed description of MIM learning.
The training procedure is summarized in Algorithm~\ref{algo:mim}.
It is very similar to training a VAE; gradients are taken by sampling using the reparameterization trick~\cite{Kingma2013, Rezende2014}.

\begin{figure}[t]
\centering
\begin{minipage}[t]{0.95\columnwidth}
\begin{algorithm}[H]
    \caption{\MIM learning of parameters $\params$}
    \label{algo:mim}
    \begin{algorithmic}[1]
        \REQUIRE Samples from anchors $\pjoint(\x),\pjoint(\z)$
        \WHILE{not converged}
        \STATE $D_\mathrm{dec} \gets \{ \x_i, \z_i \sim \Mdec(\x|\z)\pjoint(\z) \}_{i=1}^{N/2}$
        \STATE $D_\mathrm{enc} \gets \{ \x_j, \z_j \sim \Menc(\z|\x)\pjoint(\x) \}_{j=1}^{N/2}$
        \STATE $D \gets D_\mathrm{dec} \bigcup D_\mathrm{enc}$
        \STATE $\EMIMloss \left( \params ; D \right) = \frac{-1}{2N}\! \sum_{i=1}^{N}\! \big( \log \Menc(\x_i, \z_i)  + \log \Mdec(\x_i, \z_i) \big)$
        \STATE \textcolor{gray}{\textit{\# Minimize loss}}
        \STATE $\Delta \params \propto -\nabla_{\params}  \EMIMloss \left( \params ; D \right)$
        \ENDWHILE
    \end{algorithmic}
\end{algorithm}
\end{minipage}
\end{figure}


\subsection{Learning with Marginal $\Menc(\x)$}
\label{sec:learning-implicit}
When we would like to use a marginal data prior,
\begin{eqnarray}
\Menc(\x) &=& \mathbb{E}_{\Mdec(\z)}[\Mdec(\x \mid \z)] ~,
\end{eqnarray}
we must modify the learning algorithm slightly. This is because
the MIM objective in Equation~\eqref{eq:mimloss} involves a
$\log \Menc(\x)$ term, and the marginal is intractable
to compute analytically. Instead we minimize an upper bound on the
necessary integrals using Jensen's inequality. Complete details of the derivations
are given below.

For the encoder portion of $\Msamp$, the integral can be bounded and approximated as follows.
\begin{eqnarray*}
    && \hspace{-1.0cm} \mathbb{E}_{\Menc(\z \mid \x)\pjoint(\x)}[\log(\Menc(\x))] \\
    &=& \mathbb{E}_{\Menc(\z \mid \x)\pjoint(\x)}[\log(\mathbb{E}_{\Mdec(\z')}[\Mdec(\x \mid \z')])] \\
    &\geq& \mathbb{E}_{\Menc(\z \mid \x)\pjoint(\x)}\left [\log \left (\frac{\Mdec(\x \mid \z)\Mdec(\z)}{\Menc(\z \mid \x)} \right )\right ] \\
    &\approx& \log(\Mdec(\tilde{\x} \mid \tilde{\z})) - \log(\Menc(\tilde{\z} \mid \tilde{\x})) + \log(\Mdec(\tilde{\z}))~.
\end{eqnarray*}
where $\tilde{\x} \sim \pjoint(\x)$ and $\tilde{\z} \sim \Menc(\z \mid \tilde{\x})$.

Similarly, the decoder portion can be bounded and approximated as follows.
\begin{align*}
& \hspace*{-0.5cm} \mathbb{E}_{\Mdec(\x \mid \z)\pjoint(\z)}[\log(\Menc(\x))] \\
=& ~\, \mathbb{E}_{\Mdec(\x \mid \z)\pjoint(\z)}[\log(\mathbb{E}_{\Mdec(\z')}[\Mdec(\x \mid \z')])] \\
\geq& ~\, \mathbb{E}_{\Mdec(\x \mid \z)\pjoint(\z)\Menc(\z'|\x)}\left [\log\left (\frac{\Mdec(\x \mid \z')\Mdec(\z')}{\Menc(\z'|\x)}\right )\right ] \\
\approx&~\, \log(\Mdec(\tilde{\x} \mid \tilde{\z'}))) + \log(\Mdec(\tilde{\z'})) - \log(\Menc(\tilde{\z'} \mid \x')).
\end{align*}
where $\tilde{\z} \sim \pjoint(\z)$, $\tilde{\x} \sim \Mdec(\x \mid \tilde{\z})$, and $\tilde{z'} \sim \Menc(\z' \mid \tilde{\x})$.

In both cases, we use importance sampling from $\Menc(\z \mid \x)$.
The full algorithm, collecting all like terms, is summarized in 
Algorithm~\ref{algo:vae-as-mim}.

\begin{figure}[t]
    \centering
    \begin{minipage}[t]{0.95\columnwidth}
        \begin{algorithm}[H]
        \caption{\MIM learning with marginal $\Menc(\x)$}
        \label{algo:vae-as-mim}
        \begin{algorithmic}[1]
        \REQUIRE Samples from anchors $\pjoint(\x),\pjoint(\z),\Menc(\z\mid\x)$
        \REQUIRE Define $\Menc(\x) = \E{\z \sim \Mdec(\z)}{\Mdec(\x\mid\z)}$
        \WHILE{not converged}
        \STATE $D_\mathrm{enc} \gets \{ \x_i, \z_i \sim \Menc(\z\mid\x)\pjoint(\x) \}_{i=1}^{N}$
        \STATE $\EMIMlosssub{\mathrm{enc}} \gets  -\frac{1}{N}\sum_{i=1}^{N} \big( \log \Mdec(\x_i\mid\z_i)$ \\
             ~~~ $+ \log \Mdec(\z_i) \big)$
        \STATE $D_\mathrm{dec} \gets \{ \x_i, \z_i, \z_i' \sim \Mdec(\x\mid\z)\pjoint(\z)$ \\
             ~~~ $\Menc(\z'\mid\x) \}_{i=1}^{N}$
        \STATE $\EMIMlosssub{\mathrm{dec}} \gets -\frac{1}{2N} \sum_{i=1}^{N} \big(\log \Menc(\z_i\mid\x_i)$ \\
                 ~~~ $+ \log \Mdec(\x_i\mid\z_i) + \log \Mdec(\z_i) + \log \Mdec(\x \mid \z_i')$ \\
                 ~~~ $+ \log \Mdec(\z_i') - \log \Menc(\z_i' \mid \x_i)\big)$
        \STATE \textcolor{gray}{\textit{\# Minimize loss}}
        \STATE $\Delta \params \propto -\nabla_{\params} ~ \frac{1}{2} \left( \EMIMlosssub{\mathrm{enc}} + \EMIMlosssub{\mathrm{dec}} \right)$
        \ENDWHILE
        \end{algorithmic}
        \end{algorithm}
    \end{minipage}
    \vspace*{-0.3cm}
\end{figure}

\section{Experiments} \label{sec:experiments}

We next examine MIM empirically, often with a VAE baseline.  
We use both synthetic datasets and well-known image datasets, namely MNIST \citep{LeCun1998}, 
Fashion-MNIST \citep{DBLP:journals/corr/abs-1708-07747} and Omniglot \citep{Lake2015}. 
All models were trained using Adam \cite{2014arXiv1412.6980K} 
with a learning rate of $10^{-3}$, and a mini-batch size of 128. 
Following \cite{DBLP:journals/corr/abs-1711-00464} we 
stabilize training by increasing the weight on the second (decoder) CE term 
in Eqn.\ \eqref{eq:mimloss}, from 0 to 0.5, in several 'warm-up' epochs.
Training continues until the loss, Eqn.\ \eqref{eq:mimloss}, on a held-out validation set 
has not improved for the same number of epochs as the warm-up steps (\ie, defined per experiment).
The number of epochs to convergence of MIM learning is usually 2 to 5 times 
greater than a VAE with the same architecture. A more complete description of MIM learning 
is provided in the supplementary material.

\subsection{Synthetic Data} 
\label{sec:posterior-collapse-mim-vae}


\begin{figure*}[t]
    \centering
    \setlength{\tabcolsep}{0pt}
     \begin{tabular}{cccccccc} 
      {\scriptsize VAE} & {\scriptsize MIM} & \hspace*{0.25cm} & {\scriptsize VAE} & {\scriptsize MIM} & \hspace*{0.25cm} & {\scriptsize VAE} & {\scriptsize MIM} \\
      \includegraphics[width=0.235\columnwidth]{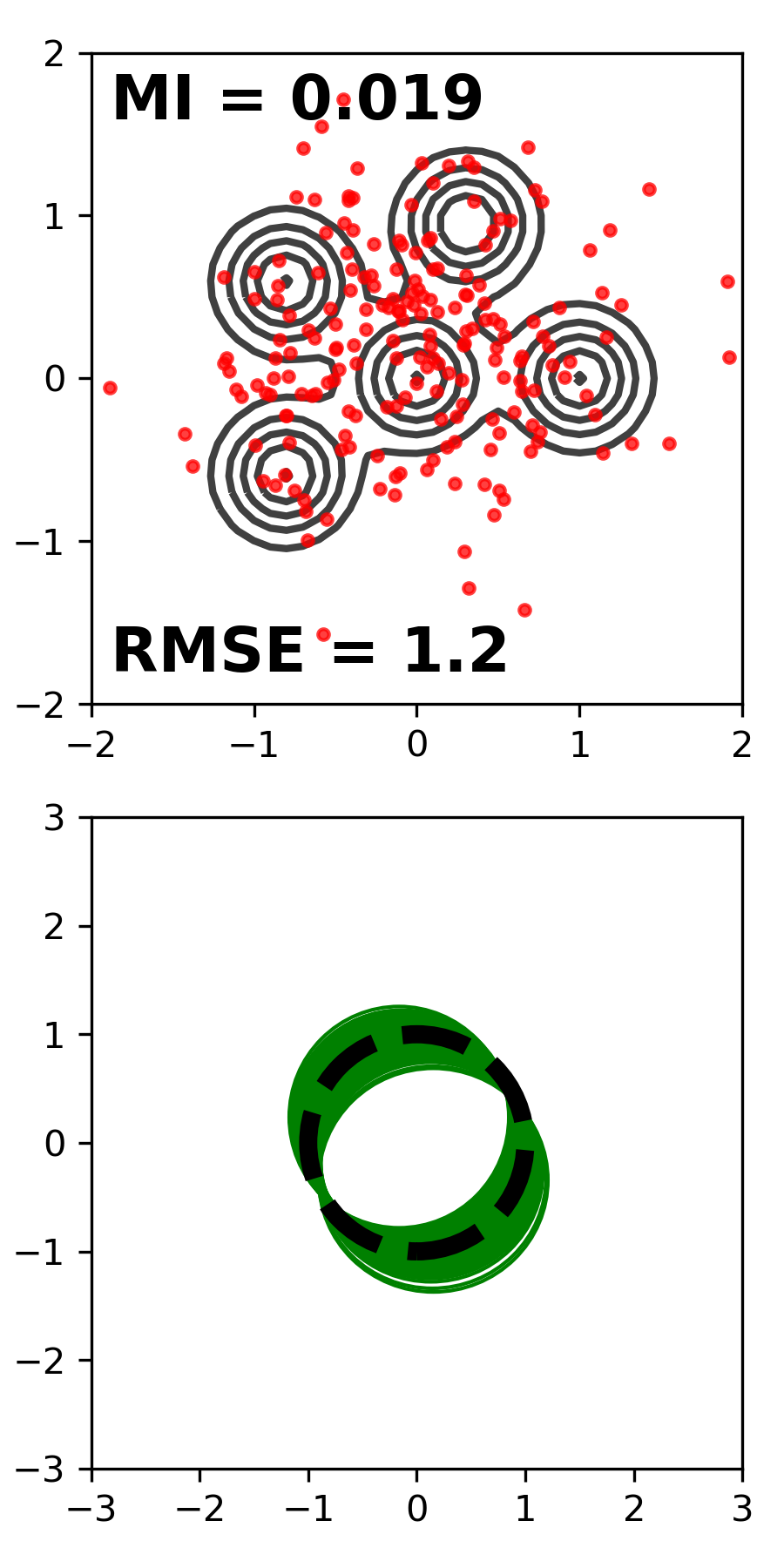}
    & \includegraphics[width=0.235\columnwidth]{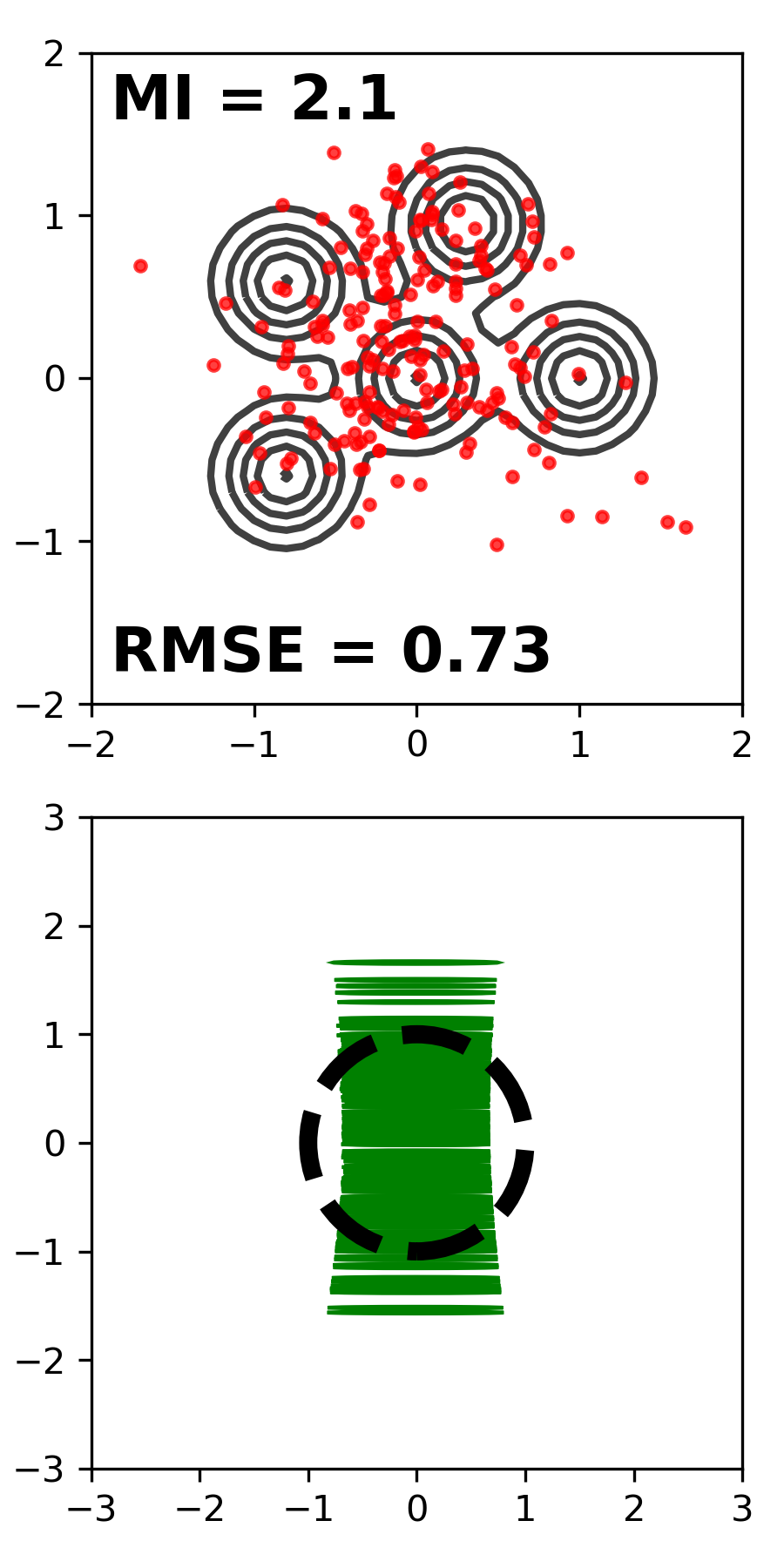}
    & & \includegraphics[width=0.235\columnwidth]{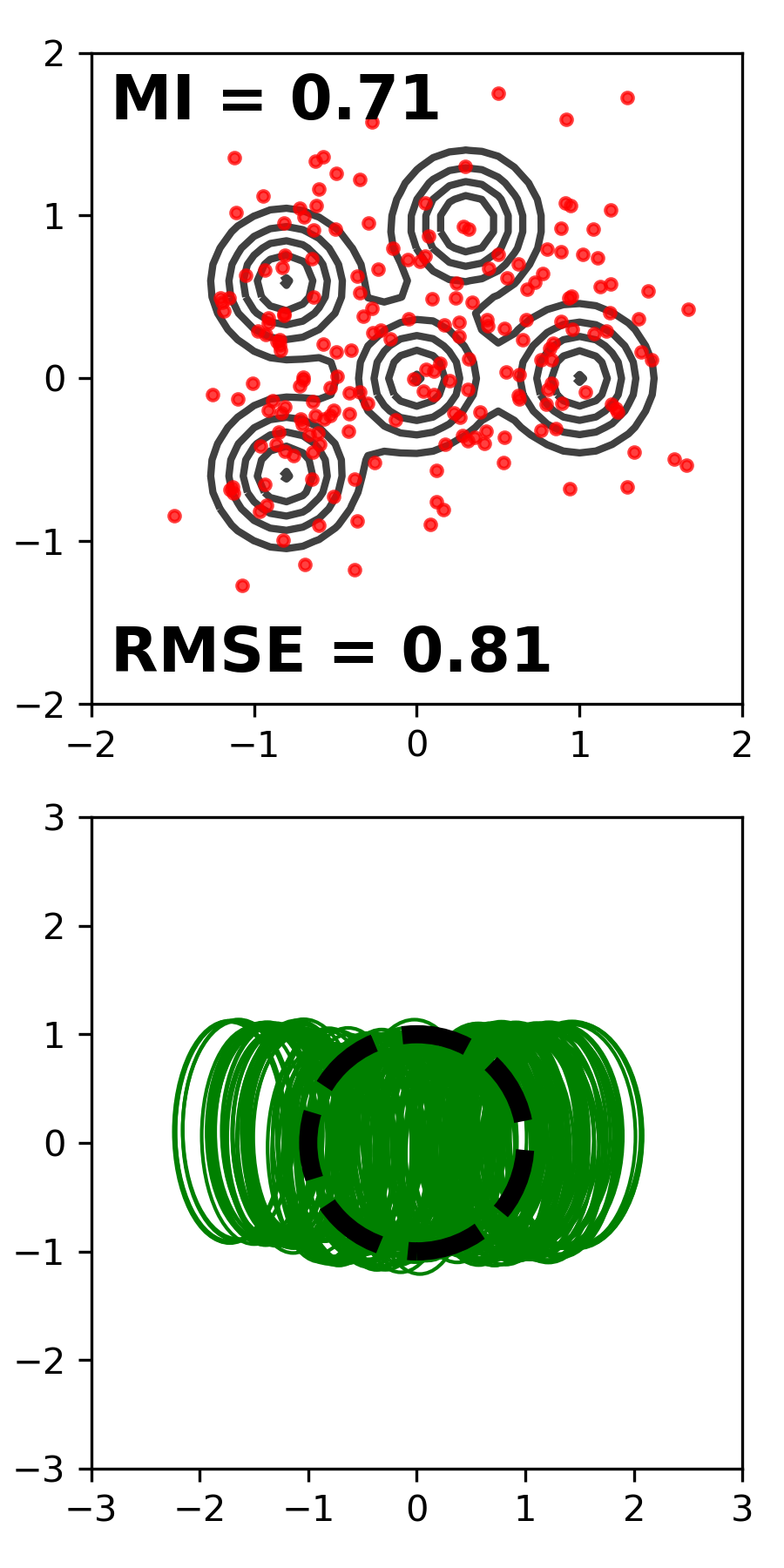}
    & \includegraphics[width=0.235\columnwidth]{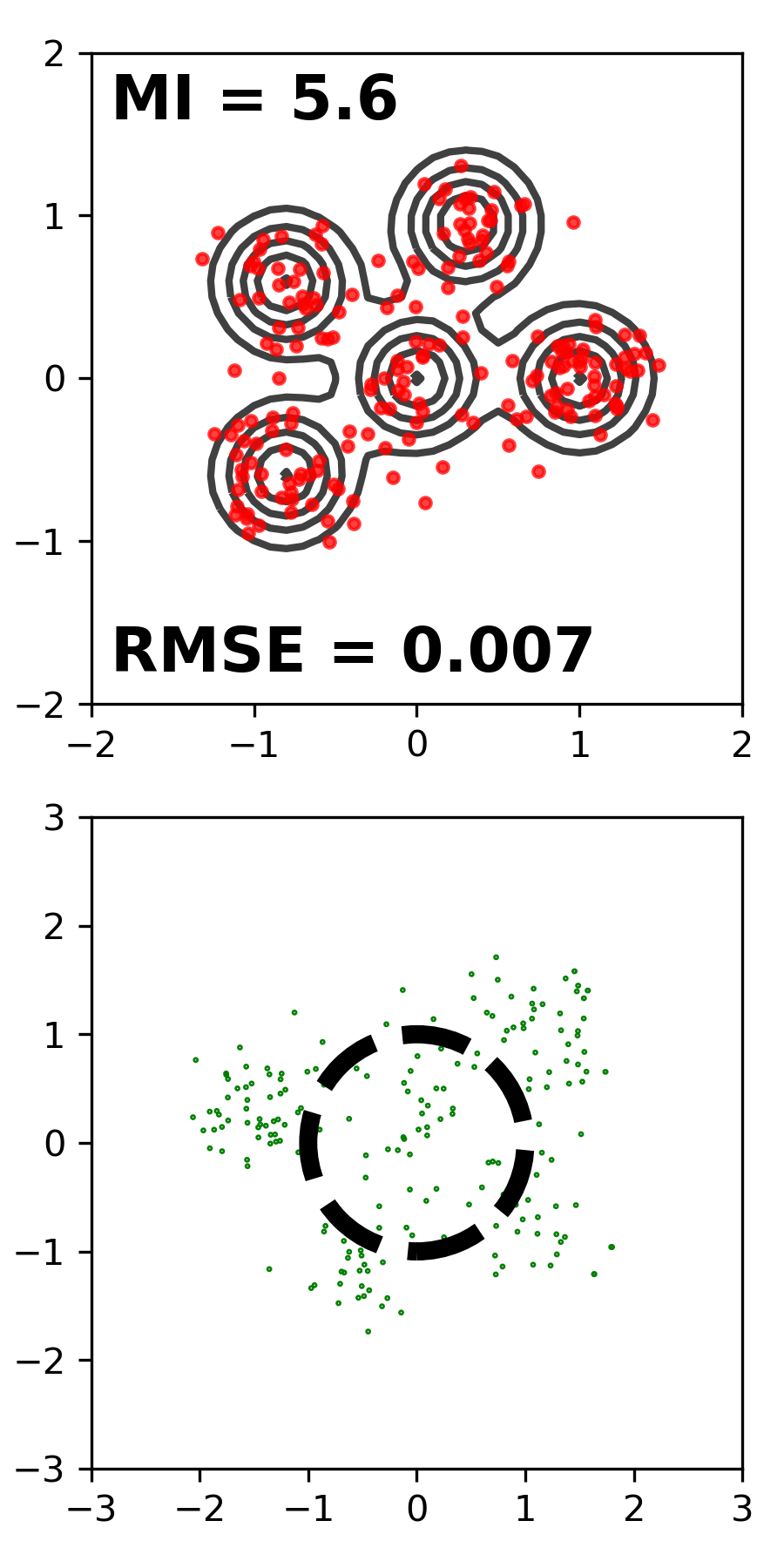}
    & &  \includegraphics[width=0.235\columnwidth]{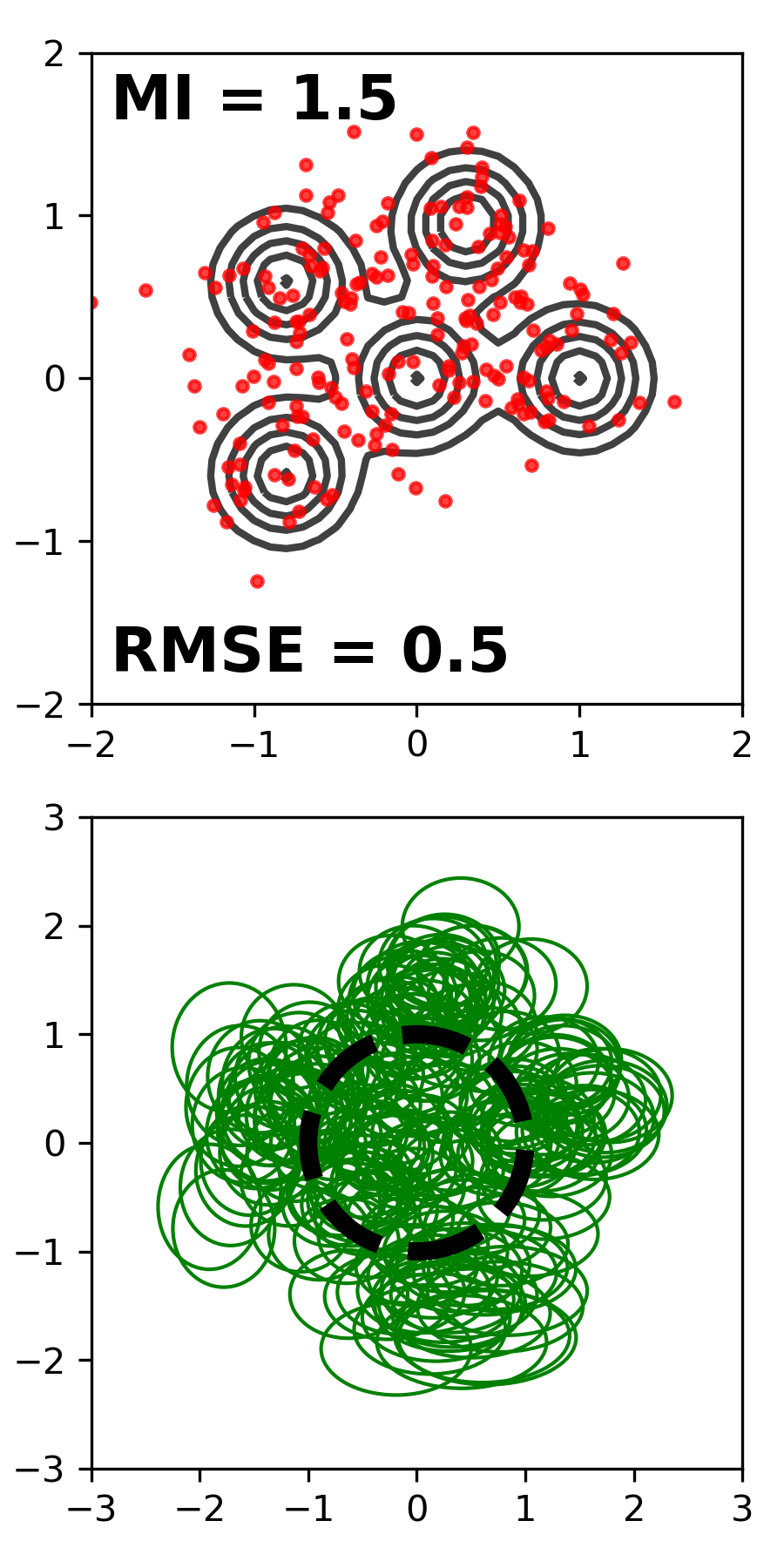}
    & \includegraphics[width=0.235\columnwidth]{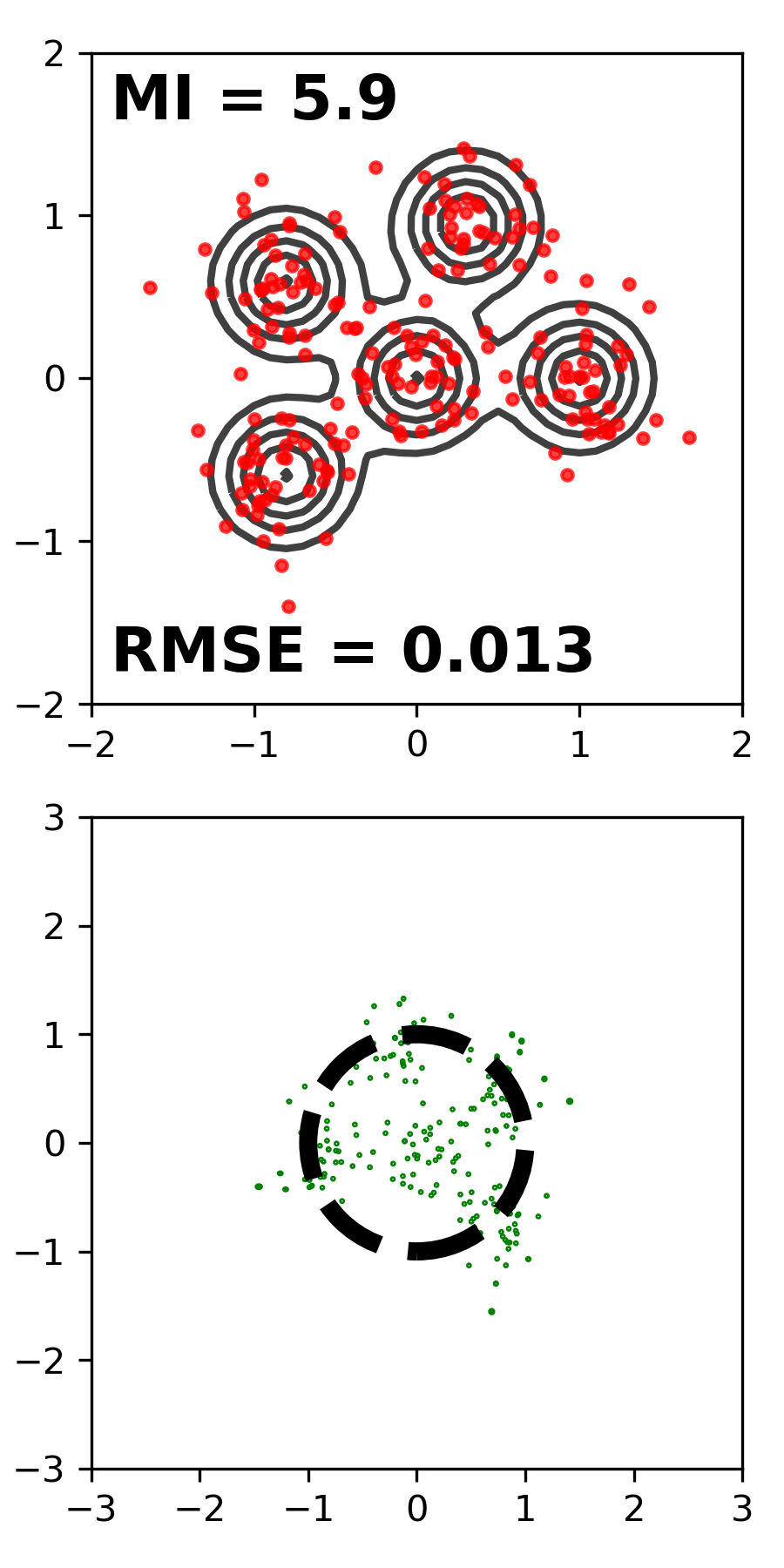}
    \\
    \multicolumn{2}{c}{\small (a) $h \in \mathbb{R}^{5}$ } & & \multicolumn{2}{c}{\small (b) $h \in \mathbb{R}^{20}$ } & & \multicolumn{2}{c}{\small (c) $h \in \mathbb{R}^{500}$ } \\
    \end{tabular}
    \vspace*{-0.25cm}
    \caption{
    VAE and MIM models with 2D inputs, a 2D latent space, and 5, 20 and 500 hidden units. 
    Top row: Black contours depict level sets of $\pjoint(\x)$; red dots are reconstructed test points.
    Bottom row: Green contours are one standard deviation ellipses of $\Menc(\z|\x)$ for test 
    points. Dashed black circles depict one standard deviation of $\pjoint (\z)$.
    The VAE predictive variance remains high, regardless of model expressiveness,
    an indication of various degrees of posterior collapse, while MIM produces lower predictive 
    variance and lower reconstruction errors, consistent with high mutual information.
    \label{fig:posterior-collapse-qualitative}
    }
     \vspace*{-0.3cm}
\end{figure*}

As a simple demonstration, Fig.\ \ref{fig:posterior-collapse-qualitative} depicts models 
learned on synthetic 2D data, $\x \in \mathbb{R}^2$, with a 2D latent space, $\z \in \mathbb{R}^2$. 
In 2D one can easily visualize the model and measure quantitative properties 
of interest (\eg, mutual information). 
Here, $\pjoint(\x)$ is a Gaussian mixture model comprising five isotropic components with 
standard deviation 0.25 (Fig.\ \ref{fig:posterior-collapse-qualitative} top), and $\pjoint(\z)$ 
is an isotropic standard Normal distribution (Fig.\ \ref{fig:posterior-collapse-qualitative} bottom).
The encoder and decoder have two fully connected layers with {\em tanh} activations,
and output the mean and variance of Gaussian densities.
The parameterized prior, $\Menc(\x)$, is defined to be the marginal of the 
decoding distribution (see Section~\ref{sec:learning-implicit})
and the model prior $\Mdec(\z)$ is defined to be $\pjoint(\z)$.
That is, the the only model parameters are those of the encoder and decoder.
We can therefore learn with MIM and VAE models using the same architecture and parameterization.
The warm-up comprises 3 steps \citep{Vaswani2017}, with 10,000 samples per epoch.
(Please see the caption for figure details.)


\begin{figure}[t]
    \centering
    \setlength{\tabcolsep}{0pt}
    \begin{tabular}{*2{>{\centering\arraybackslash}m{0.4\columnwidth}}}
      \includegraphics[width=0.375\columnwidth]{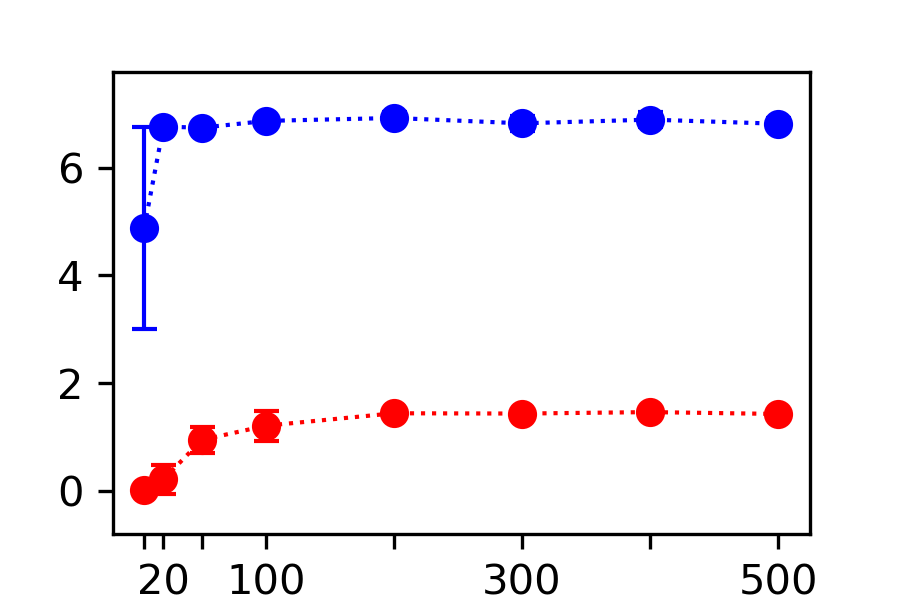}
    & \includegraphics[width=0.375\columnwidth]{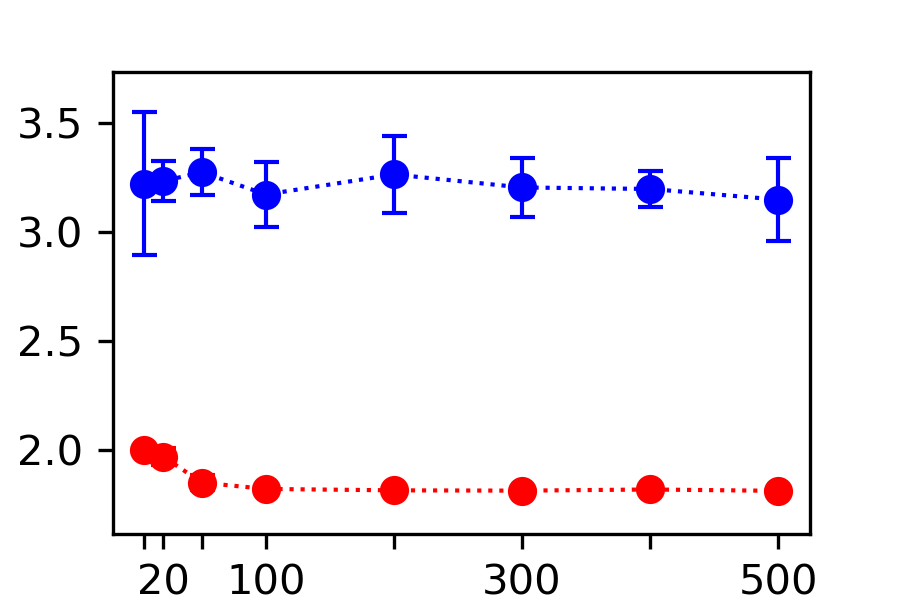} \\
    {\scriptsize (a) MI} & {\scriptsize (b) NLL}  \\
      \includegraphics[width=0.375\columnwidth]{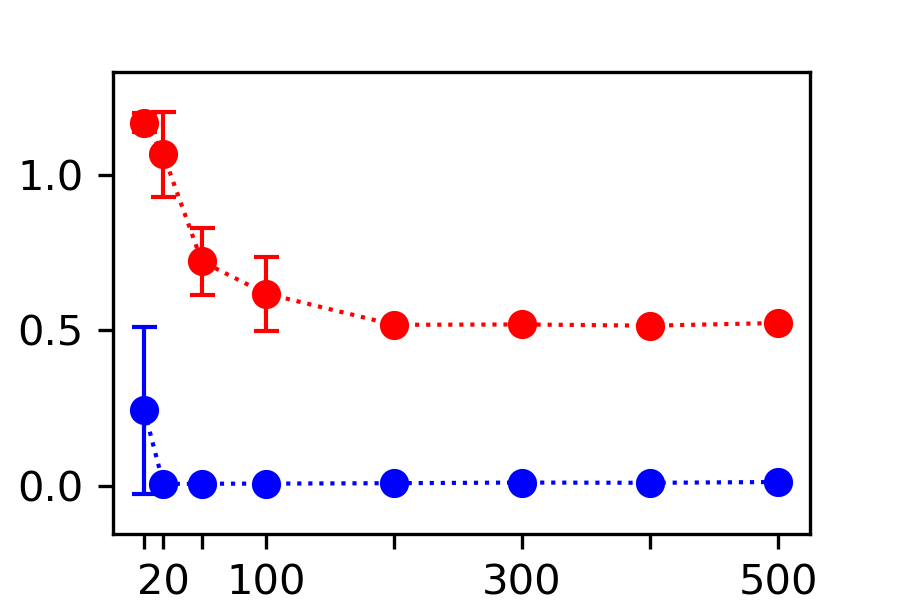}
    & \includegraphics[width=0.375\columnwidth]{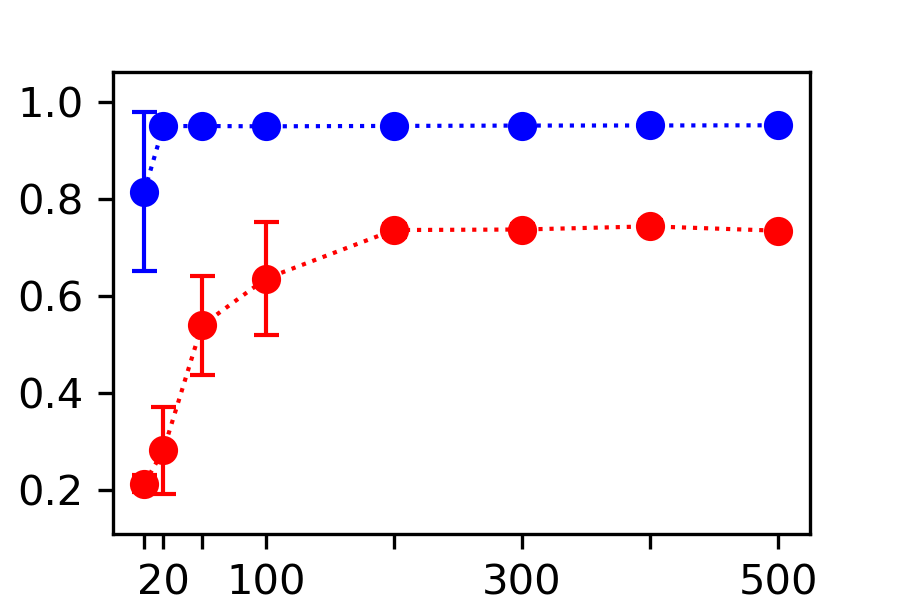} \\
    {\scriptsize (c) Reconstruction Error} & {\scriptsize (d) Classification (5-NN)}
    \end{tabular}
    \vspace*{-0.1cm}
    \caption{
    Test performance of MIM (blue) and VAE (red) for 2D GMM data 
    (cf.\ Fig.\ \ref{fig:posterior-collapse-qualitative}), versus  
    number of hidden units on x-axis. 
    Plots shows means and standard deviations for 10 experiments.
     \label{fig:posterior-collapse-quantitative}
    }
    \vspace*{-0.25cm}
\end{figure}

We train models with 5, 20 and 500 hidden units, and report the MI for the encoder $\Menc(\z \mid \x)$ and root-mean-squared test reconstruction error for each in the top row. 
(Following  \cite{Hjelm2018}, we estimate MI using the KSG estimator
\citep{PhysRevE.69.066138,DBLP:journals/corr/GaoOV16}, with 5-NNs.)
For the weakest architecture (a), with 5 hidden units, VAE and MIM posterior variances 
(green ellipses) are similar to the prior, a sign of posterior collapse, due to the limited
expressiveness of the encoder.
For more expressive architectures (b,c), the VAE posterior variance remains large,
preferring lower MI while matching the aggregated posterior to the prior.
The MIM encoder has tight posteriors, with higher MI and lower reconstruction errors. 
Indeed, one would expect low posterior variances since 
with a 2D to 2D problem, invertible mappings should be possible with a reasonably expressive model.

As functions of the number of hidden units, Fig.\ \ref{fig:posterior-collapse-quantitative} 
shows MI and average test negative log-likelihood (NLL) under the encoder, the test RMS 
reconstruction error, and 5-NN classification performance, predicting the GMM component 
responsible for each test point.
Performance on the auxiliary classification task is a proxy for representation quality.
Mutual information, reconstruction RMSE, and classification accuracy for test data 
are all as good or better with MIM, at the expense of poorer NLL.


\begin{figure}[t]
    \centering
    \setlength{\tabcolsep}{0pt}
    \begin{tabular}{*2{>{\centering\arraybackslash}m{0.4\columnwidth}}}
     \includegraphics[width=0.375\columnwidth]{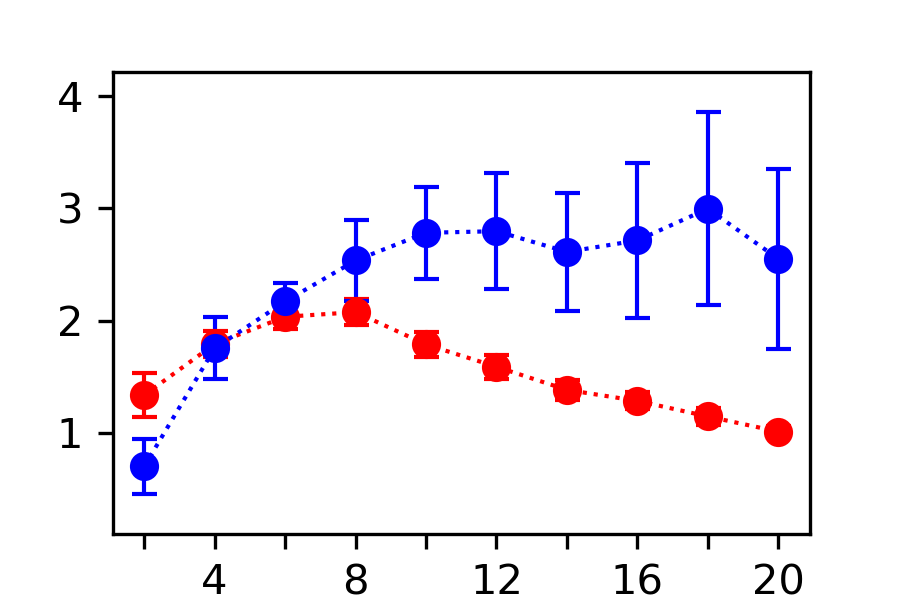}
    & \includegraphics[width=0.375\columnwidth]{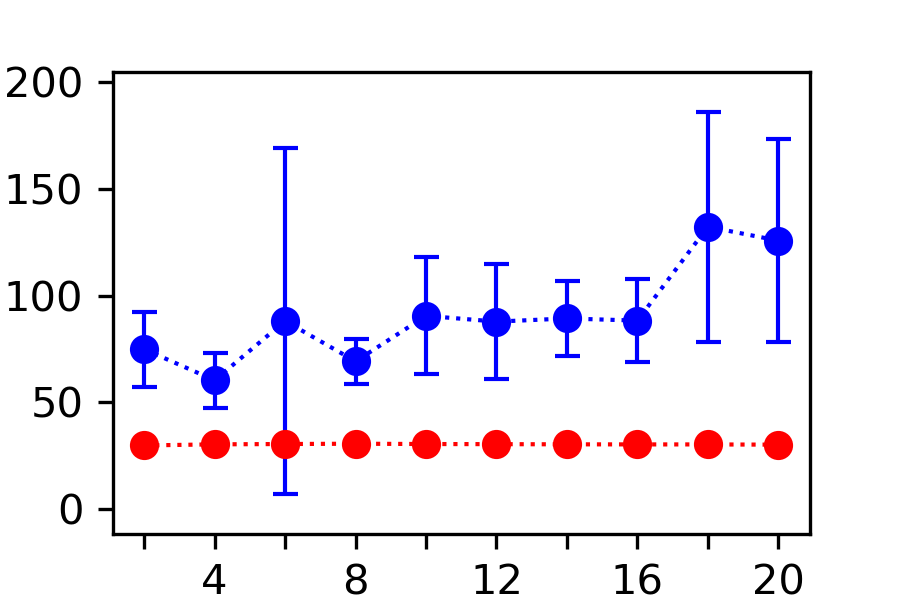} \\
     {\scriptsize (a) MI} & {\scriptsize (b) NLL}  \\
     \includegraphics[width=0.375\columnwidth]{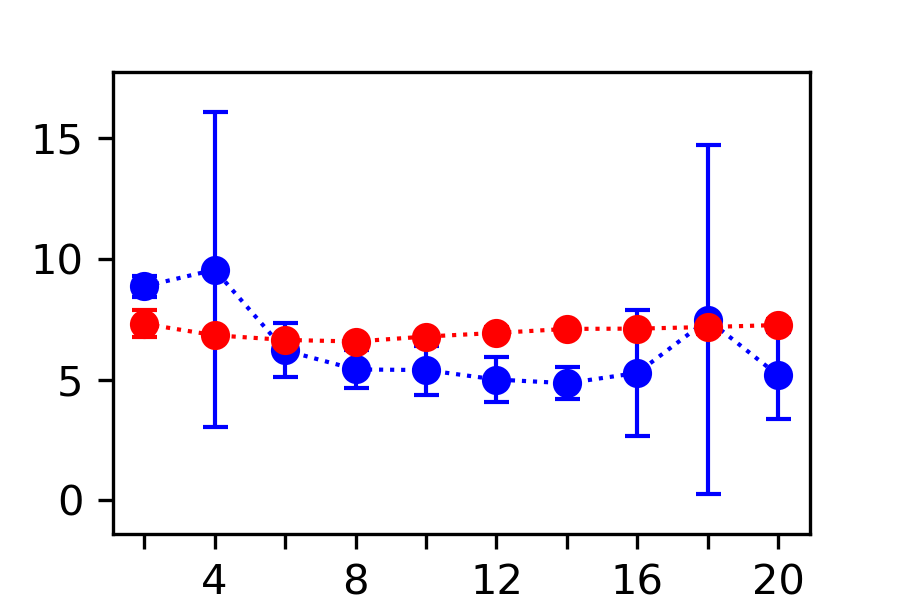}
     & \includegraphics[width=0.375\columnwidth]{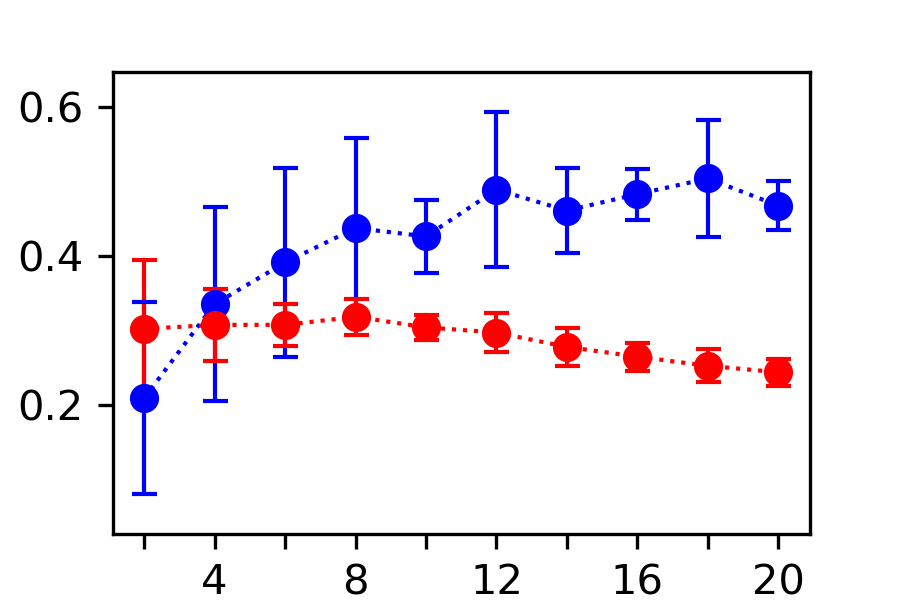} \\
     {\scriptsize (c) Reconstruction Error} & {\scriptsize (d) Classification (5-NN)}
    \end{tabular}
    \vspace*{-0.1cm}
    \caption{
    MIM (blue) and VAE (red) for 20D Fashion-MNIST, with latent dimension between 2 and 20.
    Plots depict mean and standard deviation of 10 experiments.
    \label{fig:mim-vs-vae-quantitative-bottleneck-fashion-mnist-pca}
    }
\end{figure}

{\em Low-dimensional Image Data: } 
We next consider data obtained by  projecting 784D  Fashion-MNIST images onto a 20D linear 
subspace (capturing 78.5\% of total variance), at which we can still reliably estimate MI.
The training and validation sets had 50,000 and 10,000 images.

Fig.\ \ref{fig:mim-vs-vae-quantitative-bottleneck-fashion-mnist-pca} plots results
as a function of the demension of the latent space $\z$.  MIM yields high MI and good 
classification accuracy at all but very low latent dimensions. MIM and VAE yield similar 
test reconstruction errors, with VAE having better NLL for test data.
Again VAE is prone to posterior collapse and low MI for more expressive models.
In contrast, MIM is robust to posterior collapse.  
This is consistent with Eqn.\ \eqref{eq:vi-as-ml-objective-text} and results 
of \citet{zhao2018LagrangeVAE} showing that many VAE variants designed to avoid 
posterior collapse are essentially encouraging higher MI.


\begin{figure*}[t]
    \centering
    \setlength{\tabcolsep}{0.5em} 
    {
    \renewcommand{\arraystretch}{1.2}
    \small
    \begin{tabular}{l||c|c||c|c||}
         \multicolumn{1}{l||}{} &  \multicolumn{2}{c||}{convHVAE (S)}  & \multicolumn{2}{c||}{convHVAE (VP)}   \\
         Dataset & {\small MIM} &  {\small VAE} & {\small MIM}  & {\small VAE}   \\
        \hline
        Fashion-MNIST & $272.14 \pm 0.64$ &  $\mathbf{225.40 \pm 0.05}$ & $227.61 \pm 0.34$ & $\mathbf{224.77 \pm 0.04}$  \\
        \multicolumn{1}{r||}{\scriptsize (A-MIM)} & ($263.04 \pm 0.8$) &  & ($225.53 \pm 0.11$) &   \\ \hline
        MNIST         & $126.85 \pm 0.56$ &  $\mathbf{80.50 \pm 0.05}$ & $82.73 \pm 0.08$ & $\mathbf{79.66 \pm 0.06}$   \\
        \multicolumn{1}{r||}{\scriptsize (A-MIM)} & ($122.67 \pm 0.84$) &  & ($80.4 \pm 0.84$) &   \\ \hline
        Omniglot      & $141.81 \pm 0.32$ &  $\mathbf{97.94 \pm 0.29}$ & $104.10 \pm 2.17$ & $\mathbf{97.52 \pm 0.16}$   \\
        \multicolumn{1}{r||}{\scriptsize (A-MIM)} & ($141.03 \pm 0.3$) &  & ($102.57 \pm 1.72$) &   \\
        \hline
        \hline
         \multicolumn{1}{l||}{} &  \multicolumn{2}{c||}{PixelHVAE (S)}  & \multicolumn{2}{c||}{PixelHVAE (VP)} \\
             &  {\small A-MIM} & {\small VAE} & {\small A-MIM} & {\small VAE}  \\
        \hline
        Fashion-MNIST  & $243.95 \pm 0.47$ & $\mathbf{224.65 \pm 0.07}$ & $224.94 \pm 0.34$ & $\mathbf{224.02 \pm 0.08}$ \\
        MNIST          & $114.96 \pm 0.35$ & $\mathbf{79.04 \pm 0.05}$ & $79.04 \pm 0.08$ & $\mathbf{78.60 \pm 0.04}$ \\
        Omniglot       & $126.12 \pm 0.38$ & $\mathbf{91.06 \pm 0.14}$ & $91.82 \pm 0.20$ & $\mathbf{90.74 \pm 0.15}$ \\
    \end{tabular}
    }
    \caption{
    Average test NLL (in nats) based on 10 trials per condition.
    With a more powerful prior, MIM and VAE yield comparable results.
    \label{tab:mim-vs-vae-image-NLL}
    }
    \vspace*{-0.3cm}
\end{figure*}

\begin{figure*}[t]
    \centering
    \small
    \begin{subfigure}[b]{0.23\textwidth}
    \setlength{\tabcolsep}{0pt}
    \begin{tabular}{*2{>{\centering\arraybackslash}m{0.5\textwidth}}}
    \multicolumn{2}{c}{\includegraphics[trim={10.2cm 0 0 0},clip,width=1.0\textwidth]{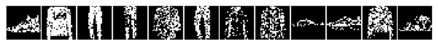}} 
    \\[-0.3cm]
    \multicolumn{2}{c}{\includegraphics[trim={10.2cm 0 0 0},clip,width=1.0\textwidth]{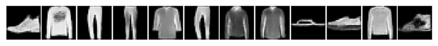}} 
    \\[-0.3cm]
    \multicolumn{2}{c}{\includegraphics[trim={10.2cm 0 0 0},clip,width=1.0\textwidth]{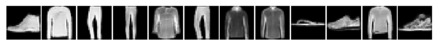}} 
    \vspace*{-0.2cm}
    \\
    \hspace{-0.04\textwidth}\includegraphics[trim={0.2cm 10.1cm 10.2cm 0},clip,width=0.48\textwidth]{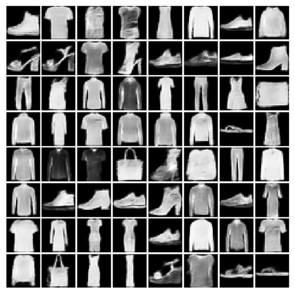}
    & \includegraphics[trim={10.2cm 10.1cm 0.2cm 0},clip,width=0.48\textwidth]{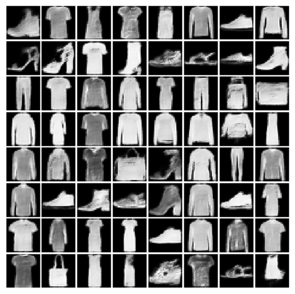}
    \\
    {\scriptsize VAE}  & {\scriptsize A-MIM}
    \end{tabular}
    \label{fig:mim-vs-vae-image-qualitative-fashion-mnist}
    \end{subfigure}
    \begin{subfigure}[b]{0.23\textwidth}
    \setlength{\tabcolsep}{0pt}
    \begin{tabular}{*2{>{\centering\arraybackslash}m{0.5\textwidth}}}
    \multicolumn{2}{c}{\includegraphics[trim={10.2cm 0 0 0},clip,width=1.0\textwidth]{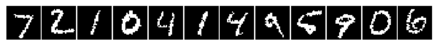}} 
    \\[-0.3cm]
    \multicolumn{2}{c}{\includegraphics[trim={10.2cm 0 0 0},clip,width=1.0\textwidth]{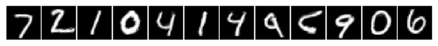}} 
    \\[-0.3cm]
    \multicolumn{2}{c}{\includegraphics[trim={10.2cm 0 0 0},clip,width=1.0\textwidth]{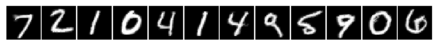}} 
    \vspace*{-0.2cm}
    \\
    \hspace{-0.04\textwidth}\includegraphics[trim={0.2cm 10.1cm 10.2cm 0},clip,width=0.48\textwidth]{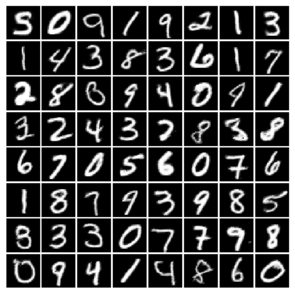}
    & \includegraphics[trim={10.2cm 10.1cm 0.2cm 0},clip,width=0.48\textwidth]{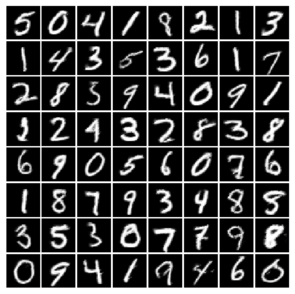}
    \\
        {\scriptsize VAE}  & {\scriptsize A-MIM}
    \end{tabular}
    \label{fig:mim-vs-vae-image-qualitative-mnist}
    \end{subfigure}
    \begin{subfigure}[b]{0.23\textwidth}
    \setlength{\tabcolsep}{0pt}
    \begin{tabular}{*2{>{\centering\arraybackslash}m{0.5\textwidth}}}
    \multicolumn{2}{c}{\includegraphics[trim={10.2cm 0 0 0},clip,width=1.0\textwidth]{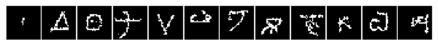}} 
    \\[-0.3cm]
    \multicolumn{2}{c}{\includegraphics[trim={10.2cm 0 0 0},clip,width=1.0\textwidth]{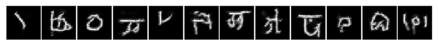}} 
    \\[-0.3cm]
    \multicolumn{2}{c}{\includegraphics[trim={10.2cm 0 0 0},clip,width=1.0\textwidth]{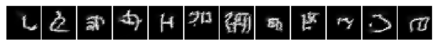}} 
    \vspace*{-0.2cm}
    \\
    \hspace{-0.04\textwidth}\includegraphics[trim={0.2cm 10.1cm 10.2cm 0},clip,width=0.48\textwidth]{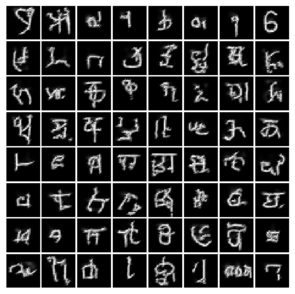}
    & \includegraphics[trim={10.2cm 10.1cm 0.2cm 0},clip,width=0.48\textwidth]{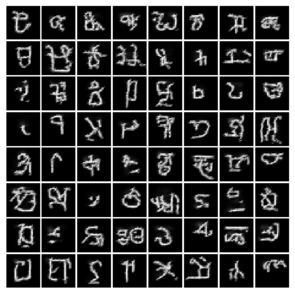}
    \\
     {\scriptsize VAE}  & {\scriptsize A-MIM}
    \end{tabular}
    \label{fig:mim-vs-vae-image-qualitative-omniglot}
    \end{subfigure}
    \vspace*{-0.3cm}
    \caption{
    MIM and VAE learning with the PixelHVAE (VP) architecture, applied to 
    Fashion-MNIST, MNIST, and Omniglot (left to right).
    The top three rows (from top to bottom)  are test data samples, 
    VAE reconstruction, and MIM reconstruction.
    Bottom: random samples from VAE and MIM.
    \label{fig:mim-vs-vae-image-qualitative}
    }
    \vspace*{-0.3cm}
\end{figure*}

\vspace{-0.1cm}
\subsection{MIM Learning with Image Data}
\label{sec:high-dimensional-image-data}
\vspace{-0.1cm}

We next consider MIM and VAE learning with three image datasets, namely, Fashion-MNIST, MNIST, and Omniglot.
Unfortunately, with high dimensional data we cannot reliably compute MI \citep{Hjelm2018}. 
Instead, for model assessment we focus on NLL, reconstruction errors, and the quality of random samples. 
In doing so we also explore multiple, more expressive architectures, including the top performing models 
from \cite{DBLP:journals/corr/TomczakW17}, namely, {\em convHVAE} (L = 2) and {\em PixelHVAE} (L = 2), 
with Standard (S) priors\footnote{$\Mdec(\z) = \mathcal{N}(\z ; \mu = \bf{0}, \sigma =  \mathbb{I})$, 
a standard Normal distribution, where $\mathbb{I}$ is the identity matrix.}, 
and {\em VampPrior} (VP) priors\footnote{$\Mdec(\z) = \frac{1}{K}\sum_{k=1}^{K} \Menc(\z|\bs{u}_k)$, 
a mixture model of the encoder posterior with optimized pseudo-inputs $\bs{u}_k$.}. 
The VP pseudo-inputs are initialized with training data samples. 
For all experiments we use the experimental setup described by \citet{DBLP:journals/corr/TomczakW17}, 
and the same latent dimensionality, \ie, $\z \in \mathbb{R}^{80}$. 
Finally, in the case of auto-regressive decoders (\eg, PixelHVAE), because sampling is 
extremely slow, we use A-MIM learning instead of MIM.

Table\ \ref{tab:mim-vs-vae-image-NLL} reports NLLs on test data.
As above, one can see that VAEs tend to yield better NLL scores. 
However, the gap becomes narrower for the more expressive PixelHVAE (S) and
largely closes for the most expressive PixelHVAE (VP). 
As the model capacity increases, we have found thatMIM tends to
better utilize the additional modelling power.

We also show qualitative results for the more expressive PixelHVAE (VP) models.
Fig.\   \ref{fig:mim-vs-vae-image-qualitative} depicts reconstructions.
Here, following \citet{DBLP:journals/corr/TomczakW17}, the random test samples
on the top row depict binary images, while the corresponding reconstructions 
depict the probability of each pixel being 1.
Examples are provided for Fashion-MNIST, MNIST, and Omniglot, indicating that 
MIM and VAE are comparable in the quality of reconstruction.
See supplementary materials for additional results.

The poor NLL and hence poor sampling for MIM when specified with a weak prior model 
can be explained by the tightly clustered latent representation that MIM often learns
(\eg, Fig.\ \ref{fig:posterior-collapse-qualitative}). 
A more expressive, parameterized prior such as the VampPrior
can capture such clusters more accurately, and produce better samples.
To summarize, while VAE opts for better NLL and sampling, at the expense of lower 
mutual information, MIM provides higher mutual information at the expense of the NLL 
for a weak prior, and comparable NLL and sampling with more expressive priors.

\begin{figure*}[t]
    \centering
    \setlength{\tabcolsep}{0.5em} 
    {
    \renewcommand{\arraystretch}{1.2}
    \small
    \begin{tabular}{l||c|c||c|c||c|c||c|c||}
         \multicolumn{1}{l||}{} &  \multicolumn{2}{c||}{convHVAE (S)}  & \multicolumn{2}{c||}{convHVAE (VP)} & \multicolumn{2}{c||}{PixelHVAE (S)}  & \multicolumn{2}{c||}{PixelHVAE (VP)} \\
         Dataset & {\small MIM (A-MIM)}  & {\small VAE} & {\small MIM (A-MIM)}  & {\small VAE} & {\small A-MIM} & {\small VAE} & {\small A-MIM} & {\small VAE} \\
        \hline
        Fashion-MNIST
        & $\mathbf{0.83} ~ (0.81)$  & $0.76$ & $0.81 ~ (\mathbf{0.82})$  & $0.78$ & $0.71$ & $\mathbf{0.76}$ & $\mathbf{0.79}$ & $0.77$ \\
        MNIST
        & $\mathbf{0.97} ~ (0.96)$  & $0.92$  & $\mathbf{0.97} ~ (0.96)$  & $0.92$ & $\mathbf{0.95}$ & $0.86$ & $\mathbf{0.96}$ & $0.81$ \\
    \end{tabular}
    }
    \caption{
    Test accuracy of 5-NN classifier.
    Standard deviations are well less than 0.001, and omitted from the table.
     \label{tab:mim-vs-vae-classification}
    }
    \vspace*{-0.2cm}
\end{figure*}

\vspace{-0.2cm}
\subsection{Clustering and Classification} 
\label{sec:representation-learning-with-mim}
\vspace{-0.1cm}

Finally, following \cite{hjelm2018learning}, we consider an auxiliary classification 
task as a further measure of the quality of the learned representations. 
We use k-nearest-neighbor (K-NN) classification as it is simple, non-parametric method that 
relies on a suitable arrangement of the latent space.
We experimented with different neighborhood sizes of 1,3, 5 and 10, but all gave comparable results.
We therefore applied a 5-NN classifier to the representations
learned in Sec.\ \ref{sec:high-dimensional-image-data} on test samples from MNIST and Fashion-MNIST.

\begin{figure}[t]
    \centering
    \setlength{\tabcolsep}{0pt}
    \small
    \begin{tabular}{cc}
     {\scriptsize VAE}  & {\scriptsize MIM} \\
        \includegraphics[width=0.4\columnwidth]{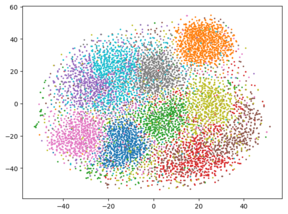}
    & \includegraphics[width=0.4\columnwidth]{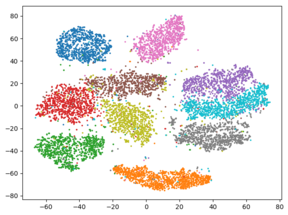}
    \\
      \includegraphics[width=0.4\columnwidth]{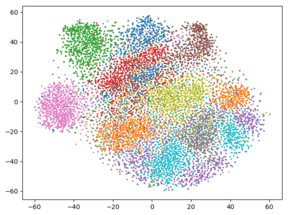}
       & \includegraphics[width=0.4\columnwidth]{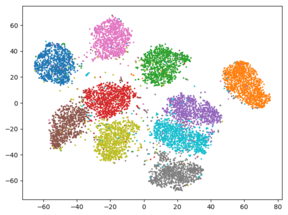}
    \end{tabular}
    \vspace*{-0.2cm}
    \caption{
        VAE and MIM embeddings for MNIST for (top) convHVAE (VP) and (bottom) PixelHVAE (VP).
          \label{fig:mim-vs-vae-embed-MNIST}
    }
  
    \vspace*{-0.25cm}
\end{figure}

\begin{figure}[t]
    \centering
    \setlength{\tabcolsep}{0pt}
    \small
    \begin{tabular}{cc}
     {\scriptsize VAE}  & {\scriptsize MIM} \\
        \includegraphics[width=0.4\columnwidth]{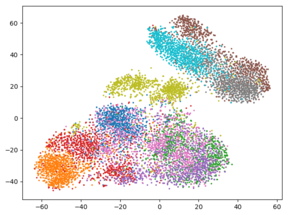}
    & \includegraphics[width=0.4\columnwidth]{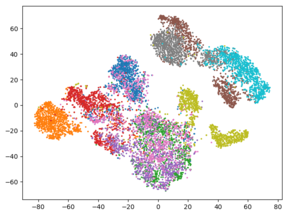}
    \\
       \includegraphics[width=0.4\columnwidth]{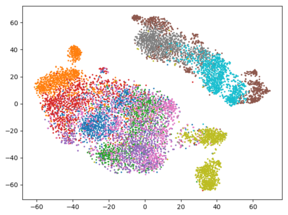}
      & \includegraphics[width=0.4\columnwidth]{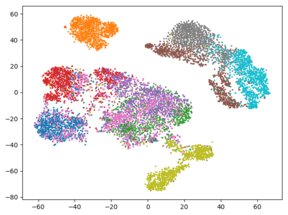}
    \end{tabular}
    \vspace*{-0.2cm}
    \caption{
    VAE and MIM embeddings for Fashiuon MNIST for (top) convHVAE (VP) and (bottom) PixelHVAE (VP).
    \label{fig:mim-vs-vae-embed-FMNIST}
    }
    \vspace*{-0.25cm}
\end{figure}
Table\ \ref{tab:mim-vs-vae-classification} shows that MIM yields more 
accurate classification results in all but one case.
We attribute the performance difference to higher mutual information of MIM representations. 
Figures\ \ref{fig:mim-vs-vae-embed-MNIST} and \ref{fig:mim-vs-vae-embed-FMNIST} provide
qualitative visualizations of the latent clustering, for which t-SNE \citep{maaten2008visualizing}) 
was used to project the latent space down to 2D. 
One can see that MIM learning tends to cluster classes in the latent representation 
more tightly, while VAE clusters are more diffuse and overlapping, consistent with 
the results in Table\ \ref{tab:mim-vs-vae-classification}.

\section{Related Work}
\label{sec:related-work}

\citet{zhao2018LagrangeVAE} establishes a unifying objective that describes many common VAE variants.
They show that these variants optimize the Lagrangian dual of a primal optimization problem involving
mutual information, subject to constraints on a chosen divergence. They use this to directly
optimize the Lagrange multipliers and explore the pareto front of mutual information maximization subject
to divergence constraints. The trade-off is that this requires a number of heuristics, such as bounding
mutual information and estimating the true divergence. Both are difficult in practice~\cite{Hjelm2018}.
As with most variants, we consider the relaxed objective and fix the weightings, however by considering
joint entropy instead of mutual information directly, we step slightly outside this framework. This
allows us to produce a tractable objective that can be solved with more direct optimization approaches.

Symmetry has been considered in the literature before to equally emphasize both the encoding
and decoding distributions. \citet{pu2017adversarial} show that an asymmetric objective can produce
overly broad marginals, and suggest using a symmetric KL divergence. 
\cite{dumoulin2016adversarially, Donahue2016BiGAN}
propose an encoder for GAN~\citep{goodfellow2014generative} models and train it with JSD. In each of these
cases, the symmetric divergence necessitates adversarial training. We show that with the right combination
of objective terms, it is possible to both target a symmetric objective and still get tractable optimization.

Mutual information, together with disentanglement, is considered to be a 
cornerstone for useful representations \citep{Hjelm2018,hjelm2018learning}. 
Normalizing flows \citep{Rezende2015,Dinh2014,Dinh2016a,Kingma2018,DBLP:journals/corr/abs-1902-00275} 
directly maximizes mutual information by restricting the architecture to be invertible and tractable. 
This, however, requires the latent dimension to be the same as the dimension of the observations 
(\ie, no bottleneck).
As a consequence, normalizing flows are not well suited to 
learning a concise representation of high dimensional data (\eg, images). 
Here, MIM often yields mappings that are approximately invertible, with high mutual 
information and low reconstruction errors.  

\citet{DBLP:journals/corr/BornscheinSFB15} share some of the same design principles as MIM, 
\ie, symmetry and encoder/decoder consistency. 
However, their formulation models the joint density
in terms of the geometric mean between the encoder and decoder, for which one must
compute an expensive partition function. \cite{pu2017adversarial} focuses on minimizing symmetric KL,
but must use an adversarial learning procedure, while MIM can be minimized directly.

\section{Conclusions} \label{sec:conclusion}

We introduce a new representation learning framework, named the {\em mutual information machine} (MIM), that defines a latent generative model built around three principles: consistency between the encoding and decoding distributions, high mutual information between the observed and latent variables, and low marginal entropy. This yields a learning objective that can be directly optimized by stochastic gradient descent with reparameterization, as opposed to adversarial learning used in related frameworks. We show that the MIM framework produces highly informative representations and is robust against posterior collapse. Most importantly, MIM greatly benefits from additional capacity; as the encoder and decoder become larger and more expressive, MIM retains its highly informative representation, and consistently shows improvement in its modelling of the underlying distribution as measured by negative log-likelihood. The stability, robustness, and ease of training make MIM a compelling framework for latent variable modelling across a variety of domains.

\bibliography{paper}
\bibliographystyle{icml2020}


\newpage
\onecolumn
\appendix
\section{Code}
Code and data are available from
{
\url{https://research.seraphlabs.ca/projects/mim/index.html}
}

\section{Derivations for MIM Formulation}

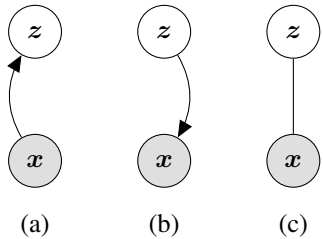
\begin{figure}[H]
    \centering
    \begin{tikzpicture}


\node[latent]                (zenc) {$\z$};
\node[obs, below=of zenc]                (xenc) {$\x$};

\node[latent, right=of zenc]               (zdec) {$\z$};
\node[obs, below=of zdec]                (xdec) {$\x$};

\node[latent, right=of zdec]             (z) {$\z$};
\node[obs, below=of z]                   (x) {$\x$};

 \node[const, below=of xenc, yshift=0.65cm]  {(a)} ; %
 \node[const, below=of xdec, yshift=0.65cm]  {(b)} ; %
 \node[const, below=of x, yshift=0.65cm]  {(c)} ; %

\edge [-] {x} {z} ; %
\edge [bend left] {xenc} {zenc} ; %
\edge [bend left] {zdec} {xdec} ; %

\end{tikzpicture}
    \caption{\MIM learning estimates two factorizations of a joint distribution:
    (a) encoding; (b) decoding factorizations.
    (c) The estimated joint distribution.
    }
    \label{fig:mim-model}
\end{figure}

In what follows we provide detailed derivations of key elements of the formulation in
the paper, namely, Equations \eqref{eq:jsd-h},  and \eqref{eq:MIM-parts}.
We also consider the relation between MIM based on the Jensen-Shannon divergence and the
symmetric KL divergence.


\subsection{Regularized JSD and Entropy}
\label{sec:jsd-entropy-supp}

First we develop the relation in Equation\ \eqref{eq:jsd-h}, between
Jensen-Shannon divergence of the encoder and decoder, the average joint entropy of the
encoder and decoder, and the joint entropy of the mixture distribution $\Msamp$.

The Jensen-Shannon divergence with respect to the encoding distribution $\Penc$
and the decoding distribution $\Pdec$ is defined as
\begin{align*}
    \mathrm{JSD}(\params) &= \frac{1}{2}\left (\DKL{\Pdec}{\Msamp} +  \DKL{\Penc}{\Msamp} \right ) \\
    &= \frac{1}{2} \bigg ( \CE{\Pdec}{\Msamp} - \HD{\Pdec}{\x, \z} \\
    &  ~~~~~ + \CE{\Penc}{\Msamp} - \HD{\Penc}{\x, \z} \bigg )
\end{align*}
Where $\Msamp=\frac{1}{2}(\Pdec + \Penc)$ is a mixture of the encoding and decoding distributions.
Adding on the regularizer $\RH(\params) = \frac{1}{2}(\HD{\Pdec}{\x, \z} + \HD{\Penc}{\x, \z})$ gives
\begin{align*}
    \mathrm{JSD}(\params) + \RH(\params) &= ~\frac{1}{2} \left ( \CE{\Pdec}{\Msamp} + \CE{\Penc}{\Msamp} \right ) \\
    &= ~\HD{\Msamp}{\x, \z}
\end{align*}


\subsection{MIM Consistency}
\label{sec:mim-consistency}

Here we discuss in greater detail how the learning algorithm encourages consistency
between the encoder and decoder of a \MIM model, beyond the fact that they are
fit to the same sample distribution. To this end we expand on several properties of the model and the optimization procedure.

\subsubsection{MIM consistency regularizer}

In what follows we derive the form of the MIM consistency regularizer,  $ \RMIM(\params)$,
given in Equation\ \eqref{eq:RMIM-DKL}.
Recall that we define $\Mmodel=\frac{1}{2}(\Pdecoderjoint + \Pencoderjoint)$.
We can show that $\MIMloss$ is equivalent to $\CEloss$ plus a regularizer by taking their difference.
\begin{align*}
    \RMIM(\params) &= \MIMloss(\params) - \CEloss(\params) \\
    &= \frac{1}{2}(\CE{\Msamp}{\Pdecoderjoint} + \CE{\Msamp}{\Pencoderjoint}) - \CE{\Msamp}{\Mmodel} \\
    &= \frac{1}{2}(\DKL{\Msamp}{\Pdecoderjoint} + \HD{\Msamp}{\x, \z} + \DKL{\Msamp}{\Pencoderjoint} + \HD{\Msamp}{\x, \z}) \\
    & ~~~~~  - \DKL{\Msamp}{\Mmodel} - \HD{\Msamp}{\x, \z} \\
    &=  \frac{1}{2}(\DKL{\Msamp}{\Pdecoderjoint} + \DKL{\Msamp}{\Pencoderjoint}) - \DKL{\Msamp}{\Mmodel}
\end{align*}
where $\RMIM(\params)$ is non-negative, and is zero only when the encoding and decoding distributions are consistent
(\ie, they represent the same joint distribution). To prove that $\RMIM(\params) \ge 0$ we now construct Equation\ \eqref{eq:RMIM-DKL}
in terms of expectation over a joint distribution, which yields
\begin{align*}
    \RMIM(\params) &= \frac{1}{2}(\CE{\Msamp}{\Pdecoderjoint} + \CE{\Msamp}{\Pencoderjoint}) - \CE{\Msamp}{\Mmodel} \\
    &= ~ \E{\x,\z \sim \Msamp}{-\frac{1}{2} \log \Pdecoderjoint - \frac{1}{2} \log \Pencoderjoint + \log \frac{1}{2} (\Menc \left(\x, \z \right) + \Mdec \left(\x, \z \right))} \\
    &= ~ \E{\x,\z \sim \Msamp}{- \log \sqrt{\Menc \left(\x, \z \right) \cdot \Mdec \left(\x, \z \right)} + \log \frac{1}{2} (\Menc \left(\x, \z \right) + \Mdec \left(\x, \z \right))} \\
    &= ~ \E{\x,\z \sim \Msamp}{-\log \, \frac{\sqrt{\Menc \left(\x, \z \right) \cdot \Mdec \left(\x, \z \right)}}{\frac{1}{2} (\Menc \left(\x, \z \right) + \Mdec \left(\x, \z \right))} } ~\ge~ 0
\end{align*}
where the inequality follows Jensen's inequality, and equality holds only when $\Menc \left(\x, \z \right) = \Mdec \left(\x, \z \right)$ (\ie,
encoding and decoding distributions are consistent).


\subsubsection{Self-Correcting Gradient}

One important property of the optimization follows directly from the difference
between the gradient of the upper bound $\MIMloss$
and the gradient of the cross-entropy loss $\CEloss$.
By moving the gradient operator into the expectation  using reparametrization, one
can express the gradient of $\MIMloss(\params)$ in terms of the gradient of
$\log \Mmodel$ and the regularization term in Equation\ \eqref{eq:RMIM-DKL}.
That is, with some manipulation one obtains
\begin{eqnarray}
\frac{\partial}{\partial\params} \left( \frac{\log\Menc+\log\Mdec}{2} \right) \, = \,
\frac{\partial}{\partial\params}\log\left(\! \frac{\Menc+\Mdec}{2}\! \right)
+ \,\frac{1}{2}\frac{\left(\frac{\Mdec}{\Menc}-1\right)\frac{\partial}{\partial\params}\Menc+\left(\frac{\Menc}{\Mdec}-1\right)\frac{\partial}{\partial\params}\Mdec}{\Menc+\Mdec} \, ,
\label{eq:tzk-gradient-correction}
\end{eqnarray}
which shows that for any data point where a gap $\Menc > \Mdec$ exists, the gradient applied
to $\Mdec$ grows with the gap, while placing correspondingly less weight on the gradient applied
to $\Menc$.
The opposite is true when $\Menc < \Mdec$.
In both case this behaviour encourages consistency between the encoder and decoder.
Empirically, we find that the encoder and decoder become reasonably consistent early in the optimization process.

\subsubsection{Numerical Stability}

Instead of optimizing an upper bound $\MIMloss$, one might consider a direct optimization of $\CEloss$.
Earlier we discussed the importance of the consistency regularizer in $\MIMloss$.
Here we motivate the use of $\MIMloss$ from a numerical perspective point of view.
In order to optimize $\CEloss$ directly, one must convert $\log \Menc$ and $\log \Mdec$ to $\Menc$ and $\Mdec$.
Unfortunately, this is has the potential to produce numerical errors,
especially with 32-bit floating-point precision on GPUs.
While various tricks can reduce numerical instability, we find that using the
upper bound eliminates the problem while providing the additional benefits outlined above.

\subsubsection{Tractability}

As mentioned earlier, there may be several ways to combine the encoder and decoder into
a single probabilistic model.
One possibility we considered, as an alternative to $\Mmodel$
in Equation \eqref{eq:LCE}, is
\begin{equation}
\Mmodel = \frac{1}{\beta} \sqrt{ \Menc \,\Mdec } ~,
\label{eq:tzk-encoder-decoder-alt}
\end{equation}
where $\beta = \int \sqrt{ \Menc \,\Mdec } \, d\x\, d\z$
is the partition function, similar to \cite{DBLP:journals/corr/BornscheinSFB15}.
One could then define the objective to be the cross-entropy as above with a
regularizer to encourage $\beta$ to be close to 1, and hence to encourage consistency between
the encoder and decoder.
This, however, requires a good approximation to the partition function.
Our choice of $\Mmodel$ avoids the need for a good value approximation by
using reparameterization, which results in unbiased low-variance gradient,
independent of the accuracy of the approximation of the value.


\subsection{MIM Loss Decomposition}
\label{sec:mim-loss-decomposition}

Here we show how to break down the $\MIMloss$ into the set of intuitive components
given in Equation\ \eqref{eq:MIM-parts}.  To this end, first note the definition of $\MIMloss$:
\begin{align}
    \MIMloss(\params) ~ =~ \frac{1}{2}(\CE{\Msamp}{\Pdecoderjoint} +
    \CE{\Msamp}{\Pencoderjoint}) \label{eq:mimloss-supp}
\end{align}
We will focus on the first half of Equation~\eqref{eq:mimloss-supp} for now,
\begin{align}
    \frac{1}{2}\CE{\Msamp}{\Pdecoderjoint} ~=~ \frac{1}{4}\bigg(\CE{\Pdec}{\Pdecoderjoint}
      + \CE{\Penc}{\Pdecoderjoint}\bigg) \label{eq:ce-decoder}
\end{align}
It will be more clear to write out the first term of Equation\ \eqref{eq:ce-decoder},
$\frac{1}{4}\CE{\Pdec}{\Pdecoderjoint}$ in full
\begin{align*}
    \frac{1}{4}\CE{\Pdec}{\Pdecoderjoint} &= -\frac{1}{4}\int_{x,z} \Pdec\log(\Pdecoderjoint)\dxz \\
    &= -\frac{1}{4}\int_{\x,\z} \Pdec \log(\Mdec(\x | \z))\dxz \\
    & ~~~~~ - \frac{1}{4} \int_{\x,\z} \Pdec \log(\Mdec(\z))\dxz
\end{align*}
We then add and subtract $\frac{1}{4}\HD{\pjoint}{\z}$, where
\begin{align*}
    \frac{1}{4}\HD{\pjoint}{\z} ~=~ -\frac{1}{4}\int_{\x,\z}\pjoint(\z)\log(\pjoint(\z))\dxz
    ~=~ -\frac{1}{4}\int_{\x,\z}\Pdec\log(\pjoint(\z))\dxz
\end{align*}
Writing out $\frac{1}{4}\CE{\Pdec}{\Pdecoderjoint} + \frac{1}{4}\HD{\pjoint}{\z} - \frac{1}{4}\HD{\pjoint}{\z}$ and combining terms, we obtain
\begin{align}
    \frac{1}{4}\CE{\Pdec}{\Pdecoderjoint} ~=~ \HD{\Pdec}{\x, \z} + \DKL{\pjoint(\z)}{\Mdec(\z)} \label{eq:ce-decoding-decoder}
\end{align}
The second term in Equation\ \eqref{eq:ce-decoder} can then be rewritten as
\begin{align}
    \frac{1}{4}\CE{\Penc}{\Pdecoderjoint} ~=~ \frac{1}{4}\DKL{\Penc}{\Pdecoderjoint}
     + \frac{1}{4}\HD{\Penc}{\x, \z} \label{eq:ce-encoding-decoder}
\end{align}

Combining Equations\ \eqref{eq:ce-decoding-decoder}~and~\eqref{eq:ce-encoding-decoder}, we get the
interpretable form for Equation\ \ref{eq:ce-decoder}, \ie,
\begin{align}
    \frac{1}{2}\CE{\Msamp}{\Pdecoderjoint} &= \frac{1}{4}(\HD{\Pdec}{\x, \z} + \HD{\Penc}{\x, \z}) \nonumber \\
    &~~~~ + \frac{1}{4}\DKL{\pjoint(\z)}{\Mdec(\z)} + \frac{1}{4}\DKL{\Penc}{\Pdecoderjoint} \nonumber \\
    &= \frac{1}{2}\RH(\params) + \frac{1}{4}\DKL{\pjoint(\z)}{\Mdec(\z)} + \frac{1}{4}\DKL{\Penc}{\Pdecoderjoint} \label{eq:mimloss-interpretable-half}
\end{align}
We can use the same basic steps to derive an analogous expression for $\CE{\Msamp}{\Pencoderjoint}$ in Equation\ \eqref{eq:mimloss-supp} and combine it with Equation\ \eqref{eq:mimloss-interpretable-half} to get the final
interpretable form:
\begin{align*}
    \MIMloss(\params)~ &=  ~
    \mathrm{R}_\mathrm{H}(\params) ~+ ~
    \frac{1}{4}\Big(\, \DKL{\pjoint(\z)}{ \Mdec(\z)}  + \DKL{\pjoint(\x)}{\Menc(\x)} \Big)
    \nonumber \\
    & ~~~~~
    + ~ \frac{1}{4}\Big(\,\DKL{\Penc }{ \Mdec(\x, \z)} + \DKL{\Pdec}{ \Menc(\z , \x)} \Big)
\end{align*}

\section{MIM in terms of Symmetric KL Divergence}

As discussed above, the VAE objective can be expressed as minimizing the KL
divergence between the joint anchored encoding and anchored decoding distributions.
Below we consider a model formulation using the symmetric KL divergence (SKL),
\begin{align*}
    \mathrm{SKL}(\params) &=
    \frac{1}{2} \left ( \,\DKL{\Pdec}{\Penc} + \DKL{\Penc}{\Pdec}\, \right )
    ~,
\end{align*}
the second term of which is the VAE objective.
The mutual information regularizer $\mathrm{R}_\mathrm{H}(\params)$ given in
Equation~\eqref{eq:RH} can be added to SKL to obtain a cross-entropy
objective that looks similar to MIM:
\begin{align*}
    \frac{1}{2}\mathrm{SKL}(\params) + \mathrm{R}_{\mathrm{H}}(\params) &= \frac{1}{2}\left (\CE{\Msamp}{\Pdec} + \CE{\Msamp}{\Penc}\right )
\end{align*}
When the model priors are equal to the anchors, this regularized SKL and MIM are equivalent.
In general, however, the MIM loss is not a bound on the regularized SKL.

In what follows we explore the relation between SKL and JSD.
In Section~\ref{sec:jsd-entropy-supp} we showed that the Jensen-Shannon divergence
can be written as
\begin{align*}
    \mathrm{JSD}(\params) &= \frac{1}{2} \bigg ( \CE{\Pdec}{\Msamp} - \HD{\Pdec}{\x, \z} \\
    & ~~~~~ + \CE{\Penc}{\Msamp} - \HD{\Penc}{\x, \z} \bigg ) \\
    &= \frac{1}{2} \left ( \CE{\Pdec}{\Msamp} + \CE{\Penc}{\Msamp} \right ) - \RH(\params)
\end{align*}
Using Jensen's inequality, we can bound $\mathrm{JSD}(\params)$ from above,
\begin{align}
    \mathrm{JSD}(\params) & \leq \frac{1}{4} \bigg( \HD{\Pdec}{\x, \z} + \CE{\Pdec}{\Penc} \nonumber \\
    &~~~~~~ + \HD{\Penc}{\x, \z} + \CE{\Penc}{\Pdec}\bigg) - \RH(\params) \label{eq:skl-rh} \\
    &= \frac{1}{4} \bigg( \CE{\Pdec}{\Penc} + \CE{\Penc}{\Pdec} \nonumber \\
    &~~~~~~ + 2\RH(\params) \bigg) - \RH(\params) \nonumber \\
    &= \frac{1}{4} \bigg( \DKL{\Pdec}{\Penc} + \DKL{\Penc}{\Pdec} \nonumber \\
    &~~~~~~ + 4\RH(\params) \bigg) - \RH(\params) \nonumber \\
    &= \frac{1}{4} \left (\DKL{\Pdec}{\Penc} + \DKL{\Penc}{\Pdec} \right ) \nonumber \\
    &= \frac{1}{2}\mathrm{SKL}(\params) \nonumber
\end{align}
From Equation~\eqref{eq:skl-rh}, if we add the regularizer $\RH(\params)$ and combine terms, we get
\begin{align*}
    \frac{1}{2}\mathrm{SKL}(\params) + \RH(\params) &= \frac{1}{2}\left ( \CE{\Msamp}{\Penc}  + \CE{\Msamp}{\Pdec}\right )
\end{align*}
Interestingly, we can write this in terms of KL divergence,
\begin{align*}
    \frac{1}{2}\mathrm{SKL}(\params) + \RH(\params) &= \frac{1}{2}\left ( \DKL{\Msamp}{\Penc}  + \DKL{\Msamp}{\Pdec} \right ) + \HD{\Msamp}{\x, \z} \\
    &= \frac{1}{2}\left (\DKL{\Msamp}{\Penc} + \DKL{\Msamp}{\Pdec}\right ) \\
    & ~~~~~~ + \mathrm{JSD}(\params) + \RH(\params)
\end{align*}
which gives the exact relation between JSD and SKL.
\begin{align*}
    \frac{1}{2}\mathrm{SKL}(\params) &= \frac{1}{2}\left ( \DKL{\Msamp}{\Penc}  + \DKL{\Msamp}{\Pdec} \right ) + \mathrm{JSD}(\params) \\
    &= \frac{1}{2}\left ( \DKL{\Msamp}{\Penc}  + \DKL{\Msamp}{\Pdec} \right ) \\
    & ~~~~~~+ \frac{1}{2}\left ( \DKL{\Penc}{\Msamp}  + \DKL{\Pdec}{\Msamp} \right )
\end{align*}



\section{Learning}


\subsection{\MIM Parametric Priors}
\label{sec:mim-parametric-priors}

There are several effective ways to parameterize the priors.
For the 1D experiments below we model $\Mdec(\z)$ using linear mixtures of isotropic Gaussians.
With complex, high dimensional data one might also consider more powerful
models (\eg, autoregressive, or flow-based priors).  Unfortunately, the use
of complex models typically increases the required computational resources,
and the training and inference time.
As an alternative we use for image data the {\em vampprior}
\cite{DBLP:journals/corr/TomczakW17}, which models the latent prior as a
mixture of posteriors, \ie, $\Mdec(\z) = \sum_{k=1}^{K} \Menc(\z|\x=\bs{u}_k)$
with learnable pseudo-inputs $\{\bs{u}_k\}_{k=1}^{K}$.  This is effective
and allows one to reduce the need for additional parameters
(see \cite{DBLP:journals/corr/TomczakW17} for details on vampprior's effect over gradient estimation).

Another useful model with high dimensional data,
following \cite{DBLP:journals/corr/BornscheinSFB15}, is to define
$\Menc(\x)$ as the marginal of the decoding distribution; \ie,
\begin{equation}
    \Menc(\x) ~=~ \E{\z \sim \Mdec(\z)}{\,\Mdec(\x|\z)\,} ~.
    \label{eqn:q-marginal}
\end{equation}
Like the vampprior, this entails no new parameters.  It also helps to
encourage consistency between the encoding and decoding distributions.
In addition it enables direct empirical comparisons of
VAE learning to MIM learning, because we can then use identical
parameterizations and architectures for both.
During learning, when $\Menc(\x)$ is defined as the marginal \eqref{eqn:q-marginal},
we evaluate $\log \Menc(\x)$ with a single sample and reparameterization.
When $\z$ is drawn directly from the latent prior:
\begin{equation}
    \log \Menc(\x) ~ =~ \log \E{\tilde{\z} \sim \Mdec(\z)}{\,\Mdec(\x|\tilde{\z})\,}     ~\approx~
    \log  \Mdec(\x|\tilde{\z}) ~.
    \nonumber
\end{equation}
When $\z$ is drawn from the encoder, given a sample observation, we use importance sampling:
\begin{equation}
    \log \Menc(\x) ~=~ \log \E{\tilde{\z} \sim \Menc(\z|\x)}{ \,\Mdec(\x|\tilde{\z})\frac{\Mdec(\tilde{\z})}{\Menc(\tilde{\z}|\x)} \, }     ~\approx~
    \log  \Mdec(\x|\tilde{\z}) + \log \Mdec(\tilde{\z})  - \log \Menc(\tilde{\z}|\x)
    \nonumber
\end{equation}
Complete derivations for learning with the marginal prior are provided in Sec.\ \ref{sec:learning-implicit}.


\subsection{Gradient Estimation}

Optimization is performed through minibatch stochastic gradient
descent. To ensure unbiased gradient estimates of $\EMIMloss$ we
use the reparameterization trick \cite{Kingma2013,Rezende2014} when
taking expectation with respect to continuous encoder and decoder
distributions, $\Menc(\z | \x)$ and $\Mdec(\x | \z)$.
Reparameterization entails sampling an auxiliary variable
$\epsilon \sim p(\epsilon) $, with known $p(\epsilon)$, followed
by a deterministic mapping from sample variates to the target random
variable, that is $p_{\params}(\z) = g_{\params}(\epsilon)$ and
$q_{\params}(\z|\x) = h_{\params}(\epsilon, \x)$ for prior and
conditional distributions. In doing so we assume $p(\epsilon)$
is independent of the parameters $\params$. It then follows that
\begin{equation*}
    \nabla_{\params} \E{\z \sim q_{\params}(\z|\x)}{f_{\params}(\z)}
    ~=~
    \nabla_{\params} \E{\epsilon \sim p(\epsilon)}{f_{\params}(h_{\params}(\epsilon, \x))}
    ~=~
    \E{\epsilon \sim p(\epsilon)}{\nabla_{\params} f_{\params}(h_{\params}(\epsilon, \x))}
\end{equation*}
where $f_{\params}(\z)$ is the loss function with parameters $\params$.
It is common to let $p(\epsilon)$ be standard normal, $\epsilon \sim \mathcal{N}(0,1)$,
and for $\z | \x$ to be Gaussian with mean $\mu_{\params}(\x)$ and
standard deviation $\sigma_{\params}(\x)$, in which case
$ \z = \sigma_{\params}(\x) \, \epsilon + \mu_{\params}(\x) $,
A more generic exact density model can be learned
by mapping a known base distribution (\eg, Gaussian) to a target distribution
with normalizing flows \cite{Dinh2014,Dinh2016a,Rezende2015}.

In the case of discrete distributions, \eg, with discrete data,
reparameterization is not readily applicable.
There exist continuous relaxations that permit reparameterization (\eg, \cite{DBLP:journals/corr/MaddisonMT16,DBLP:journals/corr/TuckerMMS17}), but
current methods are rather involved in practice, and require adaptation of the
objective function or the optimization process.
Here we simply use the REINFORCE algorithm \cite{Sutton:1999:PGM:3009657.3009806}
for unbiased gradient estimates, as follows
\begin{align*}
    \nabla_{\params} \E{\z \sim q_{\params}(\z|\x)}{f_{\params}(\z)}
    & = ~\E{\z \sim q_{\params}(\z|\x)}{\nabla_{\params} f_{\params}(\z) + f_{\params}(\z) \nabla_{\params} \log q_{\params}(\z|\x)}  ~.
\end{align*}
A detailed derivation follows the use of the relation below,
\begin{equation*}
    \nabla_{\params} q_{\params}(\z|\x) ~=~ q_{\params}(\z|\x) \nabla_{\params} \log q_{\params}(\z|\x)
\end{equation*}
in order to provide unbiased gradient estimates as follows
\begin{align*}
    \nabla_{\params} \E{\z \sim q_{\params}(\z|\x)}{f_{\params}(\z)}
    & = ~ \nabla_{\params} \int f_{\params}(\z) \, q_{\params}(\z|\x) \, d\z \\
    & = ~\int  q_{\params}(\z|\x)\, \nabla_{\params}f_{\params}(\z) \, d\z
       + \int f_{\params} (\z)\, \nabla_{\params} q_{\params}(\z|\x) \, d\z \\
    & =~ \int  q_{\params}(\z|\x)\, \nabla_{\params}f_{\params}(\z) \, d\z
       + \int f_{\params} (\z) \, q_{\params}(\z|\x) \,\nabla_{\params} \log q_{\params}(\z|\x) \, d\z \\
    & = ~\E{\z \sim q_{\params}(\z|\x)}{\nabla_{\params} f_{\params}(\z) + f_{\params}(\z) \nabla_{\params} \log q_{\params}(\z|\x)}
\end{align*}
which facilitate the use of samples to approximate the integral.


\subsection{Training Time}
\label{sec:training-time}

Training times of MIM models are comparable to training times for VAEs with
comparable architectures. One important difference concerns the time required
for sampling from the decoder during training. This is particularly significant
for models like auto-regressive decoders (\eg, \cite{Kingma2016}) for which sampling
is very slow.  In such cases, we find that we can also learn effectively with a
sampling distribution that only includes samples from the encoding distribution,
\ie, $ \pjoint(\x)\, \Menc(\z | \x)$, rather than the mixture.
We refer to this particular MIM variant as asymmetric MIM (or A-MIM).
We use it in Sec.\ \ref{sec:high-dimensional-image-data} when working with
the PixelHVAE architecture \cite{Kingma2016}.

\section{Posterior Collapse in VAE}
\label{sec:posterior-collapse-in-vae}

Here we discuss a possible root cause for the observed phenomena of posterior collapse, and show that VAE learning can be viewed as an asymmetric MIM learning
with a regularizer that encourages the appearance of the collapse. We further support that idea in the experiments in Section\ \ref{sec:posterior-collapse-mim-vae}.
As discussed earlier, VAE learning entails maximization of a variational lower
bound (ELBO) on the log-marginal likelihood, or equivalently, given
Equation  \eqref{eq:vi-as-ml-objective-text}, the VAE loss in terms of expectation
over a joint distribution:
\begin{eqnarray}
- \E{\x \sim \pjoint (\x),\z \sim \Menc(\z|\x)}
{ \, \log \Mdec(\x|\z) + \log \pjoint(\z) - \log \Menc(\z|\x) \,}
~.
\label{eq:vi-as-joint-objectiveA}
\end{eqnarray}
To connect the loss in Equation\ \eqref{eq:vi-as-joint-objectiveA} to MIM, we first add
the expectation of $\log \pjoint(\x)$, and scale the loss by a factor of $\frac{1}{2}$,
to obtain
\begin{eqnarray}
\E{\x \sim \pjoint (\x),\z \sim \Menc(\z|\x)}
{\,-\frac{1}{2} \left( \log ( \Mdec(\x|\z)\pjoint(\z) )
+ \log ( \Menc(\z|\x)\pjoint(\x) ) \right)
+ \log \pjoint(\x) + \log \Menc(\z|\x) \,}
\label{eq:vi-as-joint-objective}
\end{eqnarray}
where $\pjoint(\x)$ is the data distribution, which is assumed to be independent of model
parameters $\theta$ and to exist almost everywhere (\ie, complementing $\pjoint(\z)$).
Importantly, because $\pjoint(\x)$ does not depend on $\params$, the gradients of
Eqs.\ \eqref{eq:vi-as-joint-objectiveA} and \eqref{eq:vi-as-joint-objective}
are identical up to a multiple of $\frac{1}{2}$, so they share the same stationary points.

Combining IID samples from the data distribution, $\x^i \sim \pjoint(\x)$, with samples
from the corresponding variational posterior, $\z^i \sim \Menc(\z|\x^i)$, we obtain a
joint sampling distribution; \ie,
\begin{equation*}
 \Msamp^\mathrm{VAE}(\x,\z) ~=~ \pjoint (\x)\,\Menc(\z|\x)
\end{equation*}
where $\Msamp^\mathrm{VAE}$ comprises the encoding distribution in $\Msamp$.
With it one can then rewrite the objective in Equation\ \eqref{eq:vi-as-joint-objective}
in terms of the cross-entropy between $\Msamp^\mathrm{VAE}$ and the parametric encoding
and decoding distributions; \ie,
\begin{align}
& \frac{1}{2} \big(\, \CE{\Msamp^\mathrm{VAE}}{\Mdec(\x|\z)\pjoint(\z)}
+ \CE{\Msamp^\mathrm{VAE}}{\Menc(\z|\x)\pjoint(\x)}\, \big) + \nonumber \\
& - \HD{\Msamp^\mathrm{VAE}}{\x} - \HD{\Msamp^\mathrm{VAE}}{\z} + I_{\Msamp^\mathrm{VAE}}(\x;\z)  ~.
\label{eq:vi-as-ml-objective}
\end{align}
The sum of the last three terms in Equation\ \eqref{eq:vi-as-ml-objective}
is the negative joint entropy $-\HD{\Msamp^\mathrm{VAE}}{\z, \x}$ under the sample distribution $\Msamp^\mathrm{VAE}$.

Equations\ \eqref{eq:vi-as-ml-objective-text} and \eqref{eq:vi-as-ml-objective}, the VAE objective and
VAE as regularized cross entropy objective respectively, define
equivalent optimization problems, under the assumption that  $\pjoint(\x)$ and
samples $\x \sim \pjoint(\x)$ do not depend on the parameters $\params$, and that
the optimization is gradient-based.
Formally, the VAE objectives \eqref{eq:vi-as-ml-objective-text} and \eqref{eq:vi-as-ml-objective}
are equivalent up to a scalar multiple of $\frac{1}{2}$ and an additive constant,
namely, $\HD{\Msamp^\mathrm{VAE}}{\x}$.

Equation \eqref{eq:vi-as-ml-objective} is the average of two cross-entropy objectives (
\ie, between sample distribution $\Msamp^\mathrm{VAE}$ and the model decoding and encoding distributions,
respectively), along with a joint entropy term (\ie, last three terms), which can be viewed as a
regularizer that encourages a reduction in mutual information and increased entropy in $\z$ and $\x$.
We note that Equation \eqref{eq:vi-as-ml-objective} is similar to the \MIM objective in Equation \eqref{eq:mimloss},
but with a different sample distribution, where the priors are defined to be the anchors, and with an additional regularizer.
In other words, Equation\ \eqref{eq:vi-as-ml-objective} suggests that VAE learning
implicitly lowers mutual information.
This runs contrary to the goal of learning useful latent representations, and we
posit that it is an underlying root cause for {\em posterior collapse},
wherein the trained model show low mutual information which can be manifested as an encoder which matches the prior, and thus provides weak information about the latent state (\eg, see \citep{ChenKSDDSSA16} and others). We point the reader to Section\ \ref{sec:entropy-as-mi-regularizer}
for empirical evidence for the use of a joint entropy as a mutual information regularizer.

\section{Additional Experiments}

Here we provide additional experiments that further explore the characteristics of \MIM learning.

\subsection{Entropy as Mutual Information Regularizer} \label{sec:entropy-as-mi-regularizer}

\begin{figure}[ht]
    \centering
    \setlength{\tabcolsep}{0pt}
    \begin{tabular}{*6{>{\centering\arraybackslash}m{0.167\textwidth}}}
      {\scriptsize VAE+H} & {\scriptsize MIM-H} & {\scriptsize VAE+H} & {\scriptsize MIM-H} & {\scriptsize VAE+H} & {\scriptsize MIM-H} \\
      \includegraphics[width=0.165\columnwidth]{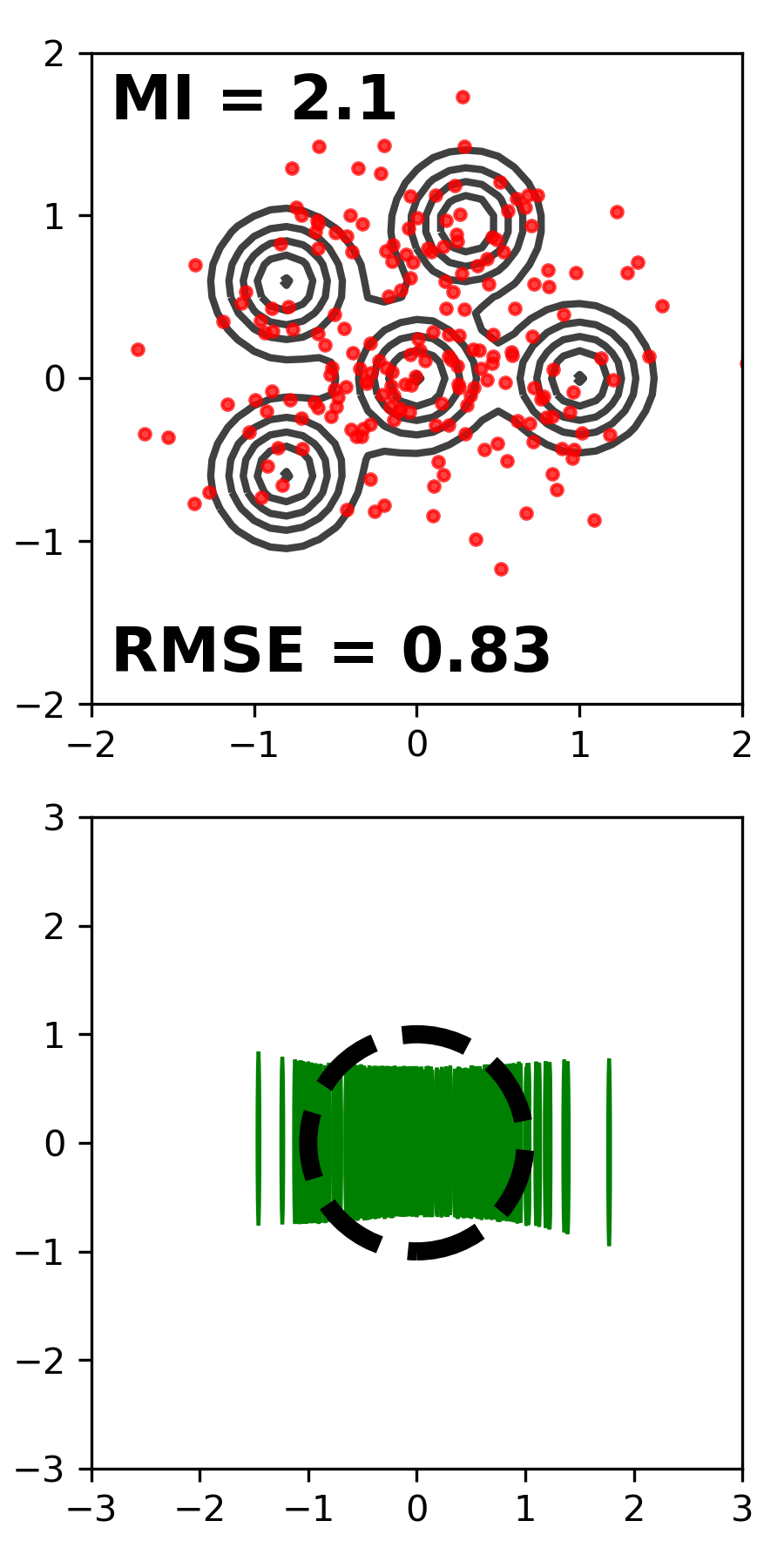}
    & \includegraphics[width=0.165\columnwidth]{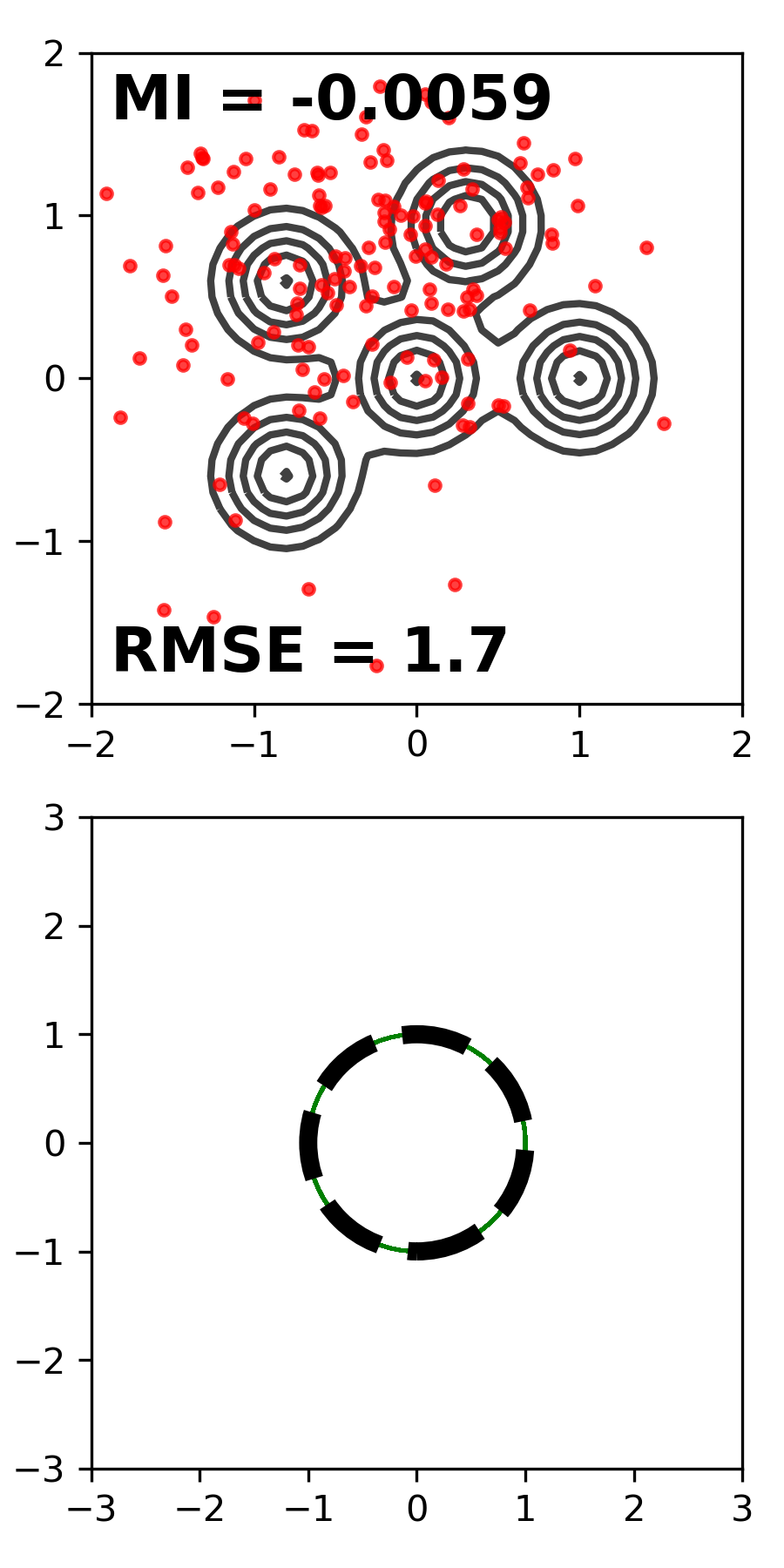}
    & \includegraphics[width=0.165\columnwidth]{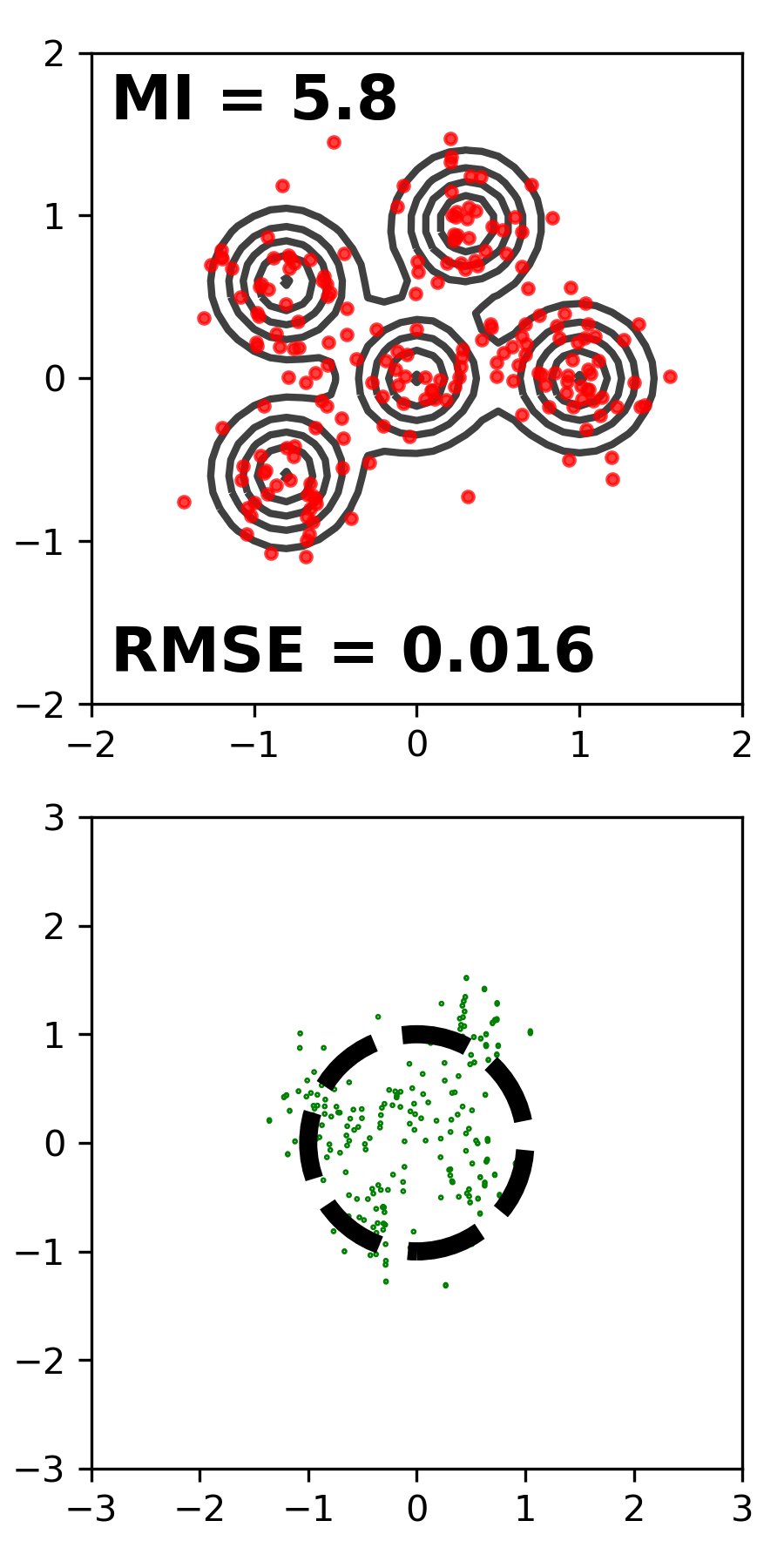}
    & \includegraphics[width=0.165\columnwidth]{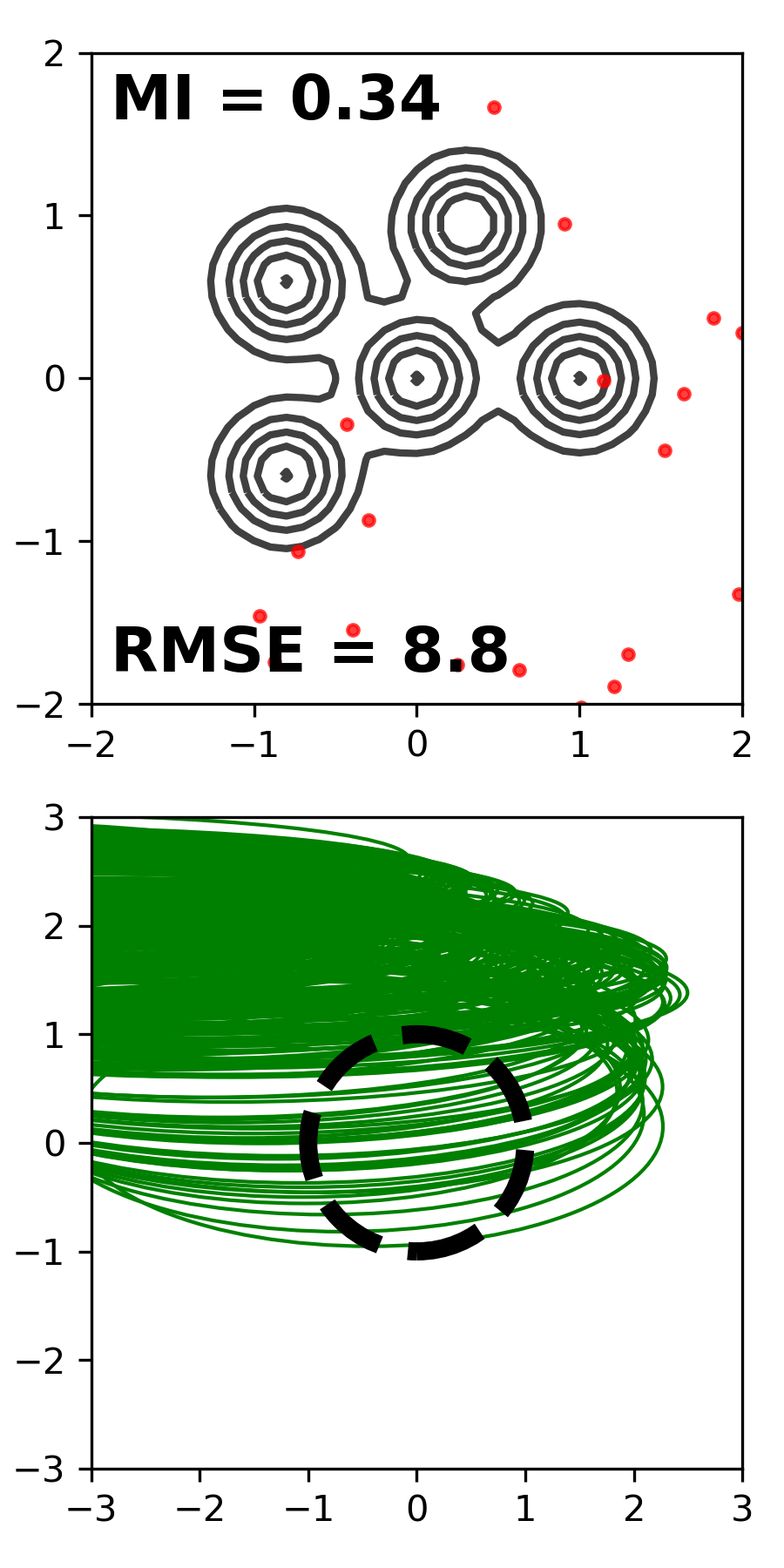}
    & \includegraphics[width=0.165\columnwidth]{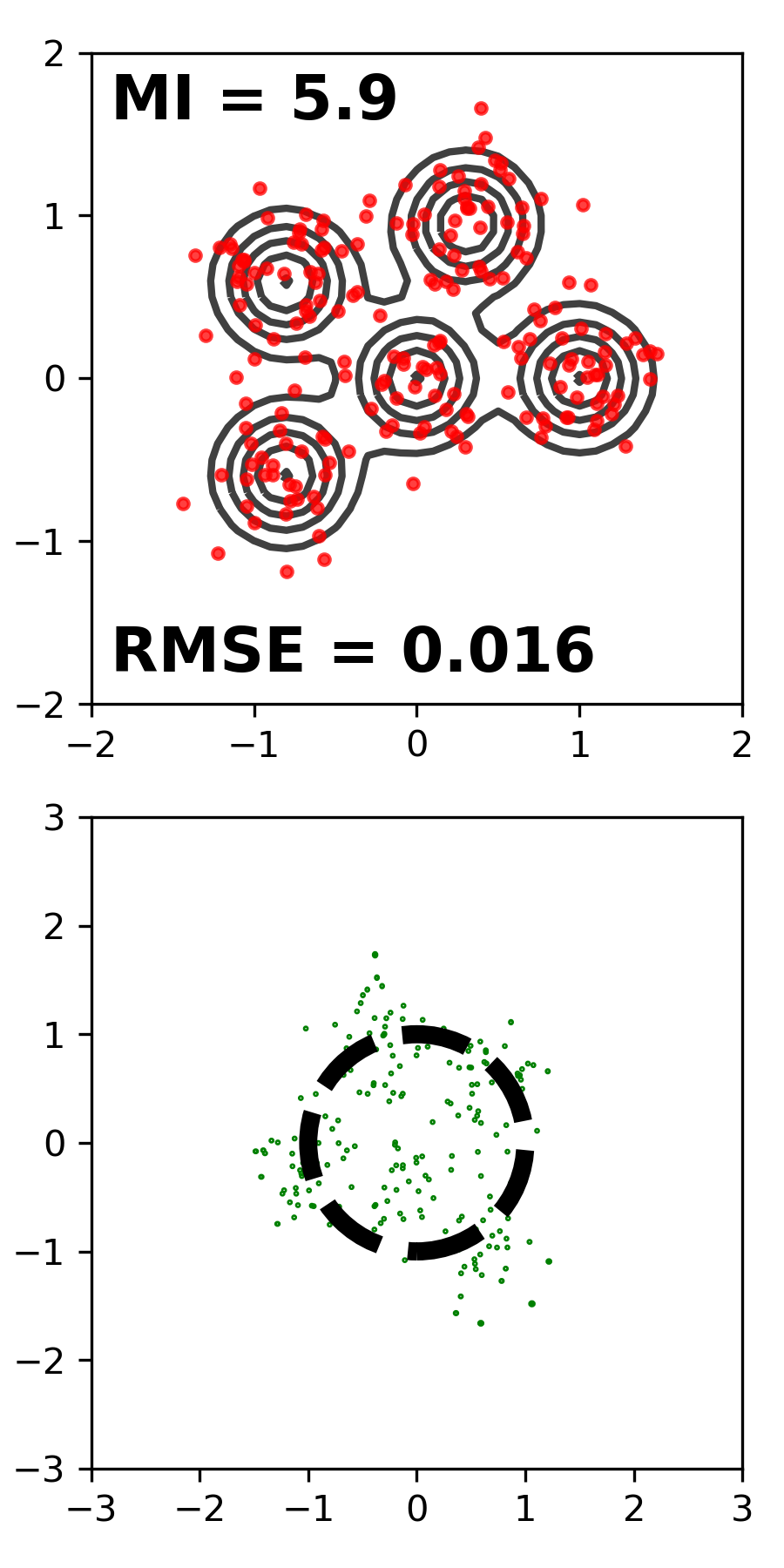}
    & \includegraphics[width=0.165\columnwidth]{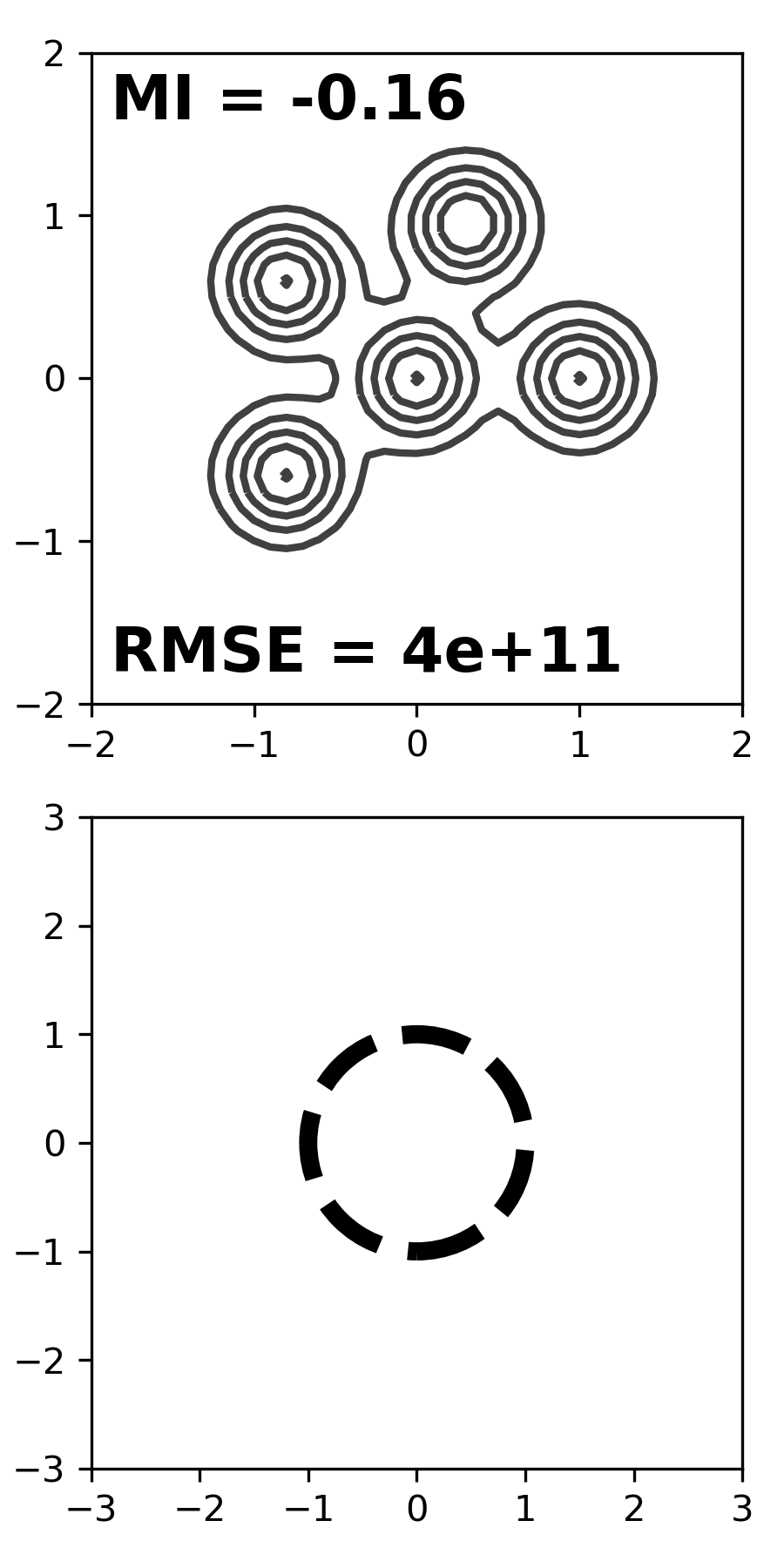}
    \\
    \multicolumn{2}{c}{(a) $h \in \mathbb{R}^{5}$ } & \multicolumn{2}{c}{(b) $h \in \mathbb{R}^{20}$ } & \multicolumn{2}{c}{(c) $h \in \mathbb{R}^{500}$ } \\
    \end{tabular}
    \caption{
    Effects of entropy as a mutual information regularizer in 2D $\x$ and 2D $\z$ synthetic problem.
    VAE and MIM models with 2D inputs, a 2D latent space, and 5, 20 and 500 hidden units.
    Top row: Black contours depict level sets of $\pjoint(\x)$; red dots are reconstructed test points.
    Bottom row: Green contours are one standard deviation ellipses of $\Menc(\z|\x)$ for test
    points. Dashed black circles depict one standard deviation of $\pjoint (\z)$.
    Here we added $H_{\Menc}(\x, \z)$ to VAE loss, and subtracted $H_{\Msamp}(\x, \z)$ from MIM loss, in order to demonstrate the effect of entropy on mutual information.
    Posterior collapse in VAE is mitigated following the increased mutual information.
    MIM, on the other hand, demonstrates a severe posterior collapse as a result of the reduced mutual information (\ie, posterior matches prior over $\z$ almost perfectly).
    (see inset quantities).
    }\label{fig:entropy-regularizer-qualitative}
\end{figure}

\begin{figure}[ht]
    \centering
    \setlength{\tabcolsep}{0pt}
    \begin{tabular}{*4{>{\centering\arraybackslash}m{0.25\textwidth}}}
      \includegraphics[width=0.24\columnwidth]{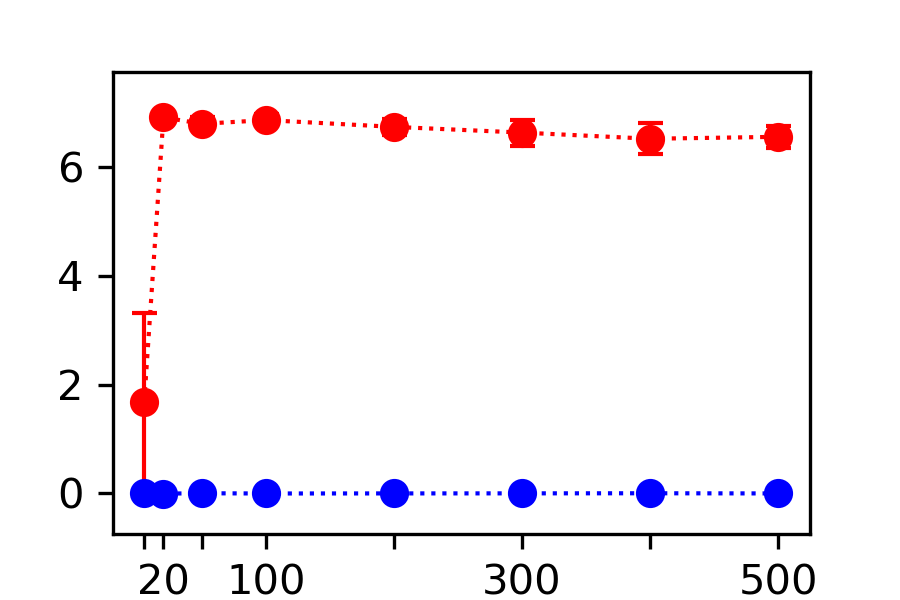}
    & \includegraphics[width=0.24\columnwidth]{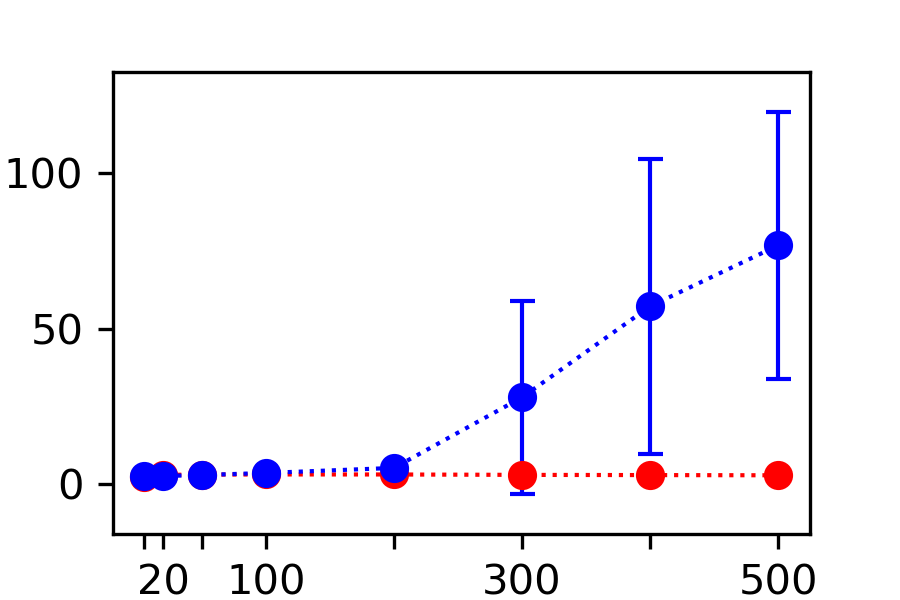}
    & \includegraphics[width=0.24\columnwidth]{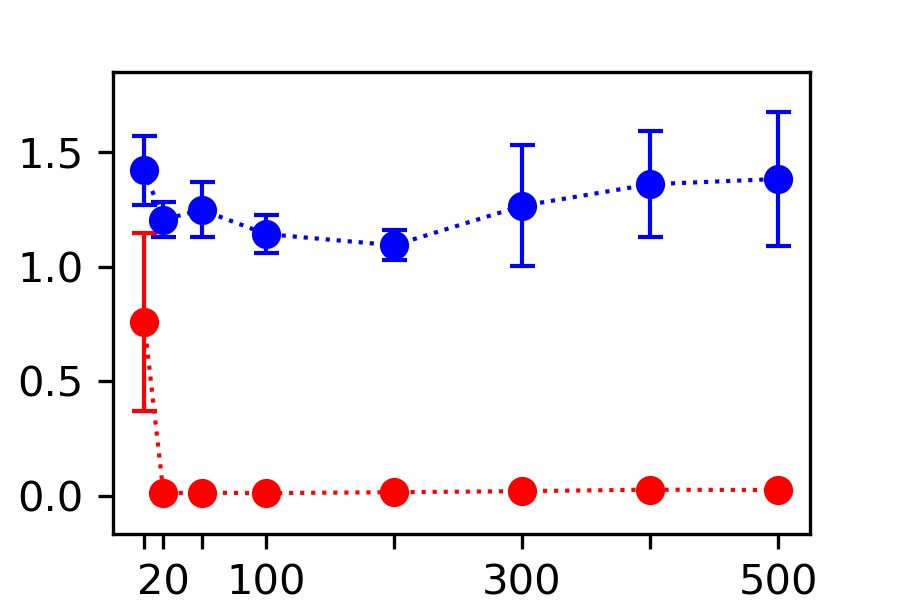}
    & \includegraphics[width=0.24\columnwidth]{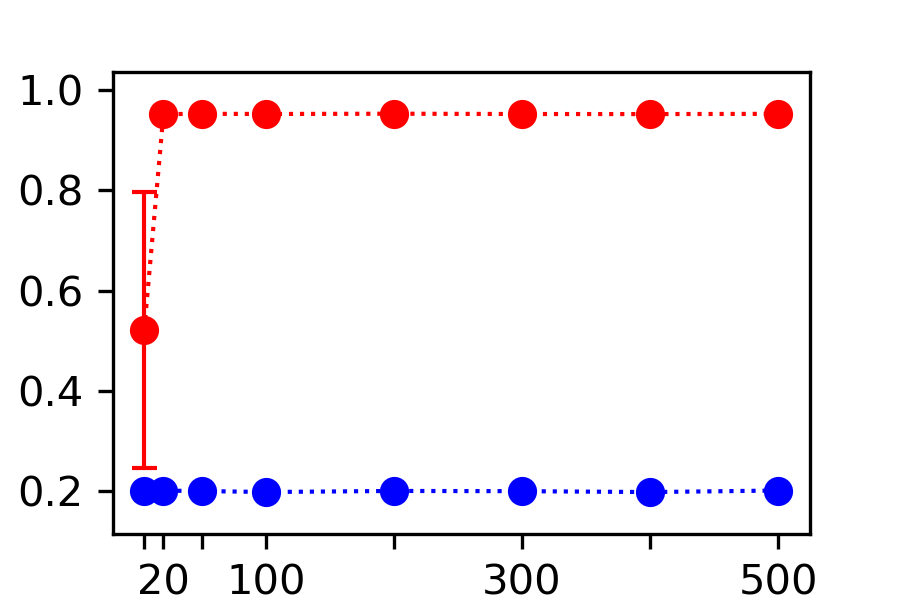}
    \\
    (a) MI & (b)  NLL & (c) Recon.\ Error & (d) Classif.\ (5-NN)
    \end{tabular}
    \caption{
    Effects of entropy as a mutual information regularizer in 2D $\x$ and 2D $\z$ synthetic problem.
    Test performance for modified MIM (blue) and modified VAE (red) for the 2D GMM data with
    (cf.\ Fig.\ \ref{fig:entropy-regularizer-qualitative}), all as functions of the
    number of hidden units (on x-axis).
    Each plot shows the mean and standard deviation of 10 experiments.
    Adding encoding entropy regularizer to VAE loss leads to high mutual information (\ie, prevent posterior collapse), low reconstruction error, and better classification accuracy.
    Subtracting sample entropy regularizer from MIM loss results in almost zero mutual information (severe collapse), which leads to poor reconstruction error and classification accuracy.
    }\label{fig:entropy-regularizer-quantitative}
\end{figure}

Here we examine the use of entropy as a mutual information regularizer. We repeat the experiment in Section \ref{sec:posterior-collapse-mim-vae} with added entropy regularizer. Figure\ \ref{fig:entropy-regularizer-qualitative} depicts the effects of an added $H_{\Menc}(\x, \z)$ to the VAE loss, and a subtracted $H_{\Msamp}(\x, \z)$ from \MIM.
The corresponding quantitative values are presented in Figure\ \ref{fig:entropy-regularizer-quantitative}. Adding the entropy regularizer leads to increased the mutual information, and subtracting it results in a strong posterior collapse, which in turn is reflected in the reconstruction quality. While such an experiment does not represent a valid probabilistic model, it supports our use of entropy as a regularizer (cf. Eq. \eqref{eq:RH}) for JSD in order to define a consistent model with high mutual information.

\FloatBarrier

\subsection{Consistency regularizer in \texorpdfstring{$\MIMloss$}{Lmim}}

\begin{figure}[ht]
    \centering
    \begin{minipage}[b]{0.33\columnwidth}
    \centering
    Reconstruction \\
    \begin{minipage}[b]{0.5\columnwidth}
    \centering
    \includegraphics[width=0.99\columnwidth]{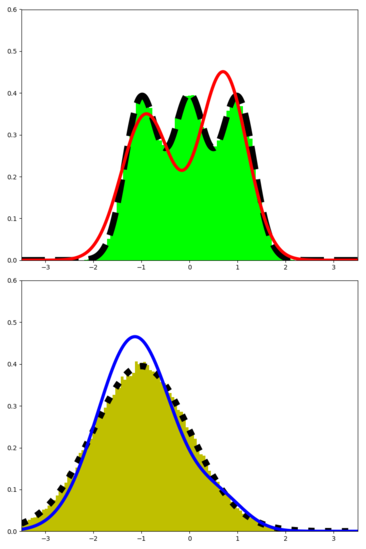}
    \subcaption{CE}
    \end{minipage}%
    \begin{minipage}[b]{0.5\columnwidth}
    \centering
    \includegraphics[width=0.99\columnwidth]{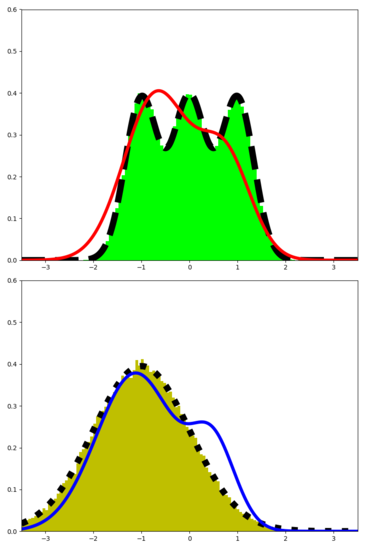}
    \subcaption{\MIM}
    \end{minipage}
    \end{minipage}%
    \begin{minipage}[b]{0.33\columnwidth}
    \centering
    Anchor Consistency \\
    \begin{minipage}[b]{0.5\columnwidth}
    \centering
    \includegraphics[width=0.99\columnwidth]{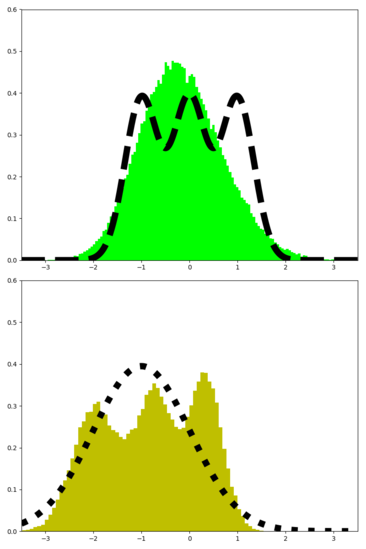}
    \subcaption{CE}
    \end{minipage}%
    \begin{minipage}[b]{0.5\columnwidth}
    \centering
    \includegraphics[width=0.99\columnwidth]{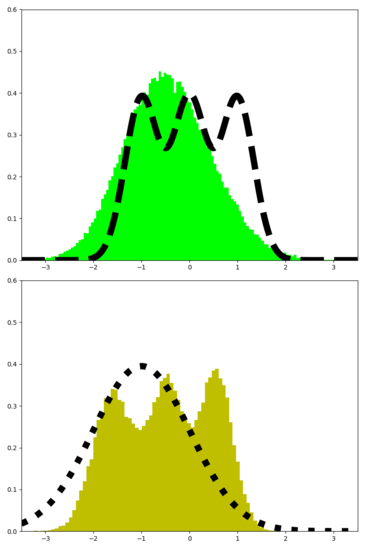}
    \subcaption{\MIM}
    \end{minipage}
    \end{minipage}%
    \begin{minipage}[b]{0.33\columnwidth}
    \centering
    Prior Consistency \\
    \begin{minipage}[b]{0.5\columnwidth}
    \centering
    \includegraphics[width=0.99\columnwidth]{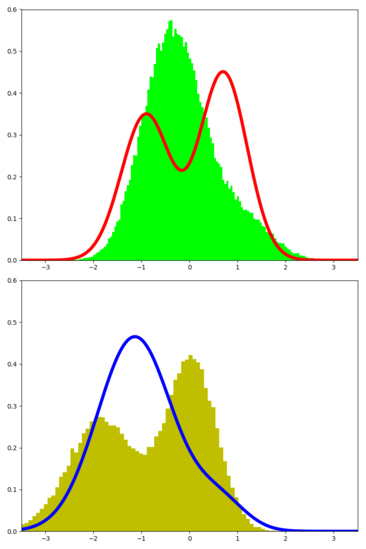}
    \subcaption{CE}
    \end{minipage}%
    \begin{minipage}[b]{0.5\columnwidth}
    \centering
    \includegraphics[width=0.99\columnwidth]{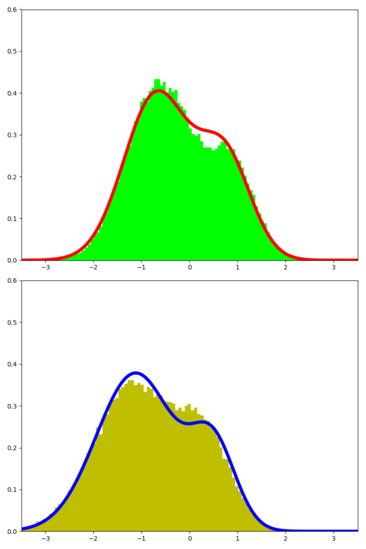}
    \subcaption{\MIM}
    \end{minipage}
    \end{minipage}
    \caption{We explore the influence of consistency regularizer $R_{\params}$.
    CE and MIM indicate the loss, $\CEloss$ or $\MIMloss$ (the regularized objective), respectively. Top row shows anchor $\pjoint(\x)$ (dashed), prior $\Menc(\x)$ (red), and reconstruction distribution $\x_i \sim \pjoint(\x)\rightarrow \z_i \sim \Menc(\z|\x_i)\rightarrow x_i' \sim \Mdec(\x|\z_i)$ (green). Bottom row mirrors the top row, with anchor $\pjoint(\z)$ (dotted), prior $\Mdec(\z)$ (blue), and reconstruction distribution $\z_i \sim \pjoint(\z)\rightarrow \x_i \sim \Mdec(\x|\z_i)\rightarrow z_i' \sim  \Menc(\z|\x)$ (yellow).
    In (c-d) the reconstruction is replaced with decoding from anchors $\z_i \sim \pjoint(\z)\rightarrow x_i' \sim \Mdec(\x|\z_i)$ (green), and encoding $\x_i \sim \pjoint(\x) \rightarrow z_i' \sim  \Menc(\z|\x)$ (yellow).
    In (e-f) the reconstruction is replaced with decoding from priors $\z_i \sim \Mdec(\z) \rightarrow x_i' \sim \Mdec(\x|\z_i)$ (green), and encoding $\x_i \sim \Menc(\x) \rightarrow z_i' \sim  \Menc(\z|\x)$ (yellow).
    While both models offers similar reconstruction (a-b), and similar consistency
    w.r.t.\ the anchors (c-d), only \MIM finds a consistent model (e-f).
    See text for details.
    }
    \label{fig:mim-consistency}
\end{figure}

Here we explore properties of models for 1D $\x$ and $\z$,
learned with $\MIMloss$ and $\CEloss$, the difference being the model consistency regularizer $\RMIM(\params)$.
All model priors and conditional likelihoods ($\Menc(\x)$, $\Menc(\z|\x)$, $\Mdec(\z)$, $\Mdec(\x|\z)$)
are parameterized as 10-component Gaussian mixture models, and optimized during training.
Means and variances for the conditional distributions were regressed with 2 fully connected
layers ($h \in \mathbb{R}^{10}$) and a swish activation function \cite{Ramachandran2017}.

Top and bottom rows in Fig.\ \ref{fig:mim-consistency} depict distributions in
observations and latent spaces respectively. Dashed black curves are anchors, $\pjoint(\x)$ on top,
and $\pjoint(\z)$ below (GMMs with up to 3 components). Learned model priors, $\Menc(\x)$
and $\Mdec(\z)$, are depicted as red (top) and blue (bottom) curves.

Green histograms in Fig.\ \ref{fig:mim-consistency}(a,b) depict
reconstruction distributions, computed by passing fair samples from $\pjoint(\x)$
through the encoder to $\z$ and then back through the decoder to $\x$.
Similarly the yellow histograms shows samples from $\pjoint(\z)$ passed through the
decoder and then back to the latent space. For both losses these reconstruction histograms match the anchor priors well. In contrast, only the priors that were learned with $\CEloss$ loss approximates the anchor well, while the $\MIMloss$ priors do not. To better understand that, we consider two generative procedures: sampling from the anchors, and sampling from the priors.

Anchor consistency is depicted in Fig.\ \ref{fig:mim-consistency}(c,d),
where Green histograms  are marginal distributions
over $\x$ from the anchored decoder (\ie, samples from $\pjoint(\z) \Mdec(\x |\z)$ ).
Yellow are marginals over $\z$ from the anchored encoders $\pjoint(\x) \Menc(\z |\x)$.  One can see that both losses results in similar quality of matching the corresponding opposite anchors.

Priors consistency is depicted in Fig.\ \ref{fig:mim-consistency}(e,f),
where Green histograms  are marginal distributions
over $\x$ from the model decoder $\Mdec(\z) \Mdec(\x |\z)$.
Yellow depicts marginals over $\z$ from the model encoder $\Menc(\x) \Menc(\z |\x)$.
Importantly, with $\MIMloss$ the encoder and decoder are consistent; \ie, $\Menc(\x)$ (red curve)
matches the decoder marginal, while $\Mdec(\z)$ (blue) matches the encoder marginal.
The model trained with $\CEloss$ (\ie, without consistency prior) fails to learn a consistent encoder-decoder pair. We note that in practice, with expressive enough priors, $\MIMloss$ will be a tight bound for $\CEloss$.

\FloatBarrier


\subsection{Parameterizing the Priors} \label{sec:parameterizing-priors}

\begin{figure}[ht]
    \centering
    \begin{minipage}[b]{0.25\columnwidth}
    \centering
    \includegraphics[width=0.65\columnwidth]{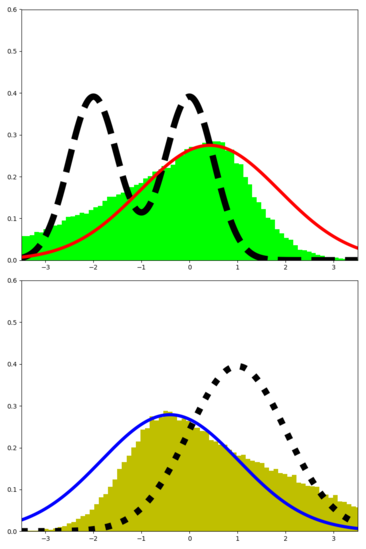}
    \subcaption{$\Menc(\x),\Mdec(\z)$}
    \end{minipage}%
    \begin{minipage}[b]{0.25\columnwidth}
    \centering
    \includegraphics[width=0.65\columnwidth]{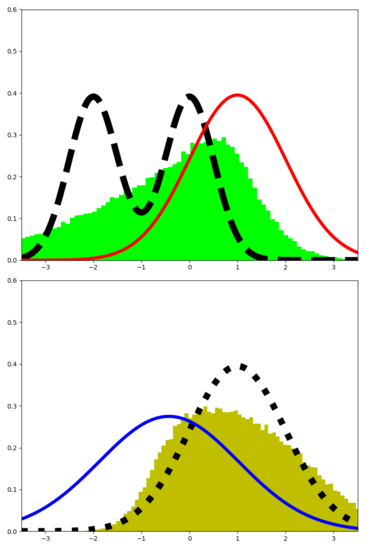}
    \subcaption{$\penc(\x),\Mdec(\z)$}
    \end{minipage}%
    \begin{minipage}[b]{0.25\columnwidth}
    \centering
    \includegraphics[width=0.65\columnwidth]{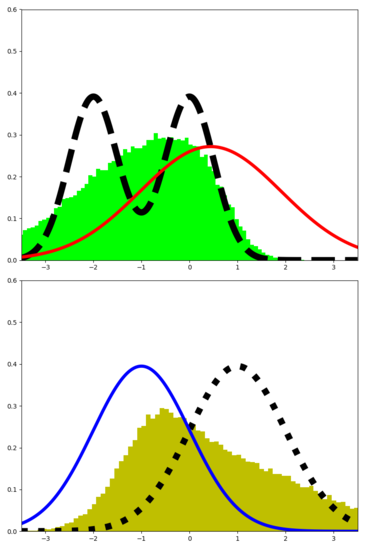}
    \subcaption{$\Menc(\x),\pdec(\z)$}
    \end{minipage}%
    \begin{minipage}[b]{0.25\columnwidth}
    \centering
    \includegraphics[width=0.65\columnwidth]{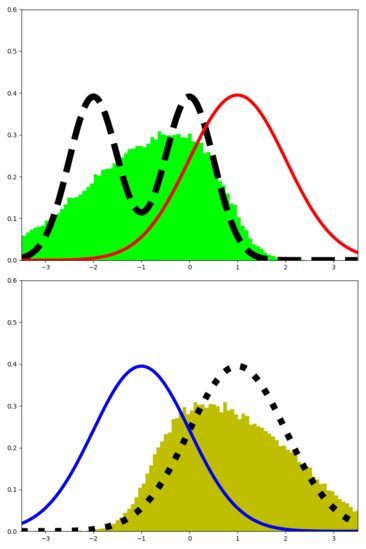}
    \subcaption{$\penc(\x),\pdec(\z)$}
    \end{minipage}
    \caption{\MIM prior expressiveness.  In this experiment we explore the effect of learning a prior, where the priors $\penc(\x)$ and $\pdec(\z)$ are normal Gaussian distributions. Top row shows anchor $\pjoint(\x)$ (dashed), prior $\Menc(\x)$ (red), and decoding distribution $\z_i \sim \Mdec(\z)\rightarrow x_i' \sim \Mdec(\x|\z_i)$ (green). Bottom row mirrors the top row, with anchor $\pjoint(\z)$ (dotted), prior $\Mdec(\z)$ (blue), and encoding distribution $\x_i \sim \Menc(\x) \rightarrow z_i' \sim  \Menc(\z|\x)$ (yellow).
    As can be seen, parameterizing priors affects all learned distributions, supporting the notion of optimization of a single model $\Mmodel$. We point that (a) demonstrates the best consistency between
    the priors and corresponding generated samples, following the additional expressiveness.
    }
    \label{fig:mim-priors}
\end{figure}

Here we explore the effect of parameterizing the latent and observed priors.
A fundamental idea in \MIM is the concept of a single model, $\Mmodel$. As such,
parameterizing a prior increases the global expressiveness of the model $\Mmodel$.
Fig.\ \ref{fig:mim-priors} depicts the utilization of the added expressiveness in order to increase
the consistency between the encoding and decoding model distribution, in addition to the consistency of $\Mmodel$ with $\Msamp$.

\FloatBarrier


\subsection{Effect of Consistency Regularizer on Optimization}

\begin{figure}[ht]
    \centering
    \begin{minipage}[b]{0.5\columnwidth}
    \centering
    Reconstruction \\
    \begin{minipage}[b]{0.5\columnwidth}
    \centering
    \includegraphics[width=0.99\columnwidth]{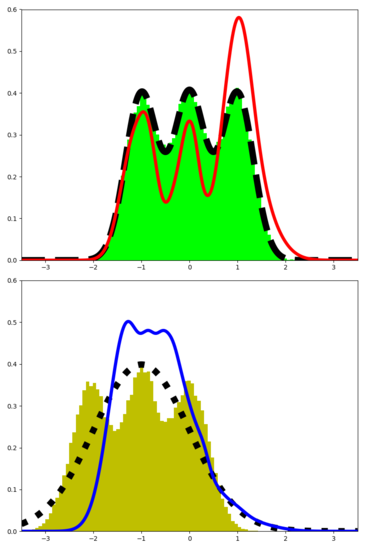}
    \subcaption{$\CEloss$}
    \end{minipage}%
    \begin{minipage}[b]{0.5\columnwidth}
    \centering
    \includegraphics[width=0.99\columnwidth]{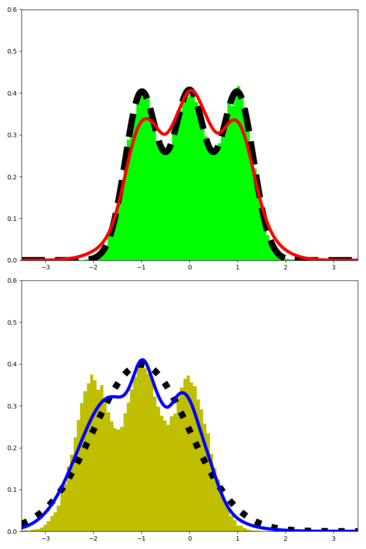}
    \subcaption{$\MIMloss$}
    \end{minipage}
    \end{minipage}%
    \begin{minipage}[b]{0.5\columnwidth}
    \centering
    Prior Sampling \\
    \begin{minipage}[b]{0.5\columnwidth}
    \centering
    \includegraphics[width=0.99\columnwidth]{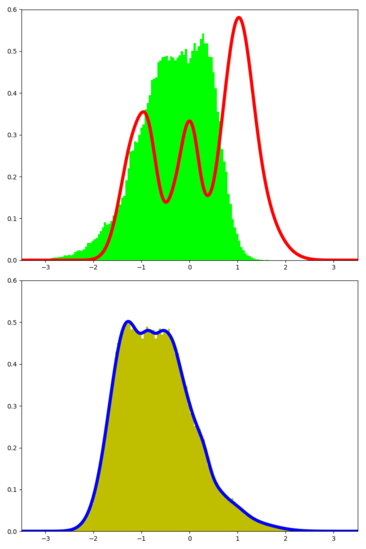}
    \subcaption{$\CEloss$}
    \end{minipage}%
    \begin{minipage}[b]{0.5\columnwidth}
    \centering
    \includegraphics[width=0.99\columnwidth]{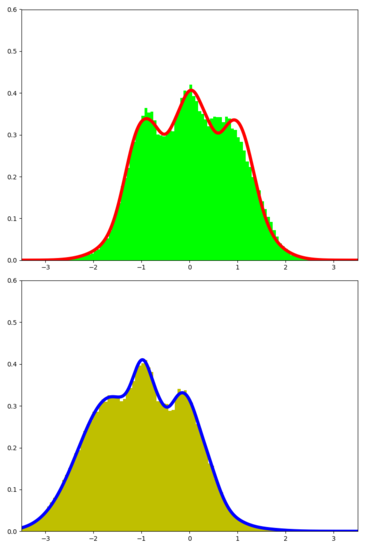}
    \subcaption{$\MIMloss$}
    \end{minipage}
    \end{minipage}
    \begin{minipage}[b]{1.0\columnwidth}
    \centering
    (i) $\CEloss \Longrightarrow \MIMloss$
    \end{minipage}
    \begin{minipage}[b]{0.5\columnwidth}
    \centering
    Reconstruction \\
    \begin{minipage}[b]{0.5\columnwidth}
    \centering
    \includegraphics[width=0.99\columnwidth]{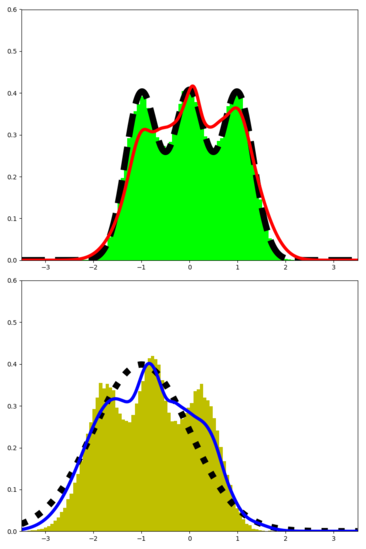}
    \subcaption{$\MIMloss$}
    \end{minipage}%
    \begin{minipage}[b]{0.5\columnwidth}
    \centering
    \includegraphics[width=0.99\columnwidth]{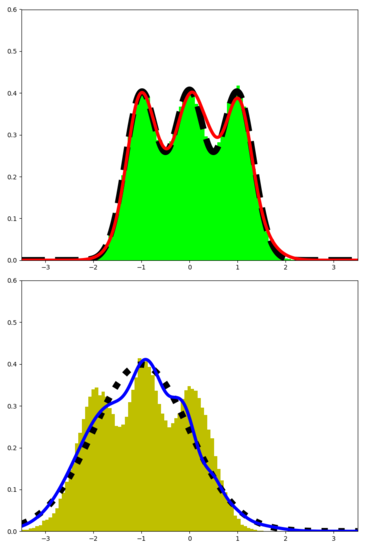}
    \subcaption{$\CEloss$}
    \end{minipage}
    \end{minipage}%
    \begin{minipage}[b]{0.5\columnwidth}
    \centering
    Prior Sampling \\
    \begin{minipage}[b]{0.5\columnwidth}
    \centering
    \includegraphics[width=0.99\columnwidth]{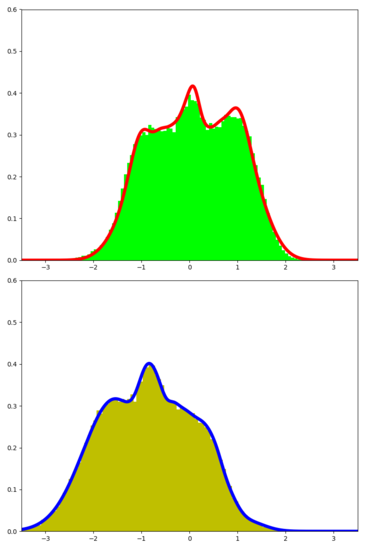}
    \subcaption{$\MIMloss$}
    \end{minipage}%
    \begin{minipage}[b]{0.5\columnwidth}
    \centering
    \includegraphics[width=0.99\columnwidth]{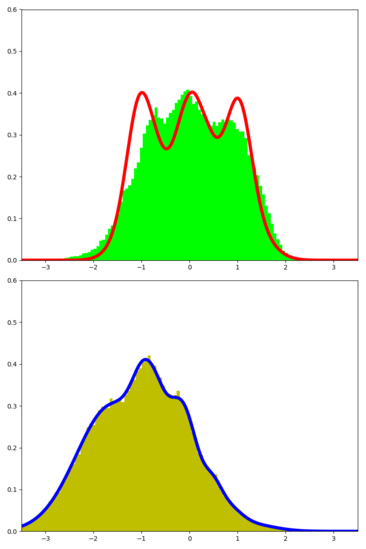}
    \subcaption{$\CEloss$}
    \end{minipage}
    \end{minipage}
    \begin{minipage}[b]{1.0\columnwidth}
    \centering
    (ii) $\MIMloss \Longrightarrow \CEloss$
    \end{minipage}
    \caption{Effects of \MIM consistency regularizer and optimization on encoding-decoding consistency.
    (i) and (ii) differ in initialization order.
    Odd rows: anchor $\pjoint(\x)$ (dashed), prior $\Menc(\x)$ (red). Even rows: anchor $\pjoint(\z)$ (dotted), prior $\Mdec(\z)$ (blue).
    (a-b,e-f) Reconstruction $\x_i \sim \pjoint(\x)\rightarrow \z_i \sim \Menc(\z|\x_i) \rightarrow x_i' \sim \Mdec(\x|\z_i)$ ($x_i'$ green, $\z_i$ yellow).
    (c-d,g-h) Prior decoding $\z_i \sim \Mdec(\z) \rightarrow x_i' \sim \Mdec(\x|\z_i)$ (green), and prior encoding $\x_i \sim \Menc(\x)\rightarrow \z_i \sim \Menc(\z|\x_i)$ (yellow).
    See text for details.
    }
    \label{fig:mim-consistency-optim-vs-loss}
\end{figure}

Here we explore whether a learned model with consistent encoding-decoding distributions (\ie, trained with $\MIMloss$) also constitutes an optimal solution of a CE objective (\ie, trained with $\CEloss$). Results are depicted in Fig. \ref{fig:mim-consistency-optim-vs-loss}. In order to distinguish between the effects of the optimization from the consistency regularizer we initialize a \MIM model by pre-training it with $\CEloss$ loss followed by $\MIMloss$ training in Fig.\ \ref{fig:mim-consistency-optim-vs-loss}(i), and vice verse in Fig.\ \ref{fig:mim-consistency-optim-vs-loss}(ii).
(a-b,e-f) All trained models in Fig. \ref{fig:mim-consistency-optim-vs-loss} exhibit similarly good reconstruction (green matches dashed black). (c-d,g-h) However, only models that were trained with $\MIMloss$ exhibit encoding-decoding consistency (green matches red, yellow matches blue). While it is clear that the optimization plays an important role (\ie, different initialization leads to different local optimum), it is also clear that encoding-decoding consistency is not necessarily an optimum of $\CEloss$, as depicted in a non-consistent model (h) which was initialized with a consistent model (g). Not surprisingly, without the consistency regularizer training with $\CEloss$ results in better fit of priors to anchors (f) as it is utilizing the expressiveness of the parametric priors in matching the sample distribution.

\FloatBarrier

\subsection{Bottleneck in Low Dimensional Synthetic Data}

\begin{figure}[ht]
    \centering
    \setlength{\tabcolsep}{0pt}
    \begin{tabular}{*4{>{\centering\arraybackslash}m{0.25\textwidth}}}
     \includegraphics[width=0.24\columnwidth]{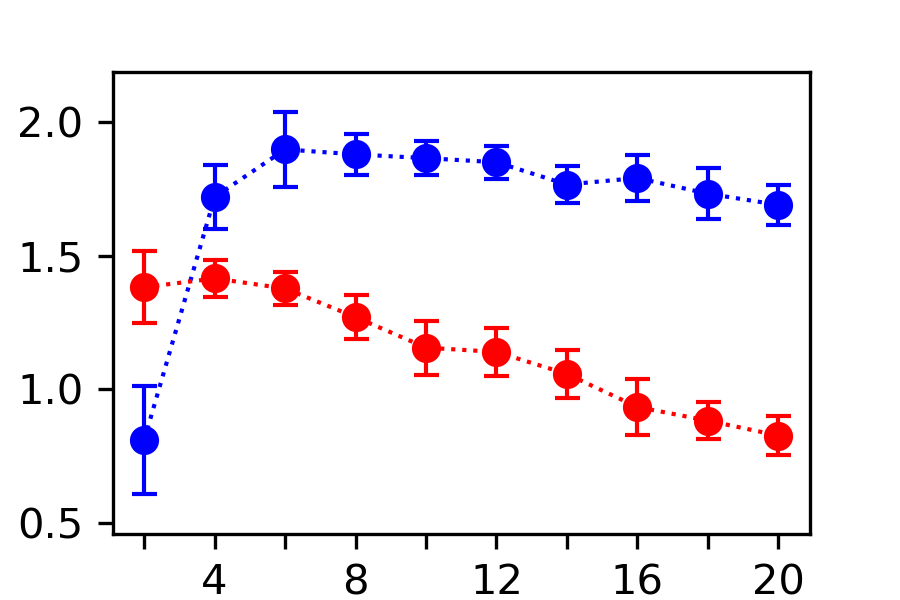}
    & \includegraphics[width=0.24\columnwidth]{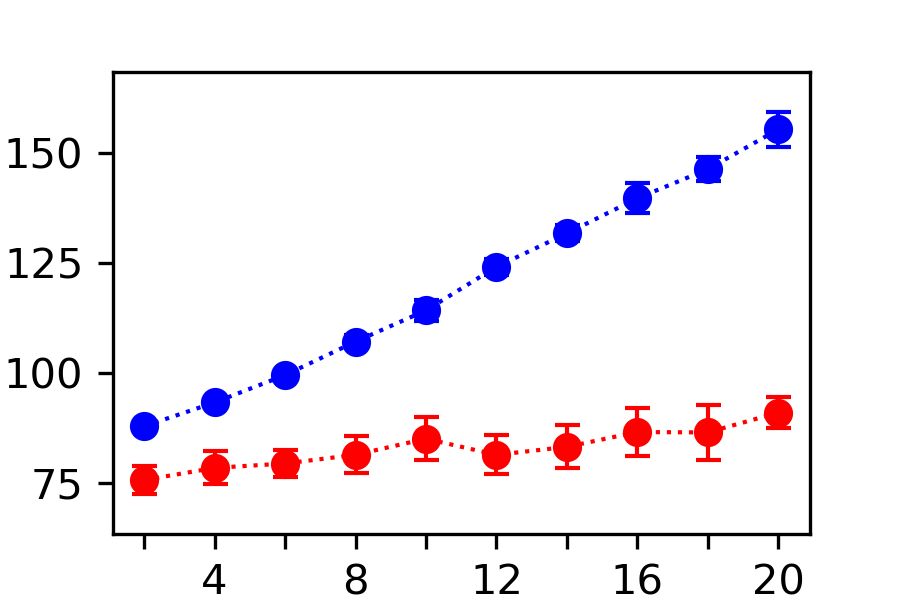}
    & \includegraphics[width=0.24\columnwidth]{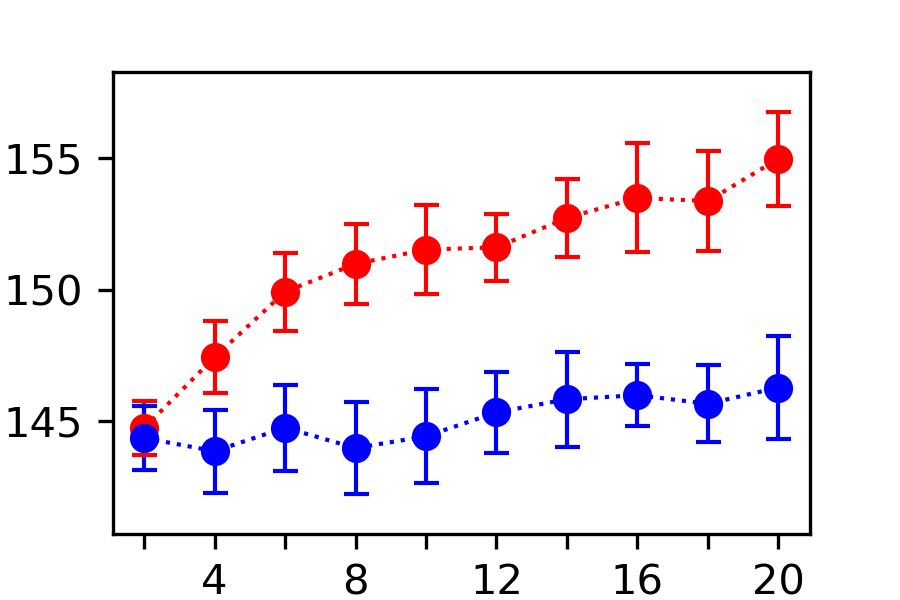}
    & \includegraphics[width=0.24\columnwidth]{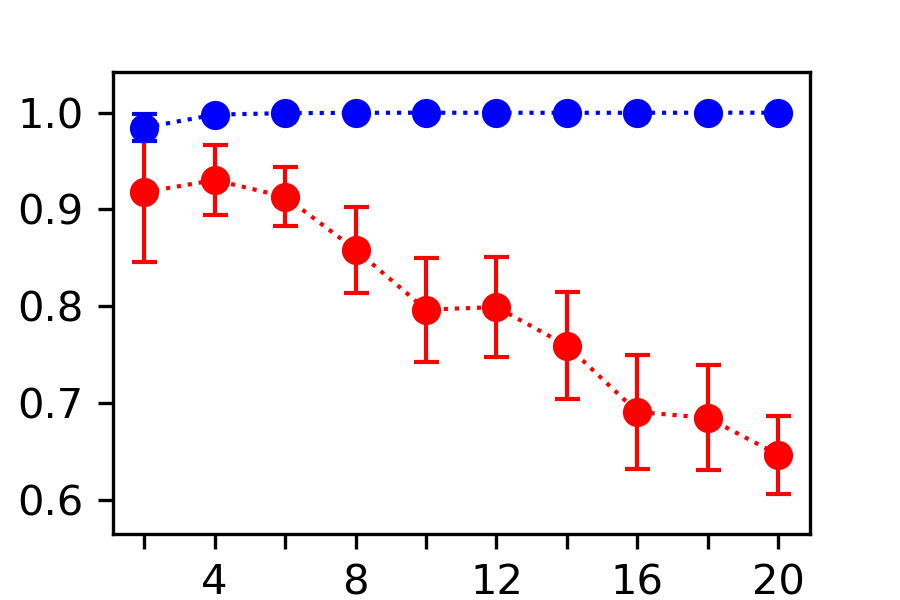}
    \\
    (a) MI & (b)  NLL  & (c) Recon.\ Error & (d) Classif.\ (5-NN)
    \end{tabular}
    \caption{
    MIM (blue) and VAE (red) for 20D GMM data,
    all as function of the latent dimensionality, from 2 to 20
    (on x-axis). Plots depict mean and standard deviation of 10 experiments.
    MIM learning produces higher mutual information and classification accuracy,
    with lower test reconstruction error, while VAE yields better data log likelihoods.
    The VAE suffers from increased collapse as latent dimensionality grows.
    }
    \label{fig:mim-vs-vae-quantitative-bottleneck-toy-x20}
\end{figure}

Here we consider synthetic 20D data from a 5-component GMM, with independent training and test sets, and latent representation between 2D and 20D.
This ensures that the distribution is well modeled with a relatively simple
architecture, like that in Sec.\ \ref{sec:posterior-collapse-mim-vae}.
This experiment extends the experiment in Fig.\ \ref{fig:posterior-collapse-quantitative} by adding a bottleneck, similar to the experiment in Fig.\ \ref{fig:mim-vs-vae-quantitative-bottleneck-fashion-mnist-pca}.
We used here the same experimental setup that was used in Fig.\ \ref{fig:posterior-collapse-quantitative}.

Results are shown in Fig.\ \ref{fig:mim-vs-vae-quantitative-bottleneck-toy-x20}.
MIM produces higher mutual information, and better classification, as the latent dimensionality grows, while VAE
increasingly suffers from posterior collapse, which leads to lower mutual information, and lower classification accuracy.
The test negative log likelihood scores (NLL) of MIM are not as good as VAE scores
in part because the MIM encoder produces very small posterior variance,
approaching a deterministic encoder. Nevertheless, MIM produces lower test
reconstruction errors.
These results are consistent with those in Sec.\ \ref{sec:posterior-collapse-mim-vae}.

\FloatBarrier



\section{Additional Results}
\label{sec:additional-results}

Here we provide additional visualization of various MIM and VAE models which were not included in the main body of the paper.

\subsection{Reconstruction and Samples for MIM and A-MIM}

In what follows we show samples and reconstruction for MIM (\ie, with convHVAE architecture), and A-MIM (\ie, with PixelHVAE architecture).
We demonstrate, again, that a powerful enough encoder allows for generation of samples which are comparable to VAE samples.

\begin{figure}[H]
    \centering
    \setlength{\tabcolsep}{0pt}
    \begin{tabular}{*4{>{\centering\arraybackslash}m{0.25\textwidth}}}
    \multicolumn{2}{c}{Standard Prior} & \multicolumn{2}{c}{VampPrior Prior} \\
    \multicolumn{2}{c}{\includegraphics[width=0.5\columnwidth]{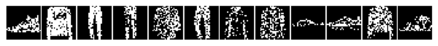}}
    & \multicolumn{2}{c}{\includegraphics[width=0.5\columnwidth]{images/vae-as-mim-image/dynamic_fashion_mnist_pixelhvae_2level_vampprior__K_500__wu_100__z1_40_z2_40/real_flat.png}}
    \\[-0.2cm]
    \multicolumn{2}{c}{\includegraphics[width=0.5\columnwidth]{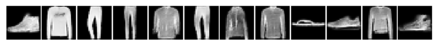}}
    & \multicolumn{2}{c}{\includegraphics[width=0.5\columnwidth]{images/vae-as-mim-image/dynamic_fashion_mnist_pixelhvae_2level_vampprior__K_500__wu_100__z1_40_z2_40/reconstructions_flat.png}}
    \\[-0.2cm]
    \multicolumn{2}{c}{\includegraphics[width=0.5\columnwidth]{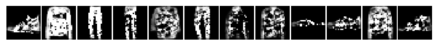}}
    & \multicolumn{2}{c}{\includegraphics[width=0.5\columnwidth]{images/vae-as-mim-image/dynamic_fashion_mnist_pixelhvae_2level-amim_vampprior__K_500__wu_100__z1_40_z2_40/reconstructions_flat.png}}
    \\
    \includegraphics[width=0.25\columnwidth]{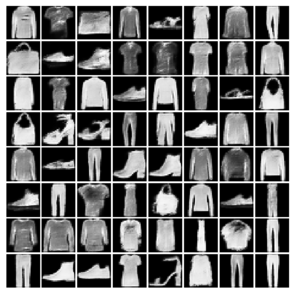}
    & \includegraphics[width=0.25\columnwidth]{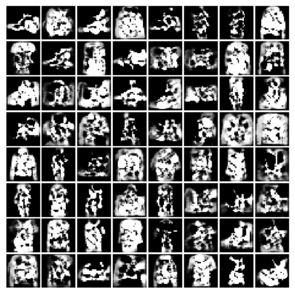}
    & \includegraphics[width=0.25\columnwidth]{images/vae-as-mim-image/dynamic_fashion_mnist_pixelhvae_2level_vampprior__K_500__wu_100__z1_40_z2_40/generations.png}
    & \includegraphics[width=0.25\columnwidth]{images/vae-as-mim-image/dynamic_fashion_mnist_pixelhvae_2level-amim_vampprior__K_500__wu_100__z1_40_z2_40/generations.png}
    \\
    (a) VAE (S) & (b) A-MIM (S) & (c) VAE (VP) & (d) A-MIM (VP)
    \end{tabular}
    \caption{
    MIM and VAE learning with PixelHVAE for Fashion MNIST.
    The top three rows (from top to bottom)  are test data samples, VAE reconstruction, A-MIM reconstruction.
    Bottom: random samples from VAE and MIM.
    (c-d) We initialized all pseudo-inputs with training samples, and used the same random seed for both models. As a result the samples order is similar.
    }
    \label{fig:mim-vs-vae-image-qualitative-fashion-mnist-pixelcnn-extra}
\end{figure}

\begin{figure}[H]
    \centering
    \setlength{\tabcolsep}{0pt}
    \begin{tabular}{*4{>{\centering\arraybackslash}m{0.25\textwidth}}}
    \multicolumn{2}{c}{Standard Prior} & \multicolumn{2}{c}{VampPrior Prior} \\
    \multicolumn{2}{c}{\includegraphics[width=0.5\columnwidth]{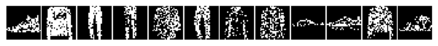}}
    & \multicolumn{2}{c}{\includegraphics[width=0.5\columnwidth]{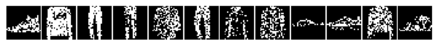}}
    \\[-0.2cm]
    \multicolumn{2}{c}{\includegraphics[width=0.5\columnwidth]{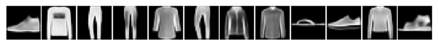}}
    & \multicolumn{2}{c}{\includegraphics[width=0.5\columnwidth]{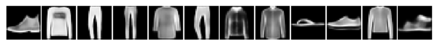}}
    \\[-0.2cm]
    \multicolumn{2}{c}{\includegraphics[width=0.5\columnwidth]{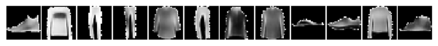}}
    & \multicolumn{2}{c}{\includegraphics[width=0.5\columnwidth]{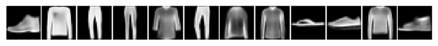}}
    \\
    \includegraphics[width=0.25\columnwidth]{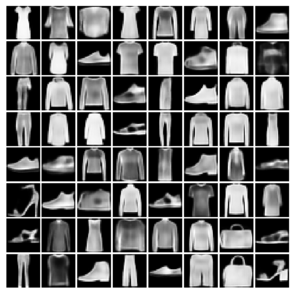}
    & \includegraphics[width=0.25\columnwidth]{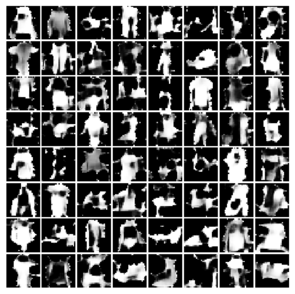}
    & \includegraphics[width=0.25\columnwidth]{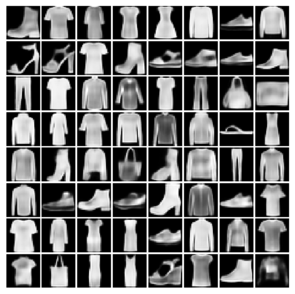}
    & \includegraphics[width=0.25\columnwidth]{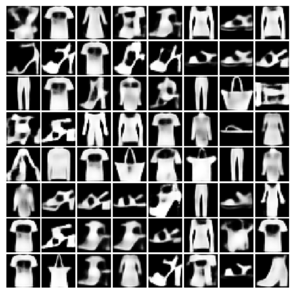}
    \\
    (a) VAE (S) & (b) MIM (S) & (c) VAE (VP) & (d) MIM (VP)
    \end{tabular}
    \caption{
    MIM and VAE learning with convHVAE for Fashion MNIST.
    The top three rows (from top to bottom)  are test data samples, VAE reconstruction, MIM reconstruction.
    Bottom: random samples from VAE and MIM.
    }
    \label{fig:mim-vs-vae-image-qualitative-fashion-mnist-conv-extra}
\end{figure}

\begin{figure}[H]
    \centering
    \setlength{\tabcolsep}{0pt}
    \begin{tabular}{*4{>{\centering\arraybackslash}m{0.25\textwidth}}}
    \multicolumn{2}{c}{Standard Prior} & \multicolumn{2}{c}{VampPrior Prior} \\
    \multicolumn{2}{c}{\includegraphics[width=0.5\columnwidth]{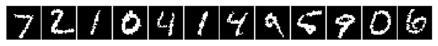}}
    & \multicolumn{2}{c}{\includegraphics[width=0.5\columnwidth]{images/vae-as-mim-image/dynamic_mnist_pixelhvae_2level_vampprior__K_500__wu_100__z1_40_z2_40/real_flat.png}}
    \\[-0.2cm]
    \multicolumn{2}{c}{\includegraphics[width=0.5\columnwidth]{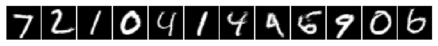}}
    & \multicolumn{2}{c}{\includegraphics[width=0.5\columnwidth]{images/vae-as-mim-image/dynamic_mnist_pixelhvae_2level_vampprior__K_500__wu_100__z1_40_z2_40/reconstructions_flat.png}}
    \\[-0.2cm]
    \multicolumn{2}{c}{\includegraphics[width=0.5\columnwidth]{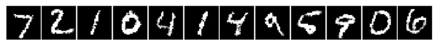}}
    & \multicolumn{2}{c}{\includegraphics[width=0.5\columnwidth]{images/vae-as-mim-image/dynamic_mnist_pixelhvae_2level-amim_vampprior__K_500__wu_100__z1_40_z2_40/reconstructions_flat.png}}
    \\
    \includegraphics[width=0.25\columnwidth]{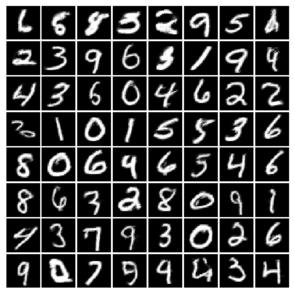}
    & \includegraphics[width=0.25\columnwidth]{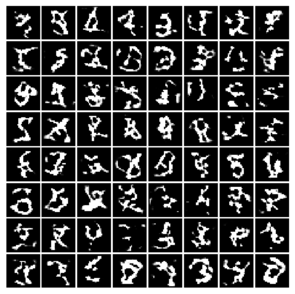}
    & \includegraphics[width=0.25\columnwidth]{images/vae-as-mim-image/dynamic_mnist_pixelhvae_2level_vampprior__K_500__wu_100__z1_40_z2_40/generations.png}
    & \includegraphics[width=0.25\columnwidth]{images/vae-as-mim-image/dynamic_mnist_pixelhvae_2level-amim_vampprior__K_500__wu_100__z1_40_z2_40/generations.png}
    \\
    (a) VAE (S) & (b) A-MIM (S) & (c) VAE (VP) & (d) A-MIM (VP)
    \end{tabular}
    \caption{
    MIM and VAE learning with PixelHVAE for MNIST.
    Top three rows are test data samples, followed by VAE and A-MIM reconstructions.
    Bottom: random samples from VAE and MIM.
    (c-d) We initialized all pseudo-inputs with training samples, and used the same random seed for both models. As a result the samples order is similar.
    }
    \label{fig:mim-vs-vae-image-qualitative-mnist-pixelcnn-extra}
\end{figure}

\begin{figure}[H]
    \centering
    \setlength{\tabcolsep}{0pt}
    \begin{tabular}{*4{>{\centering\arraybackslash}m{0.25\textwidth}}}
    \multicolumn{2}{c}{Standard Prior} & \multicolumn{2}{c}{VampPrior Prior} \\
    \multicolumn{2}{c}{\includegraphics[width=0.5\columnwidth]{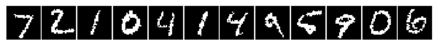}}
    & \multicolumn{2}{c}{\includegraphics[width=0.5\columnwidth]{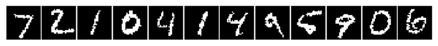}}
    \\[-0.2cm]
    \multicolumn{2}{c}{\includegraphics[width=0.5\columnwidth]{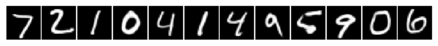}}
    & \multicolumn{2}{c}{\includegraphics[width=0.5\columnwidth]{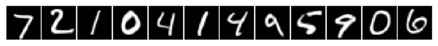}}
    \\[-0.2cm]
    \multicolumn{2}{c}{\includegraphics[width=0.5\columnwidth]{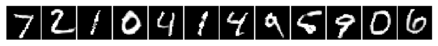}}
    & \multicolumn{2}{c}{\includegraphics[width=0.5\columnwidth]{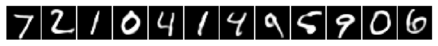}}
    \\
    \includegraphics[width=0.25\columnwidth]{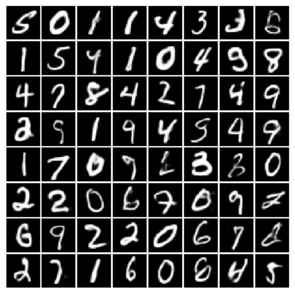}
    & \includegraphics[width=0.25\columnwidth]{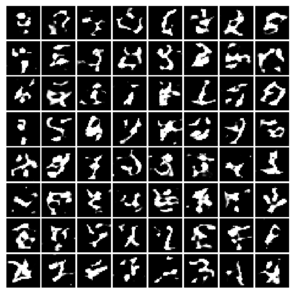}
    & \includegraphics[width=0.25\columnwidth]{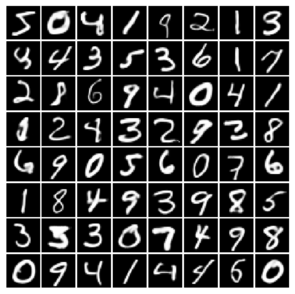}
    & \includegraphics[width=0.25\columnwidth]{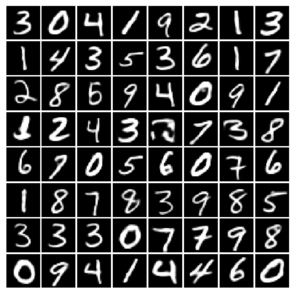}
    \\
    (a) VAE (S) & (b) MIM (S) & (c) VAE (VP) & (d) MIM (VP)
    \end{tabular}
    \caption{
    MIM and VAE learning with convHVAE for MNIST.
    Top three rows are test data samples, followed by VAE and MIM reconstructions.
    Bottom: random samples from VAE and MIM.
    }
    \label{fig:mim-vs-vae-image-qualitative-mnist-conv-extra}
\end{figure}

\begin{figure}[H]
    \centering
    \setlength{\tabcolsep}{0pt}
    \begin{tabular}{*4{>{\centering\arraybackslash}m{0.25\textwidth}}}
    \multicolumn{2}{c}{Standard Prior} & \multicolumn{2}{c}{VampPrior Prior} \\
    \multicolumn{2}{c}{\includegraphics[width=0.5\columnwidth]{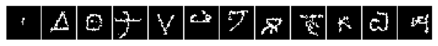}}
    & \multicolumn{2}{c}{\includegraphics[width=0.5\columnwidth]{images/vae-as-mim-image/omniglot_pixelhvae_2level_vampprior__K_500__wu_100__z1_40_z2_40/real_flat.png}}
    \\[-0.2cm]
    \multicolumn{2}{c}{\includegraphics[width=0.5\columnwidth]{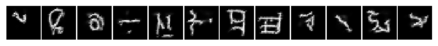}}
    & \multicolumn{2}{c}{\includegraphics[width=0.5\columnwidth]{images/vae-as-mim-image/omniglot_pixelhvae_2level_vampprior__K_500__wu_100__z1_40_z2_40/reconstructions_flat.png}}
    \\[-0.2cm]
    \multicolumn{2}{c}{\includegraphics[width=0.5\columnwidth]{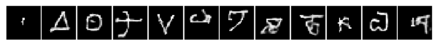}}
    & \multicolumn{2}{c}{\includegraphics[width=0.5\columnwidth]{images/vae-as-mim-image/omniglot_pixelhvae_2level-amim_vampprior__K_500__wu_100__z1_40_z2_40/reconstructions_flat.png}}
    \\
    \includegraphics[width=0.25\columnwidth]{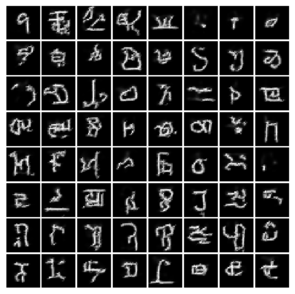}
    & \includegraphics[width=0.25\columnwidth]{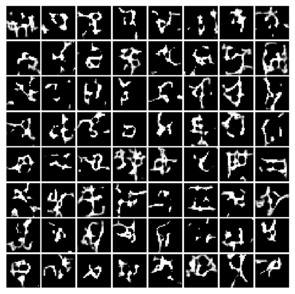}
    & \includegraphics[width=0.25\columnwidth]{images/vae-as-mim-image/omniglot_pixelhvae_2level_vampprior__K_500__wu_100__z1_40_z2_40/generations.png}
    & \includegraphics[width=0.25\columnwidth]{images/vae-as-mim-image/omniglot_pixelhvae_2level-amim_vampprior__K_500__wu_100__z1_40_z2_40/generations.png}
    \\
    (a) VAE (S) & (b) A-MIM (S) & (c) VAE (VP) & (d) A-MIM (VP)
    \end{tabular}
    \caption{
    MIM and VAE learning with PixelHVAE for Omniglot.
    Top three rows are test data samples, followed by VAE and A-MIM reconstructions.
    Bottom: random samples from VAE and MIM.
    }
    \label{fig:mim-vs-vae-image-qualitative-omniglot-pixelcnn-extra}
\end{figure}

\begin{figure}[H]
    \centering
    \setlength{\tabcolsep}{0pt}
    \begin{tabular}{*4{>{\centering\arraybackslash}m{0.25\textwidth}}}
    \multicolumn{2}{c}{Standard Prior} & \multicolumn{2}{c}{VampPrior Prior} \\
    \multicolumn{2}{c}{\includegraphics[width=0.5\columnwidth]{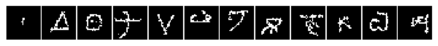}}
    & \multicolumn{2}{c}{\includegraphics[width=0.5\columnwidth]{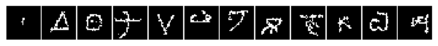}}
    \\[-0.2cm]
    \multicolumn{2}{c}{\includegraphics[width=0.5\columnwidth]{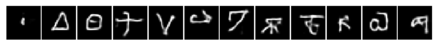}}
    & \multicolumn{2}{c}{\includegraphics[width=0.5\columnwidth]{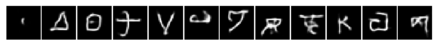}}
    \\[-0.2cm]
    \multicolumn{2}{c}{\includegraphics[width=0.5\columnwidth]{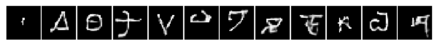}}
    & \multicolumn{2}{c}{\includegraphics[width=0.5\columnwidth]{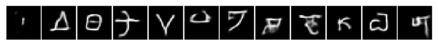}}
    \\
    \includegraphics[width=0.25\columnwidth]{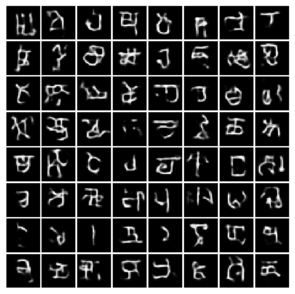}
    & \includegraphics[width=0.25\columnwidth]{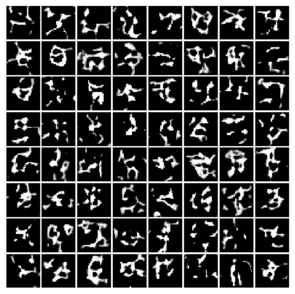}
    & \includegraphics[width=0.25\columnwidth]{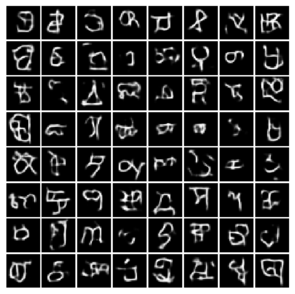}
    & \includegraphics[width=0.25\columnwidth]{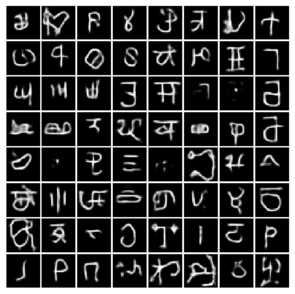}
    \\
    (a) VAE (S) & (b) MIM (S) & (c) VAE (VP) & (d) MIM (VP)
    \end{tabular}
    \caption{
    MIM and VAE learning with convHVAE for Omniglot.
    Top three rows are test data samples, followed by VAE and MIM reconstructions.
    Bottom: random samples from VAE and MIM.
    }
    \label{fig:mim-vs-vae-image-qualitative-omniglot-conv-extra}
\end{figure}

\FloatBarrier

\subsection{Latent Embeddings for MIM and A-MIM}

In what follows we show additional t-SNE visualization of unsupervised clustering in the latent representation for MIM (\ie, with convHVAE architecture), and A-MIM (\ie, with PixelHVAE architecture).

\begin{figure}[H]
    \centering
    \setlength{\tabcolsep}{0pt}
    \begin{tabular}{*4{>{\centering\arraybackslash}m{0.25\textwidth}}}
    \includegraphics[width=0.25\columnwidth]{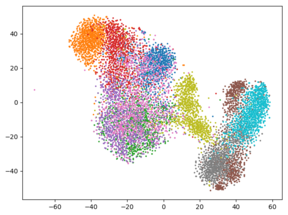}
    & \includegraphics[width=0.25\columnwidth]{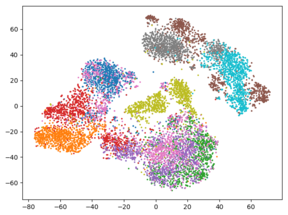}
    & \includegraphics[width=0.25\columnwidth]{images/vae-as-mim-image/dynamic_fashion_mnist_convhvae_2level_vampprior__K_500__wu_100__z1_40_z2_40/z_embed.png}
    & \includegraphics[width=0.25\columnwidth]{images/vae-as-mim-image/dynamic_fashion_mnist_convhvae_2level-smim_vampprior__K_500__wu_100__z1_40_z2_40/z_embed.png}
    \\
     (a) VAE (S)  & (b) MIM (S) & (c) VAE (VP)  & (d) MIM (VP) \\
    \end{tabular}
    \caption{MIM and VAE $\z$ embedding for Fashion MNIST with convHVAE architecture.
    }
    \label{fig:mim-vs-vae-image-z-embed-fashion-mnist-extra}
\end{figure}

\begin{figure}[H]
    \centering
    \setlength{\tabcolsep}{0pt}
    \begin{tabular}{*4{>{\centering\arraybackslash}m{0.25\textwidth}}}
    \includegraphics[width=0.25\columnwidth]{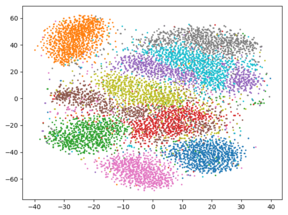}
    & \includegraphics[width=0.25\columnwidth]{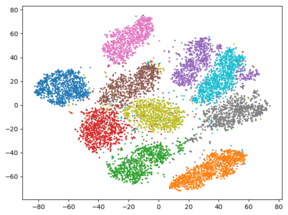}
    & \includegraphics[width=0.25\columnwidth]{images/vae-as-mim-image/dynamic_mnist_convhvae_2level_vampprior__K_500__wu_100__z1_40_z2_40/z_embed.png}
    & \includegraphics[width=0.25\columnwidth]{images/vae-as-mim-image/dynamic_mnist_convhvae_2level-smim_vampprior__K_500__wu_100__z1_40_z2_40/z_embed.png}
    \\
     (a) VAE (S)  & (b) MIM (S) & (c) VAE (VP)  & (d) MIM (VP) \\
    \end{tabular}
    \caption{MIM and VAE $\z$ embedding for MNIST with convHVAE architecture.
    }
    \label{fig:mim-vs-vae-image-z-embed-mnist-extra}
\end{figure}

\begin{figure}[H]
    \centering
    \setlength{\tabcolsep}{0pt}
    \begin{tabular}{*4{>{\centering\arraybackslash}m{0.25\textwidth}}}
    \includegraphics[width=0.25\columnwidth]{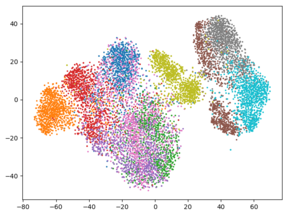}
    & \includegraphics[width=0.25\columnwidth]{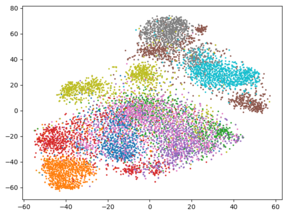}
    & \includegraphics[width=0.25\columnwidth]{images/vae-as-mim-image/dynamic_fashion_mnist_pixelhvae_2level_vampprior__K_500__wu_100__z1_40_z2_40/z_embed.png}
    & \includegraphics[width=0.25\columnwidth]{images/vae-as-mim-image/dynamic_fashion_mnist_pixelhvae_2level-amim_vampprior__K_500__wu_100__z1_40_z2_40/z_embed.png}
    \\
     (a) VAE (S)  & (b) A-MIM (S) & (c) VAE (VP)  & (d) A-MIM (VP) \\
    \end{tabular}
    \caption{A-MIM and VAE $\z$ embedding for Fashion MNIST with PixelHVAE architecture.
    }
    \label{fig:mim-vs-vae-image-z-embed-pixelcnn-fashion-mnist-extra}
\end{figure}

\begin{figure}[H]
    \centering
    \setlength{\tabcolsep}{0pt}
    \begin{tabular}{*4{>{\centering\arraybackslash}m{0.25\textwidth}}}
    \includegraphics[width=0.25\columnwidth]{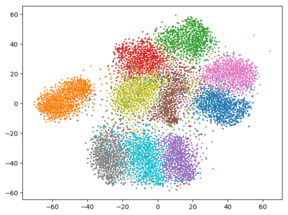}
    & \includegraphics[width=0.25\columnwidth]{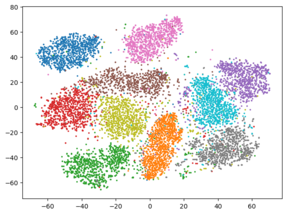}
    & \includegraphics[width=0.25\columnwidth]{images/vae-as-mim-image/dynamic_mnist_pixelhvae_2level_vampprior__K_500__wu_100__z1_40_z2_40/z_embed.png}
    & \includegraphics[width=0.25\columnwidth]{images/vae-as-mim-image/dynamic_mnist_pixelhvae_2level-amim_vampprior__K_500__wu_100__z1_40_z2_40/z_embed.png}
    \\
     (a) VAE (S)  & (b) A-MIM (S) & (c) VAE (VP)  & (d) A-MIM (VP) \\
    \end{tabular}
    \caption{A-MIM and VAE $\z$ embedding for MNIST with PixelHVAE architecture.
    }
    \label{fig:mim-vs-vae-image-z-embed-pixelcnn-mnist-extra}
\end{figure}

\FloatBarrier






\end{document}